%% file: main_TIT2025.tex
\documentclass[onecolumn,draftclsnofoot]{IEEEtran}
\IEEEoverridecommandlockouts

\usepackage[utf8]{inputenc} 
\usepackage[T1]{fontenc}
\usepackage{hyperref}
\usepackage{url}
\usepackage{ifthen}
\usepackage{cite}
\usepackage[cmex10]{amsmath} 

\usepackage{xcolor}         
\usepackage{subcaption}

\usepackage{bm,amsthm,mathtools,amsfonts,amssymb,bbm,setspace,enumitem,wrapfig}
\usepackage{graphicx} 
\graphicspath{{Fig/}} 

\input{defs_IT.tex}

\usepackage{mleftright}\mleftright

\title{Heterogeneity Matters even More in Distributed Learning: Study from Generalization Perspective} 

\author{%
  \IEEEauthorblockN{Masoud Kavian$\:^{\dagger}$$\:^{\nmid}$ \qquad \qquad Romain Chor$\:^{\dagger}$$\:^{\nmid}$ \qquad \qquad Milad  Sefidgaran$^{\:\nmid}$ \qquad \qquad Abdellatif Zaidi$\:^{\dagger}$$^{\:\nmid}$}\\
  \IEEEauthorblockA{$^{\:\nmid}$  Paris Research Center, Huawei Technologies France \qquad \qquad
  $\:^{\dagger}$ Universit\'e Gustave Eiffel, France\\
  \texttt{\{masoud.kavian, milad.sefidgaran2\}@huawei.com, abdellatif.zaidi@univ-eiffel.fr, romain.chor3@edu.univ-eiffel.fr}}
}

\begin{document}
\fontencoding{OT1}\fontsize{10}{11}\selectfont

\maketitle

\begin{abstract}
In this paper, we investigate the effect of data heterogeneity across clients on the performance of distributed learning systems, i.e., one-round Federated Learning, as measured by the associated generalization error. Specifically, \(K\) clients have each \(n\) training samples generated independently according to a possibly different data distribution, and their individually chosen models are aggregated by a central server. We study the effect of the discrepancy between the clients' data distributions on the generalization error of the aggregated model. First, we establish in-expectation and tail upper bounds on the generalization error in terms of the distributions. In part, the bounds extend the popular Conditional Mutual Information (CMI) bound, which was developed for the centralized learning setting, i.e., \(K=1\), to the distributed learning setting with an arbitrary number of clients $K \geq 1$. Then, we connect with information-theoretic rate-distortion theory to derive possibly tighter \textit{lossy} versions of these bounds. Next, we apply our lossy bounds to study the effect of data heterogeneity across clients on the generalization error for the distributed classification problem in which each client uses Support Vector Machines (DSVM). In this case, we establish explicit generalization error bounds that depend explicitly on the data heterogeneity degree. It is shown that the bound gets smaller as the degree of data heterogeneity across clients increases, 
 thereby suggesting that DSVM generalizes better when the dissimilarity between the clients' training samples is bigger. This finding, which goes beyond DSVM, is validated experimentally through several experiments.
\end{abstract}


\section{Introduction}
\IEEEPARstart{A} major focus of machine learning research over recent years has been the study of statistical learning algorithms when applied in distributed (network or graph) settings. 
In part, this is due to the emergence of new applications in which resources are constrained, data is distributed, or the need to preserve privacy. 
Examples of such algorithms include the now popular Federated Learning~\cite{mcmahan2017communication,konevcny2016federated,yuan2021we}, the Split Learning of~\cite{gupta2018distributed}, or the so-called in-network learning of~\cite{aguerri2019distributed,moldoveanu2023network}. Some of these algorithms enable collaborative training across multiple devices by distilling knowledge (features) locally, i.e., without sharing the raw data. This has given rise to privacy-preserving machine learning~\cite{li2020federatedsurvey,kairouz2021advances} and has led to rapid advancements in communication-efficient and privacy-aware distributed learning algorithms, such ass differential privacy mechanisms~\cite{dwork2014algorithmic,geyer2017differentially,abadi2016deep}. 
Despite its importance, however, little is known about the generalization guarantees of distributed statistical learning algorithms, including a lack of proper definitions~\cite{yuan2021we,mohri2019agnostic,zinkevich2010parallelized,kairouz2021advances}. Notable exceptions include the related works~\cite{yagli2020information,barnes2022improved,zhang2024improving,Sefidgaran2022,chor2023more, sefidgaran2024lessons} and \cite{hu2023generalization} or the works~\cite{aleks2020pac,neyshabur2017pac,li2020federatedpac} which essentially adapt classical bounds to account for partitioned data and constrained communication~\cite{aleks2020pac,neyshabur2017pac,li2020federatedpac} using PAC-Bayes theory. The lack of understanding of what really controls generalization in distributed learning settings is even more pronounced when the (training) data exhibits some degree of \textit{heterogeneity} across participating clients or devices. That is, when the underlying probability distributions (if there exist such distributions!) vary across those clients. From an algorithmic perspective, works appeared that combine personalized federated learning with clustered training methods in an attempt to adapt models to heterogeneous data across clients while maintaining collaborative learning benefits~\cite{smith2017federated,arivazhagan2019federated,sattler2020clustered}. In essence, these approaches strike a balance between global model performance and local data specificities. Also, compression and sparsification strategies were proposed and shown to partly alleviate the effect of data heterogeneity~\cite{aji2017sparse,reisizadeh2020fedpaq,lin2018deep}. 

The question of the effect of data heterogeneity on the performance of statistical learning algorithms is not yet fully understood from a theoretical perspective, however. This is true even from a convergence rate analysis viewpoint, a line of work that is more studied comparatively. For example, while it has been reported that non-independently and/or non-identically distributed (non-IID) data slow down convergence in FL-type algorithms~\cite{wang2022unreasonable,zhang2021fedpd,mitra2021linear}, optimal rates are still unknown in general; and, how that slowness relates to the behavior of the generalization error is yet to be un-explored.

\begin{figure}
    \centering
    \includegraphics[width=0.4\linewidth]{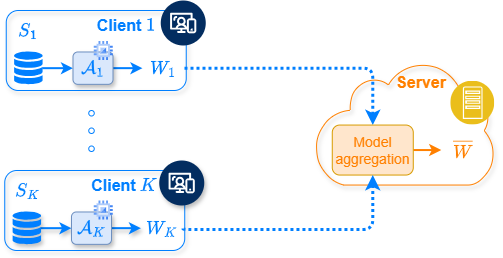}
   \caption{Studied distributed learning problem}\vspace{-0.3 cm}
    \label{fig-system-model}
\end{figure}

In this paper, we study the distributed learning system shown in Fig.~\ref{fig-system-model}. Here, there are \(K\) clients; each having access to a training dataset \(S_k = \{Z_{k,1}, \dots, Z_{k,n}\} \in \mathcal{Z}^n \) of size \(n\), where the data samples \(\{Z_{k,1}, \dots, Z_{k,n}\}\) are generated independently from each other and other clients' training samples according to a probability distribution \(\mu_k\). The probability distributions \(\{\mu_k\}_{k=1}^K\) are possibly distinct, i.e., \textit{heterogeneous} across clients. In the special case in which \(\mu_k = \mu\) for all \(k=1,\hdots, K\), we will refer to the setting as being \textit{homogeneous}. Client \(k\) applies a possibly  stochastic learning algorithm \(A_k: \mathcal{Z}^n_k \rightarrow \mathcal{W}_k\). This induces a conditional distribution \(P_{W_k | S_k}\), which together with \(\mu_k\) induce the joint dataset-hypothesis distribution $P_{W_k, S_k} = \mu_k^{\otimes n} P_{W_k | S_k}$. The server receives \( (W_1,\hdots,W_K)\)  and picks the hypothesis \(\overline{W}\) as the arithmetic average
\begin{equation}
\overline{W} = \frac{1}{K} \sum\limits_{k=1}^{K} W_k.
\label{model-aggregation-at-server}
\end{equation}
We investigate the effect of the discrepancy between the clients' data distributions on the generalization performance of the aggregated model \(\overline{W}\). In particular, for given loss function \(\ell : \mathcal{Z} \times \mathcal{W} \to [0,1]\) used to evaluate the quality of the prediction and a proper definition of the generalization error (see formal definitions in Section~\ref{sec:problem_statement}), we ask the following question:
\begin{center}
\textit{How does the generalization error of the aggregated model \(\overline{W}\) evolve as a function of a measure of discrepancy between the data distributions \(\mu_1,\hdots,\mu_K\) ?}
\end{center}

\subsection{Main contributions}
The main contributions of this paper are as follows:
\begin{itemize}[leftmargin=0.5cm]
\item We establish (general) in-expectation and tail upper bounds on the generalization error in terms of the distributions \((\mu_1,\hdots,\mu_K)\). In part, the bounds extend the Conditional Mutual Information (CMI) bound, which was developed for the centralized learning, i.e., \(K=1\) in~\cite{steinke2020reasoning}, to the distributed learning setting of Fig.~\ref{fig-system-model}.
\item We use a connection between the theory of generalization of statistical learning algorithms and information-theoretic rate-distortion theory that was introduced in~\cite{Sefidgaran2022} and subsequently used and elaborated on in~\cite{sefidgaran2024data,sefidgaran2024minimum,sefidgaran2022rate}, to obtain possibly tighter \textit{lossy} versions of these bounds. Furthermore, we also provide improved bounds that are based on the Jensen-Shannon divergence.
\item We apply our established \textit{lossy} bounds to study the effect of data heterogeneity across clients on the generalization error for a distributed classification problem in which each client uses Support Vector Machines (SVM). In this case, we establish in-expectation generalization bounds that depend explicitly on the degree of data heterogeneity across clients; and, by comparing them, we show that the bounds get better (i.e., smaller) as the degree of data heterogeneity across clients increases. Also, the bounds increase as the total variation between the distributions becomes smaller. 
\vspace{-0.1 cm}
\item We report experimental results that validate the theoretical results of this paper in various settings. This involves both feature and label heterogeneities, and for various datasets and algorithms such as D-SVM and using neural networks. Specifically, the conducted experiments on the effect of feature heterogeneity are on (i) the MNIST dataset with 
noise-induced feature heterogeneity, (ii) the MNIST dataset with digit structure-induced feature heterogeneity and (iii) synthetic data. For the experiments on the effect of label heterogeneity across clients we somewhat stretch the setup of this paper by allowing multi-round interactions between the clients and server, i.e., multi-round FL. In this case, the experiments are on (i) the MNIST dataset with unbalanced label partition across the clients and (ii) the CIFAR10 dataset with unbalanced label partition across clients.
\end{itemize}

\subsection{Relation to prior art} 

On the line of work investigating the effect of heterogeneity on the performance of distributed and FL-type learning systems, most related to our work here are the works~\cite{sun2024understanding} and~\cite{wang2025generalization} and, to a lesser extent,~\cite{liu2023exploiting}. In~\cite{sun2024understanding}, the authors analyze the generalization error of FL using algorithmic stability. Also, they report experimental results for a 10-class MNIST type classification problem which show that, when trained with the algorithm FedAvg, label heterogeneity across clients increases the generalization error (see~\cite[Fig. 1(fa)]{sun2024understanding}). In our Experiment 4 of Section~\ref{sec:experiments} we show that the increase of the generalization error with the degree of label heterogeneity observed in ~\cite[Fig. 1(a)]{sun2024understanding}  (and which seemingly contrasts with the findings of this paper) is actually attributed to a variable number of communication rounds beteen the clients and the server at distinct heterogeneity levels that is allowed therein. However, if for a fair\footnote{For a given fixed level of data heterogeneity, the generalization error was shown in~\cite{sefidgaran2024lessons} to increase with the number of communication rounds $R$ between clients and server.} comparison~\cite{sefidgaran2024lessons} one sets a same $R$ at distinct label heterogeneity levels in that experiment, then one recovers a behavior that is similar to the one observed in this paper, i.e., the generalization error decreases with the label heterogeneity across clients. Moreover, our approach to studying the generalization error and the resulting bounds, which apply to a one-round scenario, are different, being rate-distortion theoretic. In~\cite{wang2025generalization}, the authors develop an information-theoretic analysis via the CMI framework to study the effect of participating and non-participating clients in FL on the generalization error. They introduce a super-client construction to account for the problem of client participation, in addition to the super-sample CMI construction traditionally used to account for the problem of training data sample membership~\cite{steinke2020reasoning}; and they use them to establish various CMI-type bounds. However, most of the bounds of~\cite{wang2025generalization} do \textit{not} account explicitly for the distributed nature of the FL learning problem and model aggregation strategies~\footnote{Similar to in~\cite{yagli2020information} the learning problem is modeled in~\cite{wang2025generalization} as a (possibly) stochastic map  $\mathcal{A}\: : \mathcal{Z}^{nK} \rightarrow \mathcal{W}$, where $K$ is the number of clients and $n$ is the size of local datasets; and, as such, it does not account explicitly for the structure of the FL problem.}. Perhaps most close to this work is the setup of~\cite[Section 6]{wang2025generalization} in which the authors study a one-shot FL problem with local models averaging at the server, i.e., one-shot FedAvg. For this setup, the authors derive CMI-type bounds under some rather restrictive assumption on the loss function, namely losses that can be experssed as a Bregman divergence or ones that are strongly convex and smooth.

Comparatively, the effect of the data being non independent and/or identically distributed (non-i.i.d.) across clients on the performance of FedAvg has been studied more from the viewpoint of convergence rate. That is, how the convergence rate of the algorithm degrades in the presence of non-i.i.d. data across clients. To this end, the FedProx of~\cite{li2020federated2} introduces a regularization term $\|\bar{W}-W_k\|$ that helps stabilize updates and reduce divergence among local models. The SCAFFOLD of~\cite{karimireddy2020scaffold} employs variance reduction techniques to improve convergence rates; and the FedNova of~\cite{wang2020tackling} adresses the problem of objective inconsistencies caused by naive weighted aggregation of clients' individual models by means of  a normalized averaging scheme. Also, the FedMDMI of~\cite{zhang2024improving} uses model-data mutual information regularization to reduce the induced generalization error. Also relevant is the work of~\cite{JMLR:v24:21-0224} which studies the effect of data heterogeneity across clients on the excess risks of personalized federated learning and shows that, for smooth and  strongly convex loss functions, an \textit{approximate} alternative between FedAvg and pure local training exists, in the sense that FedAvg is  minimax-rate optimal when the data heterogeneity is small whereas pure local training is minimax-rate optimal when data heterogeneity is big. The interested reader may also refer to, e.g.~\cite{woodworth2020local,woodworth2020minibatch,wang2021cooperative}.

\subsection{Notation} Upper case letters denote random variables,
e.g., \(X\); lower case letters denote realizations of random variables, e.g., \(x\); and calligraphic letters denote
sets, e.g., \(\mathcal{X}\). The probability distribution of a random variable \(X\) is denoted as \(P_X\)  and its support set as \(\supp(P_X)\). For probability distributions \(P\) and \(Q\) defined over a common measurable space \(\mathcal{X}\) such that \(Q\) is absolutely continuous concerning \(P\) (i.e., \(Q \ll P\)), the relative entropy between \(Q\) and \(P\), also called the Kullback-Leibler (KL) divergence, is given by \(D_{KL}(Q \| P) \coloneqq \mathbb{E}_Q\Big[\log \Big( \frac{dQ}{dP} \Big)\Big]\). If \(Q\) is not absolutely continuous with respect to \(P\), the Radon-Nikodym derivative \(\frac{dQ}{dP}\)  is undefined and we set \(D_{KL}(Q \| P) = \infty\). The Shannon mutual information (MI) between two random variables \(X\) and \(Y\)
with joint distribution \(P_{X,Y}\) and marginals \(P_X\) and \(P_Y\) is given by 
\[
I(X;Y) = D_{KL}(P_{X,Y} \parallel P_X P_Y).
\]
Conditional mutual information, given a possibly correlated variable \(Z\), is denoted as \(I(X;Y|Z)\) and given by
\begin{equation}
I(X;Y|Z)=\mathbb{E}_{P_Z}\left[D_{KL}\left(P_{X,Y|Z} \parallel P_{X|Z} P_{Y|Z}\right)\right].\nonumber
\end{equation}
 For \(n \in \mathbb{N}\), the notation \([n]\) denotes the set \(\{1, \dots, n\}\). Also, \(\mathbbm{1}_{\{ \cdot \}}\) designates the indicator function. Finally, a set of random variables \(\{X_1, \dots, X_n\}\) is sometimes abbreviated as \(X_{[n]}\). Finally, for \((a,b) \in \mathbb{R}^2\), \([a,b]^+ = \text{max}(a,b)\).

\section{System model and preliminaries}\label{sec:problem_statement}
Consider the distributed learning system shown in Fig.~\ref{fig-system-model}. As mentioned, there are \(K\) clients; each with a training dataset \(S_k = \{Z_{k,1}, \dots, Z_{k,n}\} \in \mathcal{Z}^n \) of size \(n\), whose samples \(\{Z_{k,1}, \dots, Z_{k,n}\}\) are generated independently from each other and other clients' training samples according to some probability distribution \(\mu_k\). The probability distributions \(\{\mu_k\}_{k=1}^K\) are allowed to vary across clients, and we refer to such a setting as being \textit{heterogeneous}. This is opposed to \textit{homogeneous} data setting (across clients) in which \(\mu_k = \mu\) for all \(k=1,\hdots,K\). For example, for classification tasks we set \(Z_{k,i} = (X_{k,i}, Y_{k,i})\), where \(X_{k,i}\) denotes the feature sample and \(Y_{k,i}\) is the associated label. Client \(k\) applies a possibly  stochastic learning algorithm \(A_k: \mathcal{Z}^n_k \rightarrow \mathcal{W}_k\). This induces a conditional distribution \(P_{W_k | S_k}\), which together with \(\mu_k\) induce the joint dataset-hypothesis distribution $P_{W_k, S_k} = \mu_k^{\otimes n} P_{W_k | S_k}$. The server receives the hypotheses \( (W_1,\hdots,W_K)\)  and picks the hypothesis \(\overline{W}\) as the arithmetic average given by~\eqref{model-aggregation-at-server}. We use a loss function \(\ell : \mathcal{Z} \times \mathcal{W} \to [0,1]\) to evaluate the quality of the prediction. For a given value \(\overline{w}\) of the aggregated model, how well it performs on the training dataset of Client \(k\) is evaluated using the empirical risk
\begin{equation}
    \hat{\mathcal{L}}_k(S_k, \overline{w}) = \frac{1}{n} \sum\nolimits_{i=1}^{n} \ell(Z_{k,i}, \overline{w});
    \nonumber
\end{equation}
and how well it does on test data distributed according to \(\mu_k\) is evaluated as
\begin{equation}
\mathcal{L}_k(\overline{w}) = \mathbb{E}_{Z \sim \mu_k} [\ell(Z_k, \overline{w})].
\nonumber
\end{equation}
Setting
\begin{equation}
{\gen}_k(\overline{w}) = \mathcal{L}_k(\overline{w}) - \hat{\mathcal{L}}_k(S_k, \overline{w}),
\label{local-generalization-error-client-k}
\end{equation}
in this paper, for the dataset \(S_{[K]}=(S_1,\hdots,S_K)\) we measure the generalization error of aggregated hypothesis \(\overline{W}\) as the average (over clients)
\begin{equation}
\gen(S_{[K]},\overline{W})=\frac{1}{K}\sum\nolimits_{k=1}^K \gen_{k}(\overline{W}).
\label{definition-generalization-error}
\end{equation}
As we already mentioned in the Introduction section, we study the effect of the discrepancy between the data distributions on the generalization error~\eqref{definition-generalization-error} of the aggregated model~\eqref{model-aggregation-at-server}. Then, we apply the found results, to an example D-SVM,  to gain insights onto which of two training procedures (among heterogeneous data across clients or homogeneous data across clients) yields an aggregated model \(\overline{W}=(W_1+\hdots+W_K)/K\) that generalizes better to unseen data during test time -- for fair comparison, the test samples are generated from the same distribution for both settings.

For convenience, we define the following symmetry property, which will be instrumental throughout.

\begin{definition}[Symmetric Priors]\label{def:symmetric}
Let $\sigma:[2n] \to [2n]$ be an arbitrary permutation of the set \(\{1,\hdots,2n\}\). For a generic vector \(U^{2n}=(U_1,U_2,\hdots,U_{2n})\), we set $\sigma(U^{2n}) \coloneqq \left(U_{\sigma(1)}, \cdots, U_{\sigma(2n)}\right)$.
\begin{itemize}[leftmargin = 0.4 cm]
  \item\textbf{{Type-I symmetry}}: Define Type-I permutations as the set of permutations $\sigma:[2n]\to [2n]$ with the property that $\{\sigma(i),\sigma(i+n)\}=\{i,i+n\}$ for all \(i=1,\hdots,n\). A conditional distribution (prior) $Q(W\lvert V^{2n})$ is said to possess type-I symmetry if $Q(W|\sigma(V^{2n})$ is invariant under any type-I permutation \(\sigma :[2n]\to [2n]\).
  
 \item\textbf{Type-II symmetry}: The conditional prior $\mathbf{Q}(W|V^{2n})$ is said to satisfy Type-II symmetry if it is invariant under any arbitrary permutation $\sigma:[2n]\to [2n]$.
\end{itemize}
\end{definition}

\section{CMI-type generalization bounds} \label{sec-CMI-type-gneralization-bopunds} In our \textit{distributed} CMI framework, for every client \(k\) we generate a \textit{supersample} \((Z_{k,1}\hdots,Z_{k,n},Z'_{k,1},\hdots,Z'_{k,n}) \in \mathcal{Z}^{2n}_k\) consisting of \(n\) training samples \(S_k=(Z_{k,1}\hdots,Z_{k,n})\) as well as \(n\) \textit{ghost} samples \(S'_k=(Z'_{k,1}\hdots,Z'_{k,n})\), all drawn i.i.d. from \(\mu_k\). For the analysis, we will need to define, for every \(k \in [K]\),  a \textit{membership vector} \(\mathbf{J}_k \) that consists of \(n\) Bernoulli-1/2 random variables that are independent of each other and the supersample \((S_k,S'_k)\). Specifically, let \(\mathbf{J}_k = (J_{k,1}, \ldots, J_{k,n})\), where for \(i \in [n] \) \(J_{k,i}\) is a Bernoulli-1/2 random variable defined over the set \(\{i,i+n\}\) that is independent of everything else. Also , let  $J_{k,i}^c \in \{i, i+n\}$ be the random variable complement of \(J_{k,i}\), i.e., $J_{k,i}^c = i+n$ if $J_{k,i} = i$ and $J_{k,i}^c = i$ if $J_{k,i} = (i+n)$. Define for every \(i \in [n]\) the random variables $\mathfrak{Z}_{J_{k,i}}$ and $\mathfrak{Z}_{J_{k,i}^c}$ as $\mathfrak{Z}_{J_{k,i}} = Z_{k,i}$ and $\mathfrak{Z}_{J_{k,i}^c} = Z_{k,i}'$. Observe that the vector $\mathfrak{Z}^{2n}_{k} = (\mathfrak{Z}_{1}, \ldots, \mathfrak{Z}_{2n})$ is a \(\mathbf{J}_k\)-dependent re-arrangement of the samples of the training and ghost datasets \(S_k\) and \(S'_k\) in a manner that, without knowledge of the value of \(\mathbf{J}_k\) every element of that re-arrangement has equal likelihood to be picked from \(S_k\) or \(S'_k\). Occasionally, we will also need the size-\(n\) sub-vectors of the vector \(\mathfrak{Z}^{2n}_{k}\) with elements determined by  \(\mathbf{J}_k\) or  \(\mathbf{J}^c_k\), i.e., $\mathfrak{Z}^{2n}_{\mathbf{J}_k} = (\mathfrak{Z}_{{J}_{k,1}}, \ldots, \mathfrak{Z}_{{J}_{k,n}})$ and $\mathfrak{Z}^{2n}_{\mathbf{J}^c_k} = (\mathfrak{Z}_{{J}^c_{k,1}}, \ldots, \mathfrak{Z}_{{J}^c_{k,n}})$.

\subsection{In-expectation bound}

The next theorem, the proof of which can be found in Appendix~\ref{proof:inexp:cmi:general}, states a bound on the generalization error~\eqref{definition-generalization-error} that holds in expectation over all datasets and hypotheses.

\vspace{0.4cm}

\begin{theorem}\label{CMI:FL:TH}
     Let , for $k \in [K]$, $\mathcal{Q}_k$ denote the set of type-I symmetric conditional priors on $W_k$ given 
       $(S_k, S'_k)$. Then, 
\begin{align}
    \mathbb{E}_{S_{[K]}, \overline{W}} \left[ \gen(S_{[K]}, \overline{W}) \right] \leq \sqrt{\frac{2E}{n}},\nonumber
    \end{align}
    where
    \begin{align}
        E &= \frac{1}{K} \sum_{k=1}^K \inf_{{Q}_k \in \mathcal{Q}_k} \mathbb{E}_{S_k, S_k'} \left[ D_{KL}\left( P_{W_k \mid S_k, S_k'} \parallel {Q}_{k}\right) \right]\nonumber\\
        &= \frac{1}{K} \sum_{k=1}^K I(W_k; \mathbf{J}_k \mid \mathfrak{Z}_k^{2n}),\nonumber
    \end{align}
    with the mutual information computed with respect to
    \begin{align}
    P_{\mathbf{J}_k, W_k, S_k, S_k'} = \text{Bern} \left( \frac{1}{2} \right)^{\otimes n} \otimes \mu_k^{\otimes 2n} \otimes P_{W_k \mid S_k}.\nonumber
    \end{align} 
\end{theorem}

 The result of Theorem~\ref{CMI:FL:TH} can be seen as an extension, to the distributed learning setting with arbitrary number of clients, of that of~\cite{steinke2020reasoning} which introduced the concept of CMI and derived a bound on the average generalization error in the centralized learning setting, i.e., \(K=1\). Alternatively, Theorem~\ref{CMI:FL:TH} also extends a bound of \cite{sefidgaran2022rate,yagli2020information} and (a special case~\footnote{The bound of~\cite{sefidgaran2024lessons} accounts for multiple rounds communications between the clients and the server.} of)  a bound of \cite{sefidgaran2024lessons} developed for Federated Learning and expressed therein in terms of mutual information. Comparatively, a clear advantage of our CMI bound of Theorem~\ref{CMI:FL:TH} is that it is inherently bounded, while bounds based on mutual information (such as those of \cite{sefidgaran2022rate,sefidgaran2024lessons}) are possibly vacuous and unbounded in certain cases. 

As it will become clearer from the rest of this paper, a suitable generalization of Theorem~\ref{CMI:FL:TH} (that we call \textit{lossy} bound) will be used to study the effect of data heterogeneity across clients in the case of distributed support vector machines. In that case, our bounds will have closed-form expressions with explicit dependence on \(n,K\) and parameters of the distributions \(\mu_1,\hdots,\mu_K\).

\subsection{Tail bound}

In this section, we provide a tail bound on the generalization error of distributed learning algorithms, using the CMI framework of \cite{steinke2020reasoning}.
\begin{theorem}\label{tailbound:cmi:general}
Let , for every $k \in [K]$, $\mathcal{Q}_k$ denote the set of type-I symmetric conditional priors on $W_k$ given $(S_k,S_k')$. Then, for every $\delta > 0$  we have that with probability at least  $(1 - \delta)$ under \(S_{[K]}\sim\prod_{k=1}^K \mu_k^{\otimes n}\), the generalization error~\eqref{definition-generalization-error} is bounded from the above by
    \begin{align}
   \inf  \sqrt{\frac{E+ K \log \left( \sqrt{2n} \right) + \log \left( \frac{1}{\delta} \right)}{(2n-1)K/4}},\nonumber
    \end{align}
    where
    \begin{align}
    E = \sum_{k \in [K]} &\mathbb{E}_{S'_{[K]}} \left[ D_{KL} \left( P_{W_k|S_k,S'_k}\parallel{Q}_k \right) \right],\nonumber
    \end{align}
    and the infimum is over conditional priors \(\{{Q}_k \in \mathcal{Q}_k \}_{k=1}^K\). The proof of this result can be found in Section \ref{Tail:proof:CMI}.
\end{theorem}

\subsection{Lossy bound}

In this section, we use a connection between the theory of generalization of statistical learning algorithms and information theoretic rate-distortion theory that was introduced in~\cite{Sefidgaran2022}, and subsequently used and elaborated on in~\cite{sefidgaran2022rate}, to tighten the bound of Theorem~\ref{CMI:FL:TH}. The proof is given in Appendix~\ref{Lossy:general:CMI:1}.

\begin{theorem}\label{Lossy:inexpectation:general}
Let \(\epsilon \in \mathbb{R}\) and let for every \(k \in [K]\), $\hat{\overline{W}}_k$ be a (compressed) hypothesis generated according to some conditional $P_{\hat{\overline{W}}_k | S_k, W_{[K]\setminus{k}}}$ such that 
\begin{equation}
\mathbb{E}\left[\gen_k(\overline{W})-\gen_k(\hat{\overline{{W}}}_k)\right]\leq\epsilon.
\label{distortion-constraint}
\end{equation}
Then, we have 
\begin{equation}
\mathbb{E}\left[\gen(S_{[K]},\overline{W})\right]\leq\sqrt{\frac{2\sum_{k\in[K]}R_{\mathcal{D}_k}(\epsilon)}{nK}}+\epsilon,
\label{lossy-bound-expected-generalization-error}
\end{equation}
where 
   \begin{align}
       R_{\mathcal{D}_k}(\epsilon):=\inf I\Big(\hat{\overline{W}}_k;\mathbf{J}_k\lvert \mathfrak{Z}_{k}^{2n},W_{[K]\setminus{k}}\Big), 
       \label{rate-distortion-term}
    \end{align}
the infimum is over all conditional distributions \(P_{\hat{\overline{W}}_k\lvert \mathfrak{Z}_{k}^{2n},W_{[K]\setminus{k}},\mathbf{J}_k} = P_{\hat{\overline{W}}_k\lvert \mathfrak{Z}_{\mathbf{J}_k}^{2n},W_{[K]\setminus{k}}} \) and the mutual information is calculated according to the joint distribution \(P_{\mathfrak{Z}_{k}^{2n},W_{[K]\setminus{k}},\mathbf{J}_k} \times P_{\hat{\overline{W}}_k\lvert \mathfrak{Z}_{k}^{2n},W_{[K]\setminus{k}},\mathbf{J}_k}\).

\end{theorem}

Few remarks are in order. First, it is not difficult to see that setting $\epsilon=0$ in Theorem~\ref{Lossy:inexpectation:general} one recovers Theorem~\ref{CMI:FL:TH}. By allowing non-zero values of $\epsilon \geq 0$, one possibly tightens the result of Theorem~\ref{CMI:FL:TH}. The advantage of the \textit{lossy compression}, i.e., \(\epsilon > 0\), can be seen as follows. Consider the specific choice of \(\hat{\overline{W}}_k\) given by
\begin{equation}
\hat{\overline{W}}_k=\Big(\hat{W}_k + \sum_{i \: \in \: [K]\setminus{k}} W_i\Big)/K\nonumber
\end{equation}
such that~\eqref{distortion-constraint} is satisfied. This choice generally does not achieve the infimum on the RHS of~\eqref{rate-distortion-term} and so is not optimal in general. Also, with such a choice the RHS of ~\eqref{rate-distortion-term} reduces to $I(\hat{W}_k;\mathbf{J}_k\lvert \mathfrak{Z}_{k}^{2n})$. On one side, relaxing the constraint that \(P_{\hat{W}_k|S_k}\) should induce a generalization error that equals \(\text{gen}(S_{k},\overline{W})\); and, instead, only requiring that that constraint be satisfied approximately, i.e.,~\eqref{rate-distortion-term} with \(\epsilon > 0\), leads to a possibly smaller rate (since the set of distributions over which the infimum is taken is bigger). This, however, comes at the expense of an additional (distortion) term in the bound (the additive constant \(\epsilon\) on the RHS of~\eqref{lossy-bound-expected-generalization-error}). In certain cases, the net effect can be positive as already exemplified in the centralized learning setting in~\cite{Sefidgaran2022,sefidgaran2024data}.  


\section{Improved generalization bounds in terms of Jensen-Shannon divergence} \label{sec:hd}
In this section, we develop another type of generalization bound, which improves over the CMI-type generalization bounds in some cases. These bounds are expressed in terms of the Jensen-Shannon divergence. Let $h_D\colon [0,1]\times [0,1] \to [0,2]$ be the function defined as, for \((x_1,x_2) \in [0,1]^2\), 
\begin{equation}\nonumber
h_D(x_1,x_2) := 2h_b\Big(\frac{x_1 + x_2}{2}\Big) - h_b(x_1) - h_b(x_2),
\end{equation}
with \( h_b(x)\) denoting the binary entropy of parameter \(x \in [0,1]\), i.e., \(h_b(x) := -x \log x - (1 - x) \log (1 - x) \). It is easy to see that $h_D(x_1,x_2)$ equals two times the Jensen-Shannon Divergence between Bernoulli distributions with parameters $x_1$ and $x_2$. The reader is referred to Lemma~\ref{h_d:lemma} for further results on the properties of this function.

Next, for \(c \in [0,1]\) let $h_D^{-1}(\cdot | c) \colon [0,2] \to [0,1]$ denote the function \textit{inverse} of \(h_D(\cdot, c)\), defined as
\begin{align}\nonumber
        h_D^{-1}(y|c) = \sup \left\{x \in[0,1]\colon h_D(x,c) \leq y\right\}.
\end{align}

The function $h_D$ has several interesting properties, as shown in the following lemma.
\begin{lemma}\label{h_d:lemma}
For any $x_1,x_2 \in [0,1]$, $y\in[0,2]$,  
    \begin{itemize}
        \item[A)] $h_D(x_1,x_2)\geq(x_1-x_2)^2$,
        \item[B)] $h_D(x_1,0)\geq x_1$,
        \item[C)] $h_D(x_1,x_2)$ is increasing with respect to $x_1$ in the range $[x_2,1]$,
        \item[D)] $h_D(x_1,x_2)$ is convex with respect to both inputs,
         \item[E)] $h^{-1}_D(y|0)\leq y$,
        \item[F)] $h_D^{-1}(y|x_1)\leq x_1+\sqrt{y}$,
        \item[G)]  for $a,b\in[0,1/2]$, the function $h_D(a+x,b+x)$ is decreasing in the range $$x\in\left[0,\frac{1}{2}-\max(a,b)\right].$$ 
    \end{itemize}
\end{lemma}

The proof of items (A-D) and (E-G) can be found in {\cite[Lemma 1]{sefidgaran2024minimum}} and  Appendix~\ref{Proof:Lemma:auxiliary:h_d:property}, respectively.

Intuitively, the results established using the $h_D$ function are achieved by considering a suitable arrangement of the elements of \((S_k, S'_k)\) which is different from the arrangement considered in CMI-type of bounds. Specifically, let \(\mathbf{T}_k\sim\unif(2n)\) where indicates that $\mathbf{T}_k$ is a subset of indices of the set \(\{1,\hdots, 2n\}\) of size \(n\), chosen uniformly with probability \(1/{\binom{2n}{n}}\). Furthermore, set \(\mathbf{T}^c_k\) be the set complement in \(\{1,\hdots,2n\}\). That is, \(\mathbf{T}^c_k = \{1,\hdots, 2n\} \setminus \mathbf{T}_k\). We set \(S_k = \mathfrak{Z}^{2n}_{\mathbf{T}_k}\) and \(S'_k = \mathfrak{Z}^{2n}_{\mathbf{T}^c_k}\). Now, we are ready to state our generalization bounds.

\subsection{In-expectation bound} \label{sec:hd_expectation}
We start with the lossless generalization bound, proved in Appendix~\ref{h_D:inexpectation:general:proof}.
  \begin{theorem}\label{h_D:inexpectation:general}
  Let , for $k \in [K]$, $\mathcal{Q}_k$ denote the set of type-II symmetric conditional priors on $W_k$ given 
       $(S_k, S'_k)$. Then, for $n\geq 10$,  

         \begin{align}
     nh_{D}\left(\mathbb{E}_{\overline{W}}\big[\mathcal{L}(\overline{W})\big],\mathbb{E}_{S_{[K]},\overline{W}}\big[{\hat{\mathcal{L}}(S_{[K]},\overline{W})}\big]\right)
        &\leq\frac{1}{K}\sum_{k=1}^{K}\inf_{Q_k\in \mathcal{Q}_k}\mathbb{E}_{S_k,S^{\prime}_k}\left[D_{KL}(P_k\parallel Q_k)\right]+\log n\label{general:h_d:1}\\
         &=\frac{1}{K}\sum_{k=1}^{K}I(\mathbf{T}_k;W_k\lvert \mathfrak{Z}_k^{2n})+\log n,\label{general:h_d:2}
         \end{align}
              with the mutual information computed with respect to
    \begin{equation}
P_{\mathbf{T}_k,W_k,S_k,S^{\prime}_k}=P_{\mathbf{T}_k}\otimes\mu^{\otimes 2n}_k\otimes P_{W_k\lvert S_k}.\nonumber
    \end{equation}
    \end{theorem}

  The proof consists of two parts. In the first part, similar to in
 the proof of Theorem~\ref{CMI:FL:TH}, we establish~\eqref{general:h_d:2}. In the second part, we derive an upper bound on  $n h_D(\cdot,\cdot)$ using Jensen's inequality, since the function $h_D(\cdot,\cdot)$ is convex. The rest of the proof follows by an application of Donsker-Varadhan variational lemma to get a bound in terms of the KL-divergence of~\eqref{general:h_d:1} and a residual term that can be bounded by $\log n$ using Lemma~\ref{simp:expect:loss:1} that follows.

 \begin{lemma}\label{simp:expect:loss:1}
Let $\mathbf{T}$ be a subset of length $n$ randomly chosen from $[2n]$ with distribution $\unif(2n)$. Let $\mathbf{T}^c$ be the complement of $\mathbf{T}$ with respect to $[2n]$, i.e., $\mathbf{T}^c = [2n] \setminus \mathbf{T}$. Then for any set of $\ell_i\in[0,1]$, $i\in[2n]$, we have
\begin{align}
    \mathbb{E}_{\mathbf{T} \sim \unif(2n)}\left[\exp\left(n h_D\left(\frac{1}{n}\sum_{i \in \mathbf{T}} \ell_i, \frac{1}{n}\sum_{i' \in \mathbf{T}^c} \ell_{i'}\right)\right)\right] \leq n. \nonumber
\end{align}
The proof of this lemma appears in Appendix~\ref{Proof:auxiliary:lemma:h_d}.
\end{lemma}

Next, we state the lossy version of this result, whose proof is deferred to Appendix~\ref{Lossy:h_d:inverse}. 

\begin{theorem}\label{h_d:ingeneral} Let $\epsilon \in \mathbb{R}$ and assume that, for every \(k \in [K]\), $\hat{\overline{W}}_k$ is a (compressed) hypothesis generated according to some conditional $P_{\hat{\overline{W}}_k|S_k,W_{[K]\setminus{k}}}$ that satisfies
\begin{equation}
\frac{1}{K}\sum_{k\in[K]}\mathbb{E}\left[\gen_k(\overline{W})-\gen_k(\hat{\overline{{W}}}_k)\right]\leq\epsilon,
\nonumber
\end{equation}
where the expectation is with respect to $P_{S_{[K]},W_{[K]},\overline{W}}P_{\hat{\overline{W}}_k|S_{k},W_{[K]\setminus k}}$. Then, $\mathbb{E}\left[\gen(S_{[K]},\overline{W})\right]$ is upper bounded by
\begin{equation}
   h^{-1}_D\bigg(\frac{1}{nK}\sum\limits_{k\in[K]}(\tilde{E}_k+\log n)\Big| \hat{\mathcal{L}}_{[K]}\bigg)-\hat{\mathcal{L}}_{[K]}+\epsilon,\nonumber
    \end{equation}
where $\hat{\mathcal{L}}_{[K]}=\frac{1}{K}\sum\nolimits_{k\in[K]}\mathbb{E}\big[\hat{\mathcal{L}}(S_k,\hat{\overline{W}}_k)\big]$ and 
    \begin{align}    
    \tilde{E}_k=& \inf_{Q_k\in \mathcal{Q}_k}\mathbb{E}_{S_k,S^{\prime}_k}\left[D_{KL}\Big(P_{\hat{\overline{W}}_k|S_{k},W_{[K]\setminus k}}\parallel Q_k\Big)\right] \nonumber \\
    = &{I(\mathbf{T}_k;\hat{\overline{W}}_k\lvert\mathfrak{Z}_{k}^{2n},W_{[K]\setminus{k}})}.\nonumber
\end{align}
Here \(\mathcal{Q}_k\) denotes the set of type-II symmetric conditional priors (Definition~\ref{def:symmetric}) of \(\overline{W}_k\) given \((S_k,S_k',W_{[K]\setminus k})\) and the mutual information is taken with respect to $\mu_k^{\otimes 2n} \otimes P_{T_k} \otimes P_{W_{[K]\setminus k}|\mathfrak{Z}^{2n}_{\mathbf{T}_k}} Q_k(\hat{\overline{W}}_k|\mathfrak{Z}_k^{2n},W_{[K]\setminus k})$. \qed 
\end{theorem}

For the lossless case, i.e. when \(\epsilon = 0\) and $P_{\hat{\overline{W}}_k | S_k, W_{[K]\setminus{k}}}\equiv P_{{\overline{W}} | S_k, W_{[K]\setminus{k}}}$, the bound simplifies as
\begin{equation}
 h^{-1}_D\bigg(\frac{1}{nK}\sum\limits_{k\in[K]}(I(\mathbf{T}_k;W_k\lvert \mathfrak{Z}_k^{2n})+\log n)\Big| \hat{\mathcal{L}}_{[K]}\bigg)-\hat{\mathcal{L}}_{[K]}, \label{eq:lossless_hd}
\end{equation}
where $\hat{\mathcal{L}}_{[K]}=\mathbb{E}_{S_{[K]},\overline{W}}\big[{\hat{\mathcal{L}}(S_{[K]},\overline{W})}\big]$. Furthermore, since by Lemma~\ref{h_d:lemma}, we have $h_D^{-1}(y|c)\leq c+\sqrt{y}$ and  $h_D^{-1}(y|0) \leq y$ for any $y,c\geq 0$, \eqref{eq:lossless_hd} results in generalization bounds $\sqrt{\frac{1}{nK}\sum_{k\in[K]}({C}_k+\log n)}$ and $\frac{1}{nK}\sum_{k\in[K]}({C}_k +\log n)$ where ${C}_k=I(\mathbf{T}_k;W_k\lvert \mathfrak{Z}_k^{2n})$, for the non-realizable and realizable setups, respectively.

In particular, as it will be shown in the sections that follow, for Distributed Support Vector Machines, Theorem~\ref{h_d:ingeneral} gives a generalization bound of order $\mathcal{O}\left(\frac{\log(K)\log(nK)}{nK^2}+\frac{\log(n)}{n}\right)$ when the empirical loss is sufficiently small. When $n>K^2$, this bound improves over the generalization bound of order $\mathcal{O}\left(\sqrt{\frac{\log(K)\log(nK)}{nK^2}+\frac{\log\left(\left[1,n/K\right]^{+}\right)}{n}}\right)$ that is established using Theorem~\ref{Lossy:inexpectation:general}.

It is worth noting that, even for the specific case $K=1$, the result of Theorem~\ref{h_d:ingeneral} possibly improves upon the classical CMI result of \cite{steinke2020reasoning} (i.e. Theorem~\ref{CMI:FL:TH} with $K=1$), for small values of empirical risk.

\subsection{Tail bound} \label{sec:hd_tail}
Here, we establish a tail bound in terms of the $h_D$ function.
\begin{theorem}\label{h_d:tail:general} 
Let , for every $k \in [K]$, $\mathcal{Q}_k$ denote the set of type-II symmetric conditional priors on $W_k$ given $(S_k,S_k')$. Then, for every $\delta > 0$ with probability at least  $(1 - \delta)$ under \((S'_{[K]},S_{[K]})\sim\prod_{k=1}^K \mu_k^{\otimes 2n}\), $$nh_{D}\Big(\mathbb{E}_{\overline{W}\lvert S_{[K]}}\big[{\hat{\mathcal{L}}(S^{\prime}_{[K]},\overline{W})}\big],\mathbb{E}_{\overline{W}\lvert S_{[K]}}\big[\hat{\mathcal{L}}(S_{[K]},\overline{W})\big]\Big)$$ can be upper bounded by
    \begin{align}
      \inf_{Q_k,\cdots,Q_k}\frac{1}{K}\sum_{k=1}^{K}D_{KL}\left(P_{W_k\lvert S_k,S^{\prime}_k}\parallel{Q}_{k}\right)+\log (n/\delta),\nonumber
    \end{align}
    for $n\geq 10$. 
    \end{theorem}
The detailed proof can be found in Appendix~\ref{Tail:h_d}.


\vspace{-0.2cm}

  \section{Effect of data heterogeneity on generalization: A warm up}\label{sec:het_svm}

\begin{figure}
    \centering
    \includegraphics[width=0.6\linewidth]{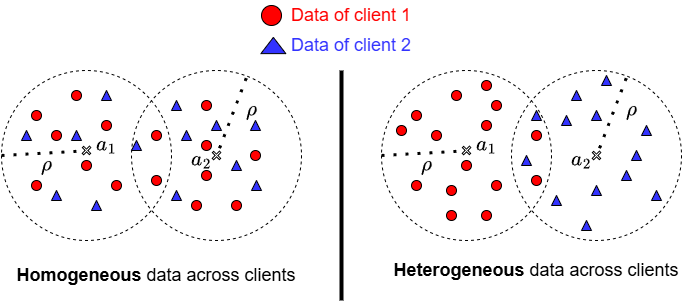}
     \caption{Illustration of (training) data generation for an example D-SVM problem with \(K=2\) clients.}
    \label{Data:distribution:Het:K=2}
\end{figure}

For convenience, we start with $K=2$. Consider an instance of the system of Fig.~\ref{fig-system-model} used for \textit{distributed} binary classification with two clients. In this case, \( \mathcal{Z}_1 = \mathcal{X}_1 \times \mathcal{Y} \) and \( \mathcal{Z}_2 = \mathcal{X}_2 \times \mathcal{Y}\), with \( \mathcal{Y} = \{-1, +1\} \). In accordance with the general setup of Section~\ref{sec:problem_statement}, Client 1 has \(n\) training samples $S_1=(Z_{1,1},\hdots,Z_{1,n})$ and Client 2 has \(n\) training samples $S_2=(Z_{2,1},\hdots,Z_{2,n})$. Both clients use Support Vector Machines (SVM) to obtain respective models $W_1$ and $W_2$; and the aggregated model is \(\overline{W} {=} (W_1+W_2)/2\).

We investigate the question of the effect of the data heterogeneity across the two clients on the generalization error of the model \(\overline{W}\). To this end, we compare the performance of \(\overline{W}\) (from a generalization error perspective) in the following two settings, depicted pictorially in Fig.~\ref{Data:distribution:Het:K=2}.
\begin{itemize}[leftmargin=0.3 cm]
\item \textit{Heterogeneous data setting:} In this case, for \(k=1,2\) the training samples \(\{(X_{k,j},Y_{k,j})\}_{j=1}^n\) of Client \(k\) are drawn independently at random from an arbitrary distribution \(\mu_k^{\text{Het}}\) which satisfies
\begin{equation}
\mathbb{P}\big(\parallel X_{k,j}-a_k\parallel \leq \rho \big) = 1,\quad \forall\: j\in [n].
\label{data-distribution-heterogeneous-setting-two-clients}
\end{equation}
For example, the data of Client 1 drawn independently at random from uniform distribution over a \(d\)-dimensional ball with center \(a_1\) and radius \(\rho\), for some \(a_1 \in \mathbb{R}^{d}\) and \(\rho \in \mathbb{R}^+\); and, similarly, the training samples of Client 2 drawn independently at random from the uniform distribution over a \(d\)-dimensional ball with center \(a_2\) and radius \(\rho\). That is,  \(X_{k,j} \sim \text{Unif}(\mathcal B(a_k,\rho))\).

\item \textit{Homogeneous data setting:} In this case, both clients have their training samples picked independently at random from the same distribution \(\mu^{\text{Hom}}=(\mu_1^{\text{Het}}+\mu_2^{\text{Het}})/2\). In particular, \(\mu^{\text{Hom}}\) satisfies, for \( k = 1, 2 \) and every \( j \in [n] \), 
 \begin{equation}
 \mathbb{P}\Big(\parallel X_{k,j}-a_1\parallel \leq\rho \:\: \text{or} \:\: \parallel X_{k,j}-a_2\parallel \leq\rho \Big) = 1.
 \label{data-distribution-homogeneous-setting-two-clients}
\end{equation}

\end{itemize}
For both settings, we measure the generalization error as given by~\eqref{definition-generalization-error}. For~\eqref{local-generalization-error-client-k}, we use the \(0\)-\(1\) loss function \(\ell_{0}(z, w) = \mathbbm{1}_{\big\{ y f(x, w) < 0 \big\}}\), where the sign of \(f(x,w)\) is the label prediction by hypothesis \(w\) and \(\mathbbm{1}\) is the indicator function, for the evaluation of the population risk; and, as it is common in related literature \cite{gronlund2020}, we use the \(0\)-\(1\) loss function with margin \(\theta\), for some \(\theta \in \mathbb{R}^+\), defined as \(\ell_{\theta}(z, w) = \mathbbm{1}_{\big\{ y f(x, w) < \theta \big\}}\),  for the evaluation of the empirical risk. That is, 
\begin{align}
    \gen_{\theta}(\overline{w}) = \mathcal{L}(\overline{w}) - \hat{\mathcal{L}}_{\theta}(S, \overline{w}).\nonumber
\end{align}

\subsection{Generalization bounds for heterogeneous and homogeneous data settings}
The next theorem states bounds on the expected generalization error of the distributed SVM classification problem with $K=2$ clients for both heterogeneous and homogeneous data settings.

\vspace{0.1cm}

\begin{theorem}\label{SVM:K=2:het-hom}
Let $\theta \in (0,1]$. The expected margin generalization error \(\mathbb{E}[\text{gen}_{\theta}(S_{[2]},\overline{W})]\) is bounded by
    \begin{equation*}
        \mathcal{O}\left(\sqrt{\frac{(\frac{\rho}{\theta})^2\log\left( [3,\frac{\theta}{\rho}]^{+}\right) \log n+\frac{1}{2}\log\left(\frac{n^2\parallel a_1\parallel \parallel a_2\parallel}{\theta^2}\right)}{n}}\right).
        \label{SVM:K=2:1}
        \end{equation*}
in the heterogeneous data setting~\eqref{data-distribution-heterogeneous-setting-two-clients}; and in the homogeneous data setting~\eqref{data-distribution-homogeneous-setting-two-clients} by
    \begin{align}
        \mathcal{O}\left(\sqrt{\frac{1}{n}\left[{\left(\frac{\rho+\parallel a_1-a_2\parallel}{\theta}\right)^2\log\left(\left[ 3,\frac{\theta}{\rho+\parallel a_1-a_2\parallel} \right]^{+}  \right)\log n +\log\left(\frac{n\parallel {a_1+a_2}\parallel}{\theta}\right)}\right]}\right).
        \label{SVM:K=2:2}\nonumber
        \end{align}  
\end{theorem}

Theorem~\ref{SVM:K=2:het-hom} is a special case of a more general one that will follow, Theorem~\ref{general:svm:cmi}; and, for this reason, we state it without proof here. Also, the constants of the $\mathcal{O}$ approximation are provided explicitly in the proof of Theorem~\ref{general:svm:cmi}.


\begin{figure}[htbp]
    \centering
    \begin{subfigure}[b]{0.45\linewidth}
        \centering
        \includegraphics[width=\linewidth]{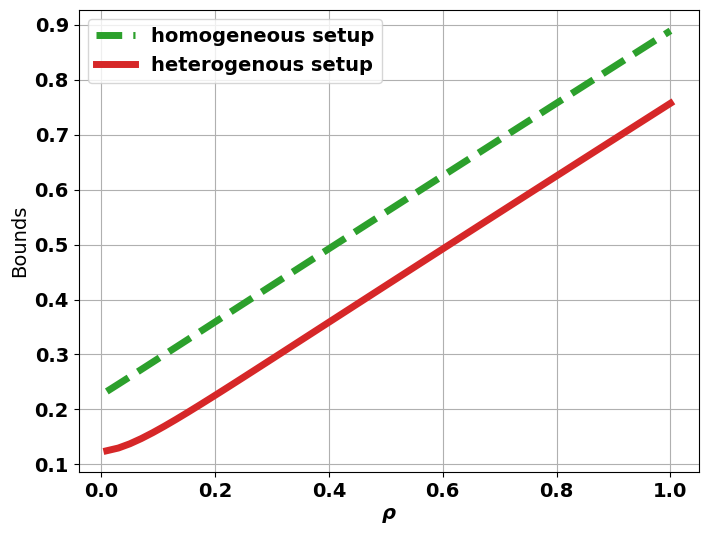}
    \end{subfigure}
    \hfill
    \begin{subfigure}[b]{0.45\linewidth}
        \centering
        \includegraphics[width=\linewidth]{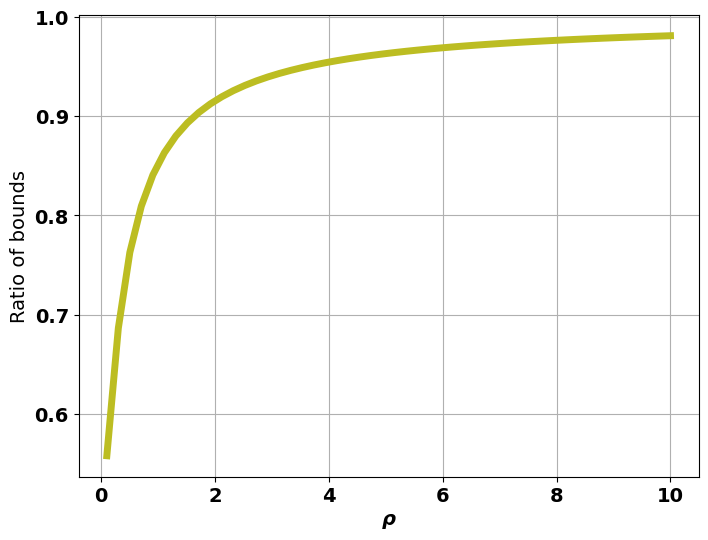}
    \end{subfigure}
    \caption{Evolution of the exact generalization bounds of Theorem~\ref{SVM:K=2:het-hom} (with the constants of the $\mathcal O$ approximation substituted as indicated in~\eqref{eq:svm_pr_rate_def_9}), as well as their ratio, as functions of the ball radius $\rho$, for both heterogeneous and homogeneous data settings. Parameters: $n = 1000$, $\theta = 1$, $K = 2$, $a_1 = (0.2, \mathbf{0}_{d-1})$, and $a_2 = (0.6, \mathbf{0}_{d-1})$.}
    \label{fig:exact:Compare:homo:hetero}
\end{figure}




\vspace{0.4cm}

\subsection{Comparison} 
We compare the bounds of Theorem~\ref{SVM:K=2:het-hom}. Note that this comparison amounts at selecting which training procedure among \textit{Option 1} (the \(n\) training samples of Client 1 drawn according to \(\mu_1^{\text{Het}}\) and those  of Client 2 drawn according to \(\mu_2^{\text{Het}}\)) or \textit{Option 2} (both clients have their \(n\) samples drawn according to \(\mu^{\text{Hom}}=(\mu_1^{\text{Het}}+\mu_2^{\text{Het}})/2\)) yield an aggregated model \(\overline{W}=(W_1+W_2)/2\) that generalizes better during test time. It is important to note that this comparison is \textit{fair}, since the test samples are generated according to the same distribution in both settings, which is \(\mu^{\text{Hom}}=(\mu_1^{\text{Het}}+\mu_2^{\text{Het}})/2\). This is easy to see as, for every \(\overline{w}\) we have
\begin{align}
\mathbb{E}_{Z_1\sim\mu^{\text{Het}}_1}[\ell(Z_1,\overline{w})]&+ \mathbb{E}_{Z_2 \sim \mu^{\text{Het}}_2} [\ell(Z_2,\overline{w})]\nonumber\\
&=2\mathbb{E}_{Z\sim\mu^{\text{Hom}}} [\ell(Z, \overline{w})].\nonumber
\end{align}
Fig.~\ref{fig:exact:Compare:homo:hetero} depicts the evolution of the bounds of Theorem~\ref{SVM:K=2:het-hom} (computed by substituting the constants of the big $\mathcal O$ approximation with their exact expressions as provided in the proof of the theorem)  as function of the ball radius \(\rho\), for both heterogeneous and homogeneous data settings (across clients). Note that, for fixed values of \(n,a_1,a_2,\theta\), increasing  \(\rho\) is equivalent to diminishing the Total Variation distance between the distributions induced by~\eqref{data-distribution-heterogeneous-setting-two-clients}  and~\eqref{data-distribution-homogeneous-setting-two-clients}. In fact, for large values of \(\rho\), the volume of the intersection of the two balls is big, and this augments the probability of the two clients picking `similar' samples. It is observed that the bound for the heterogeneous data setting is tighter (i.e., is smaller) than that for the associated homogeneous data setting. This suggests that, for this example, D-SVM generalizes better when the training data is heterogeneous across clients.    

Finally, for both heterogeneous and homogeneous data settings, the bounds increase with \(\rho\). This is somewhat expected as the ball volume increases with \(\rho\), making it less likely for the generated training samples per-client (whose number \((n)\) is fixed) to be `representatives' of all possible sample realizations over the ball during test time.

\vspace{-0.2cm}

\section{Effect of data heterogeneity on
generalization for D-SVM: General case}\label{SMI:h_d:limit:rho}

In Section~\ref{sec:het_svm}, we considered a distributed SVM setting with two extreme data-heterogeneity setups across two clients: full homogeneity or full heterogeneity. In this section, we generalize the setting of Section~\ref{sec:het_svm} to an arbitrary number of; and, most importantly, with gradually increasing data-heterogeneity setups. 

More formally, fix $M\in \mathbb{N}^*$ arbitrary data distributions $\nu_1,\ldots,\nu_M$ over \(\mathcal{Z} = \mathcal{X} \times \mathcal{Y}\). Denote the \(X\)-marginal of \(\nu_m\), \(m \in [M]\), as \(\nu_{m,X}\).

In the study of the generalization error of SVM, it is common to assume that the data is bounded \cite{gronlund2020,sefidgaran2022rate}. Hence, we assume that there exists $a_m\in \mathbb{R}^d$, $m\in[M]$, and $\rho \in \mathbb{R}^+$, such that
\begin{equation}
    \supp\left(\nu_{m,X}\right) \subseteq B(a_m,\rho),\quad m\in[M],\nonumber
\end{equation}
where $B(a_m,\rho)$ denotes the $d$-dimensional ball with the center $a_m$ and radius $\rho$. Alternatively, we have that
\begin{equation*}
    \mathbb{P}_{X\sim\nu_{m,X}}\left(\left\|X - a_m\right\| \leq \rho \right) = \mathbb{P}_{X\sim\nu_{m,X}}\left(X \in B(a_m,\rho) \right) = 1.
\end{equation*}

Now, we define a family of setups indexed by \(r=1,\hdots,M\) with gradually decreasing levels of data-heterogeneity across \(K \geq M\) clients. Specifically, for every \(r \in [M]\) and every \(k \in [K]\) let \( c_k^{(r)} = (k \mod [M - r + 1]) + 1 \). For \(r=1,\hdots,M\) the \( r \)-th Setup has the clients' data distributions defined each over exactly \(r\) balls, as a suitable mixture of \(r\) measures from the aforementioned set of distributions \(\{\nu_1,\hdots,\nu_M\}\). In particular, this allows for investigating the effect of the clients picking their training samples from partially overlapping data distributions, with the amount of overlap controlled by the value of \(r\). Specifically:

\textbf{\( r \)-th Setup:}  the data distribution $\mu_k^{(r)}$ of client $k$ is 
\begin{equation}
    \mu_k^{(r)} = \sum\nolimits_{m = c_k^{(r)}}^{c_k^{(r)} + r - 1} 
 \alpha_{k,m}^{(r)} \nu_m , \label{setup:rth}
\end{equation}
where $\{\alpha_{k,m}^{(r)}\}$ are arbitrary non-negative coefficients chosen such that $\sum_{k\in[K]}\alpha_{k,m}^{(r)}= 1/M$ for every $(r,m) \in[M]^2$,  $\sum_{m=c^{(r)}_k}^{c^{(r)}_k+r-1}\alpha_{k,m}^{(r)}=1$ for every $(k,r) \in[K]\times [M]$, and $\alpha_{k,m}^{(r)}=\alpha_{k',m}^{(r)}$ if $c_k^{(r)}=c_{k'}^{(r)}$. It is easy to check that for every $r\in[M]$, the set $\left\{\alpha_{k,m}^{(r)}\right\}$ always exists.

Notice that with the data distribution defined as~\eqref{setup:rth}, in the \( r \)-th Setup, client \(k\) picks its training samples from the union of exactly \(r\) balls; namely, those whose indices are in the set \(\{c_k^{(r)},\hdots, c_k^{(r)} +r -1\}\). That is, 
\[
\mathbb{P}_{X_k \sim \mu_{k,X}^{(r)}} \left( X_k\in \bigcup\nolimits _{m = c_k^{(r)}}^{c_k^{(r)} + r - 1} B(a_m, \rho)\right) = 1,
\]
where $\mu_{k,X}^{(r)}$ stands for the $X_k$-marginal of $\mu_{k}^{(r)}$. In particular, this allows distinct clients to pick their samples from a partially overlapping set of balls. For example, the setup $r=M$ has all \(K\) clients picking their training samples from the same distribution \(
    \mu_k^{(M)} = \Big(\sum\nolimits_{m\in[M]}\nu_m\Big)/M\), 
i.e., the data is \emph{homogeneous} across clients. As the value of \(r\) decreases, the level of data heterogeneity across clients increases, reaching its maximum for \(r=1\), a setup for which \(M\) clients (among the \(K\) participating ones) pick their samples from \textit{distinct} distributions over distinct balls. In what follows, we will develop setup-dependent generalization bounds whose comparison will provide insights into the effect of data heterogeneity on the generalization error of the studied D-SVM problem. It should be emphasized that, for the sake of fair comparison, the data distribution during ``test'' time is set to be identical for all clients and setups, given by \((\nu_1+\hdots+\nu_M)/M\). This follows since using~\eqref{setup:rth} and substuting using \(\sum_{k\in[K]}\alpha_{k,m}^{(r)}= 1/M\) we get $\Big(\sum\nolimits_{k\in[K]} \mu_{k}^{(r)}\Big)/K=\Big(\sum\nolimits_{m\in[M]}\nu_m\Big)/M$.

\subsection{Generalization bound for the \texorpdfstring{$r$}{r}'th setup}

Define, for $r\in[M]$,   
\begin{equation}
    D_{k,r}=\max_{(i,j)} \:\: \|a_{i} - a_{j}\|, \nonumber
\end{equation}
where the maximization is over all pairs  $(i,j)\in[{c^{(r)}_k},{c^{(r)}_{k}+r-1}]^2$.

\begin{theorem}\label{general:svm:cmi} Let $\theta \in (0,1]$. Then, for the $r$'th setup defined by \eqref{setup:rth} the expected margin generalization error $\mathbb{E}[\gen_{\theta}(S_{[K]},\overline{W})]$ is upper bounded by
    \begin{equation*}
    \mathcal{O}\left( \sqrt{\frac{\sum_{k=1}^K\biggl[\Bigl(\frac
    {\rho_{k}^{(r)}}{K\theta}\Bigr)^2\log(nK)\log\Bigl(\Bigl[3,\frac{K\theta}{\rho^{(r)}_k}\Bigr]^+ \Bigr) + \log\Bigl(\Bigl[1,\frac{4n\| b^{(r)}_k\|}{K\theta}\Bigr]^{+}\Bigr)\biggr]}{nK}} \right),
    \end{equation*}
    where
    \begin{equation*}
        \rho_{k}^{(r)}{=}\rho+D_{k,r},\quad \quad
    b^{(r)}_k{=}{\sum\nolimits_{m=c_k^{(r)}}^{c_k^{(r)}+r-1}\alpha^{(r)}_{k,m}a_m}.
    \end{equation*}
    \end{theorem}
This Theorem is proved in Appendix~\ref{Proof:SVM:General:K}.

We pause to discuss the result of Theorem~\ref{general:svm:cmi}. First, note that for every setup \(r=1,\hdots,M\) the contribution of Client \(k\) to the bound is, up-to an additive logarithm term, proportional to the squared radius of the smallest ball that contains the union of the \(r\) balls from which this client picks its training sample, i.e., \(\rho_{k}^{(r)}{=}\rho+D_{k,r}\). Interestingly, shifts of these balls (through the values of \((a_1,\hdots,a_m)\) only change marginally the value of the bound. This is by the intuition that the classification error of a cloud of points should depend primarily on the relative spatial repartition of data points of distinct labels concerning each other, rather than the distance to the origin of the entire cloud. Second, the bound depends essentially on \((r,K,M)\) as well as the parameters of the data support for every client, i.e., the values of \(\{\rho^{(r)}_1,\hdots,\rho^{(r)}_K\}\).

Now, we discuss a few special cases and the relation to some known prior art bounds. For the $K=M=2$ setting \(r=1\) one recovers the first bound of Theorem~\ref{SVM:K=2:het-hom}, and setting \(r=2\), one recovers the second bound therein. For $M=1$ and $r=1$ Theorem~\ref{general:svm:cmi} reduces to a bound of order
\begin{equation}
    \mathcal{O}\Bigg(\sqrt{\frac
    {\rho^2 \log(nK)\log\big(\bar{E}\big) + K^2 \theta^2 {\log}\Big(\tilde{E}\Big) }{nK^2\theta^2}}\Bigg),\nonumber
    \end{equation}
    with $\bar{E}=\left[\frac{K\theta}{\rho},3\right]^+\quad \text{and} \quad \tilde{E}=\left[1,\frac{4 n\| a_1\|}{K\theta}\right]^{+}$,
which is better than a previously established bound by~\cite[Theorem~5]{sefidgaran2022rate} which is of order 
\begin{equation}
    \mathcal{O}\Bigg(\sqrt{\frac{(\rho + \|a_1\|)^2 \log(nK){\log}\Big(\Big[\frac{K\theta}{\rho + \|a_1\|},3\Big]^+\Big)}{nK^2\theta^2}}\Bigg).\nonumber
    \end{equation}

\subsection{Improved generalization bound for DVSM in terms of Jensen-Shannon divergence}~\label{sec:svm_hd}
The following theorem, whose proof appears in Appendix~\ref{pr:svm_hd}, provides a possibly better bound in terms of the Jensen-Shannon divergence as captured by \(h_D(\cdot,\cdot)\). 

\begin{theorem}\label{SVM:h_d}
Let $\theta \in (0,1]$. Then, for the $r$'th setup defined by \eqref{setup:rth} the expected margin generalization error $\mathbb{E}[\gen_{\theta}(S_{[K]},\overline{W})]$ is upper bounded by
\begin{equation}
 \mathcal{O}\left( h^{-1}_D\bigg( \hat{E}+\log(n)\Big| \mathbb{E}\left[\hat{\mathcal{L}}_{\theta}(S_{[K]},\overline{W})\right]-\frac{9}{nK\sqrt{K}}\bigg)-\mathbb{E}\left[\hat{\mathcal{L}}_{\theta}(S_{[K]},\overline{W})\right]+\frac{1}{nK\sqrt{K}}\right).\nonumber
    \end{equation}

    where
    \begin{equation*}
        \hat{E}=\frac{1}{nK}\sum\limits_{k\in[K]}\left[\left(\frac
    {\rho^{(r)}_{k}}{K\theta}\right)^2\log(nK)\log\left( \left[ 3,\frac{K\theta}{\sigma} \right]^{+}\right)+\log\left(\left[1,\frac{4n\| b^{(r)}_k\|}{K\theta}\right]^{+}\right)\right],
    \end{equation*}
\end{theorem}
Using this result and Lemma~\ref{h_d:lemma}, it can be easily seen that if the empirical risk is negligible, then the expected margin generalization error is upper bounded by
        $\mathcal{O}\left(\frac{\log(K)\log(nK)}{nK^2}+\frac{\log(n)}{n}\right).$

\subsection{D-SVM with unbounded data support}
So far, we have analyzed SVM algorithms when applied to data with bounded support. In this section, we extend the derived result to the general case of $K$ clients for data with unbounded support. Fix $M\in\mathbb{N}^{\ast}$ and consider the data distributions $\nu_1,\cdots,\nu_M$ such that if $X\sim\nu_m$ then $\|x-a_m\|$ has Folded Gaussian distribution corresponding to a Gaussian variable with zero mean and variance $\sigma^2$, \ie, $\|x-a_m\|=|U|$, where  $U\sim \mathcal{N}(0,\sigma^2)$. More precisely, we consider the probability density function (PDF) of $X$ as 
\begin{align}
    f_{X}(x)=\frac{1}{S_m^{d-1}\left(\|x-a_m\|\right)}\left(\frac{\exp(-\|x-a_m\|^2/(2\sigma^2))}{\sqrt{2\pi\sigma^2}}\right),\label{Uniform_distribution_over_surface_gaussian}
\end{align}  
where $(x, a_m)\in\mathbb{R}^{2d}$, $\sigma^2\in\mathbb{R}^{+}$ and, for $r\in\mathbb{R}^+$, $S_m^{d-1}(r)$ is the surface of a sphere in $\mathbb{R}^d$ with radius $r$, i.e., 
\begin{equation}
S_m^{d-1}(r)=\frac{2\pi^{\frac{d}{2}}r^{d-1}}{\Gamma\left(\frac{d}{2}\right)}.\nonumber
\end{equation}
Similar to in the previous section, we consider a hierarchy of setups with an increasing degree of heterogeneity. Speficially, for the \( r \)-th setup the distribution of the data observed by the \( k \)-th client is given by
\begin{align}
\mu^{(r)}_k =\sum\nolimits_{m=c^{(r)}_k}^{c^{(r)}_k + r - 1}\alpha^{(r)}_{k,m}  \nu_{m},\label{gaussian_mixture_distribution}
\end{align}
 where the coefficients $\{\alpha_{k,m}^{(r)}\in\mathbb{R}^+\}$ are chosen such that $\sum_{k\in[K]}\alpha_{k,m}^{(r)}= 1/M$ for every $(r,m) \in[M]^2$ and  $\sum_{m=c^{(r)}_k}^{c^{(r)}_k+r-1}\alpha_{k,m}^{(r)}=1$ for every $(k,r) \in[K]\times [M]$. Also, $\alpha_{k,m}^{(r)}=\alpha_{k',m}^{(r)}$ if $c_k^{(r)}=c_{k'}^{(r)}$.

\begin{theorem}\label{Gaussian_generalization_theorem}
Let $\theta\in(0,1]$. Then, the expected margin generalization error $\mathbb{E}\left[\gen_{\theta}(S_{[K]},\overline{W})\right]$ in the $r$-th setup is upper bounded by
    \begin{align}
      \mathcal{O}\left(\sqrt{\frac{1}{nK}{\sum\nolimits_{k=1}^{K}\Biggl[\biggl(\frac{\rho^{(r)}_{k}}{K\theta}\biggr)^2\log\left(\left[ 3,\frac{K\theta}{\sigma} \right]^{+}  \right)\log (nK) +\log\left(\Bigl[1,\frac{4n\parallel b^{(r)}_{k}\parallel}{K\theta}\Bigr]^{+}\right)\Biggr]}}\right),
    \end{align}
    where $\rho_{k}^{(r)} {=} D_{k,r}+\sigma\sqrt{\log(nK)}$, 
    with $D_{k,r} {=} \max_{(i,j)\in[M]^2} \|a_{i} - a_{j}\|$, and $
    b^{(r)}_k{=}{\sum\nolimits_{m=c_k^{(r)}}^{c_k^{(r)}+r-1}\alpha^{(r)}_{k,m}a_m}$.      

\end{theorem}
The proof of Theorem~\ref{Gaussian_generalization_theorem} is given in Appendix \ref{Proof_Gaussian_generalization_theorem}.

\subsection{Comparison}

A particularly interesting special case is when the balls are equally spaced, say by some \(\Delta \in \mathbb{R}^{+}\), i.e., $\|a_{m}-a_{m-1}\|=\Delta$ for every $m\in[2:M]$. For simplicity, let \(a_1=\vc{0}_d\). In this case, it is easy to see that the bound of Theorem~\ref{general:svm:cmi} reduces to
\begin{equation}
\mathcal{O}\left(\sqrt{\frac{\bar{A}(K,r,\theta)\log(nK)}{nK^2\theta^2}{+}\frac{1}{n}\log\left(\tilde{A}(M,K,r,\Delta)\right)} \right),
        \label{bound-special-case-DSVM-general-case}
\end{equation}
where $\tilde{A}(M,K,r,\Delta)=\left[1,\frac{n\Delta (2M+r-1)}{K\theta}\right]^{+}$ and 
\begin{equation*}
\bar{A}(K,r,\theta){=} \Big(\rho{+}(r-1)\Delta\Big)^2 \log\left(\left[3,\frac{K\theta}{\rho+(r-1)\Delta}\right]^+\right).
\end{equation*}

Similarly, in this case, the bound of Theorem~\ref{Gaussian_generalization_theorem}  reduces to 
\begin{equation}
\mathcal{O}\left(\sqrt{\frac{\bar{B}(K,r,\theta)\log(nK)}{nK^2\theta^2}{+}\frac{1}{n}\log\left(\tilde{B}(M,K,r,\Delta)\right)} \right),
        \label{bound-special-case-Gaussian-DSVM-general-case}
\end{equation}
 where $\tilde{B}(M,K,r,\Delta)=\left[1,\frac{n\Delta (2M+r-1)}{K\theta}\right]^{+}$
 and 
\begin{equation*}
\bar{B}(K,r,\theta){=} \left({{{D_{k,r}}}+\sigma\sqrt{\log (nK)}}\right)^2 \log\left(\left[3,\frac{K\theta}{\sigma}\right]^+\right).
\end{equation*}

\begin{figure}
    \centering
    \includegraphics[width=0.4\linewidth]{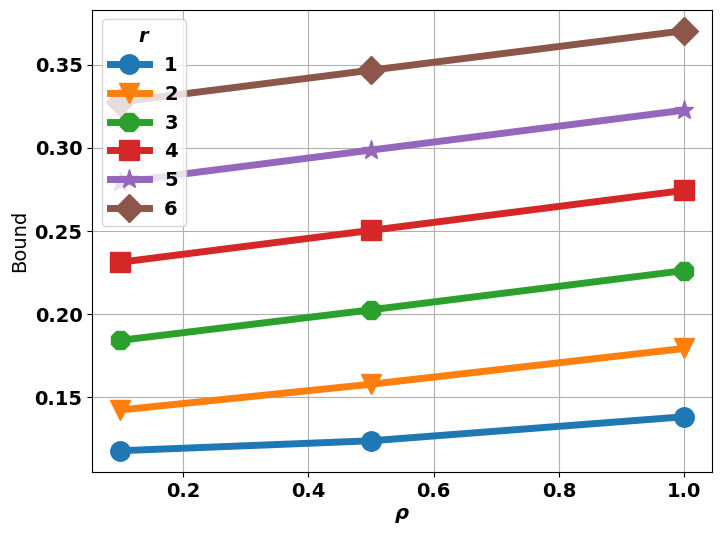}
    \caption{Evolution of the generalization bound~\eqref{bound-special-case-DSVM-general-case} for various degrees of data heterogeneity across clients.}
    \label{fig:bound_vs_rho}
\end{figure}

\begin{figure}
    \centering
    \includegraphics[width=0.4\linewidth]{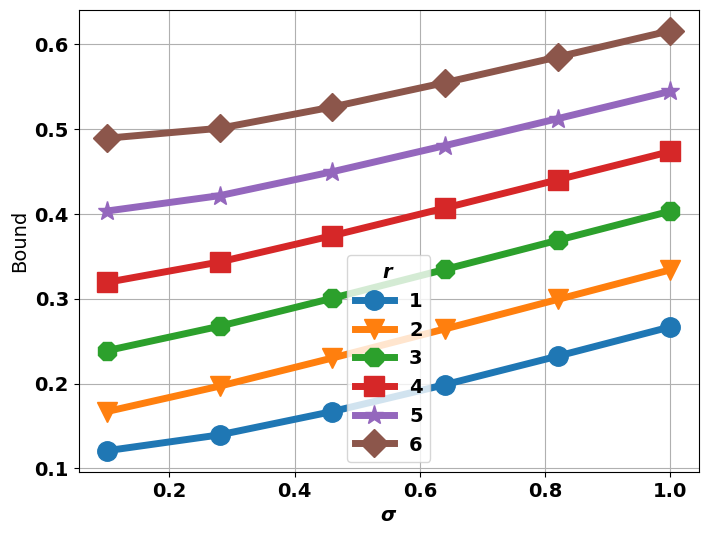}
    \caption{Evolution of the generalization bound~\eqref{bound-special-case-Gaussian-DSVM-general-case} for various degrees of data heterogeneity across clients.}
    \label{fig:bound_vs_sigma}
\end{figure}

\begin{figure}[htbp]
    \centering
    \begin{subfigure}[b]{0.45\linewidth}
        \centering
        \includegraphics[width=\linewidth]{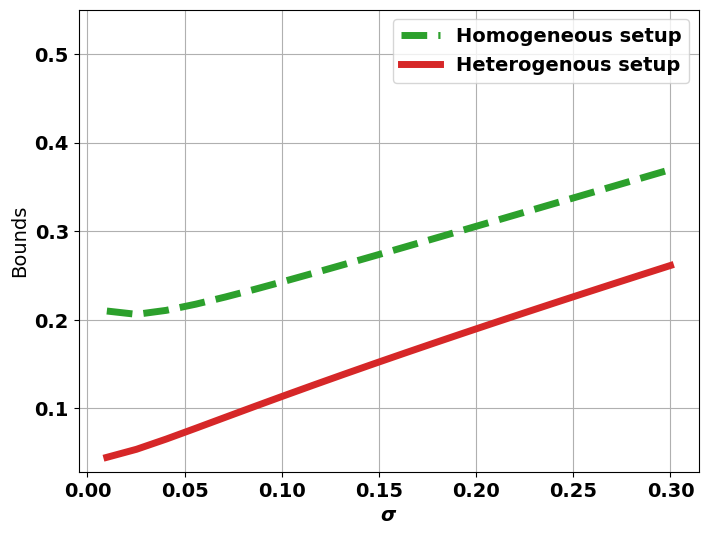}
    \end{subfigure}
    \hfill
    \begin{subfigure}[b]{0.45\linewidth}
        \centering
        \includegraphics[width=\linewidth]{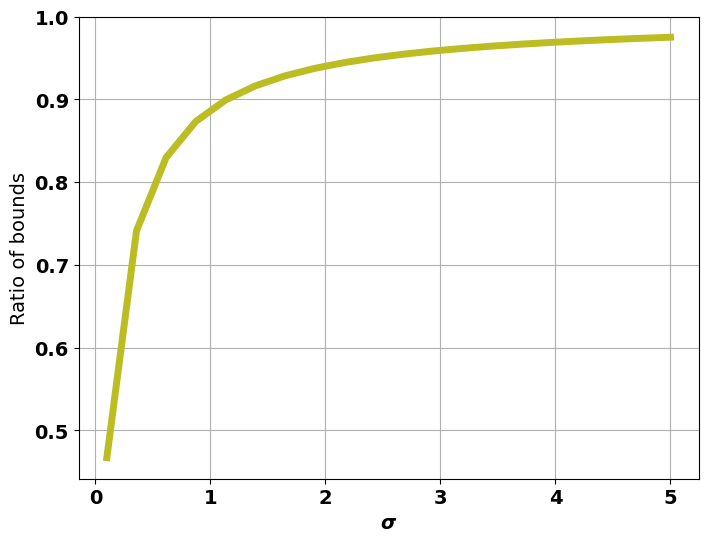}
    \end{subfigure}
    \caption{Evolution of the exact generalization bounds of \eqref{Hetero_Gaussian_comparing_K=2_1} and \eqref{Hetero_Gaussian_comparing_K=2_2} (with the constants of the $\mathcal O$ approximation), as well as their ratio, as functions of the ball radius $\sigma$, for both heterogeneous and homogeneous data settings. Parameters: $n = 10000$, $\theta = 1$, $K = 2$, $a_1 = (0.2, \mathbf{0}_{d-1})$, and $a_2 = (0.6, \mathbf{0}_{d-1})$.}
    \label{Homo_Hetro_gaussian_comparing}
\end{figure}

Fig.~\ref{fig:bound_vs_rho} depicts the evolution of the bound~\eqref{bound-special-case-DSVM-general-case} versus $\rho$ for various values of \(r=1,\hdots,M\), for an example D-SVM setting with \(K=50\), \(M=6\), \(n=1000\), \(\theta=1\) and \(\Delta=1\). As it is visible from the figure the bound on the expected generalization is better (i.e., smaller) for smaller values of \(r\), indicating that the aggregated model \(\overline{W}=(W_1+\hdots+W_K)/K\) generalizes better as the degree of training data heterogeneity across clients is bigger. Fig.~\ref{fig:bound_vs_sigma} shows similar results for the bound~\eqref{bound-special-case-Gaussian-DSVM-general-case} whose evolution is depicted as a function of $\sigma$ for the same setting. For the special case $K=2$ the margin generalization bound derived of Theorem~\ref{Gaussian_generalization_theorem} reduces to  
 \begin{equation}  \mathcal{O}\left(\sqrt{\frac{\tilde{E}\log\left( [3,\frac{\theta}{\sigma}]^{+}\right)\log n+\frac{1}{2}\log\left(\frac{n^2\parallel a_1\parallel \parallel a_2\parallel}{\theta^2}\right)}{n}}\right)
 \label{Hetero_Gaussian_comparing_K=2_1}
        \end{equation}
for the heterogeneous data setting (i.e., $r=1$); and to
    \begin{align}     \mathcal{O}\left(\sqrt{\frac{\bar{E}\log\left([ 3,\frac{\theta}{\sigma} ]^{+}  \right)\log n +\log\left(\frac{n\| a'\|}{\theta}\right)}{n}}\right)
    \label{Hetero_Gaussian_comparing_K=2_2}
        \end{align}   
        for the homogeneous data setting (i.e., $r=2$), where $\tilde{E}=\left(\frac{\sigma\sqrt{\log n}}{\theta}\right)^2$, $\bar{E}=\left(\frac{{{\|a_1-a_2\|}}+\sigma\sqrt{\log n}}{\theta}\right)^2$  and \(a'=\frac{(a_1+a_2)}{2}\). These bounds~\eqref{Hetero_Gaussian_comparing_K=2_1} and~\eqref{Hetero_Gaussian_comparing_K=2_2} are compared in Fig. ~\ref{Homo_Hetro_gaussian_comparing}, from which it can be seen that the result of Theorem~\ref{Gaussian_generalization_theorem} is tighter in smaller (i.e., better) in the across-clients heterogeneous data setting.

\subsection{Discussion}

The aforementioned results advocate in favor of data heterogeneity across clients during training phase, in the sense that this \textit{provably}\footnote{It is shown in Section~\ref{sec:experiments} that data heterogeneity across clients not only makes the bounds smaller but also the actual, measured, generalization error for the experiments therein.}  helps for a better generalization. However, caution should be exercised in the interpretation of such findings regarding the effect of data heterogeneity on the population risk.  In particular, while there are reasons to believe that there might indeed exist cases in which heterogeneity helps also for a better (i.e., smaller) population risk (such as for \textit{realizable} setups for which generalization error equals population risk), we make \textit{no} such claim in general. This is because the positive decrease of the generalization error enabled by data heterogeneity may not compensate for the caused increase of the empirical risk, causing the population risk to be larger - see Fig.~\ref{fig:gen_and_risks_feature_balls}, which shows the empirical and population risks for Experiment 1 that will follow.

\section{Experimental results}~\label{sec:experiments}

We report the results of various experiments, the first about DSVM \ie a distributed learning setup where each client trains an SVM model, but with different datasets with heterogeneity. Then, we go beyond DSVM with neural network local models. We also conducted experiments in the more general multi-round Federated Learning setup, which yielded the same observations as in the distributed learning setup considered so far. This is an important insight for future theoretical work.
Full details of all experiments, as well as additional experiments, are given in the appendices.




\begin{figure}[htpb!]
    \centering
    \begin{subfigure}{0.49\linewidth}
        \centering
        \includegraphics[width=\textwidth]{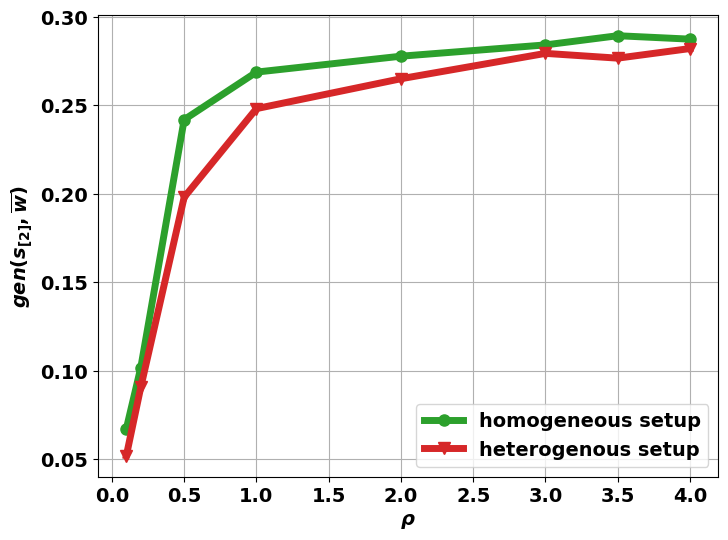}
    \end{subfigure}
    \begin{subfigure}{0.49\linewidth}
        \centering
        \includegraphics[width=\textwidth]{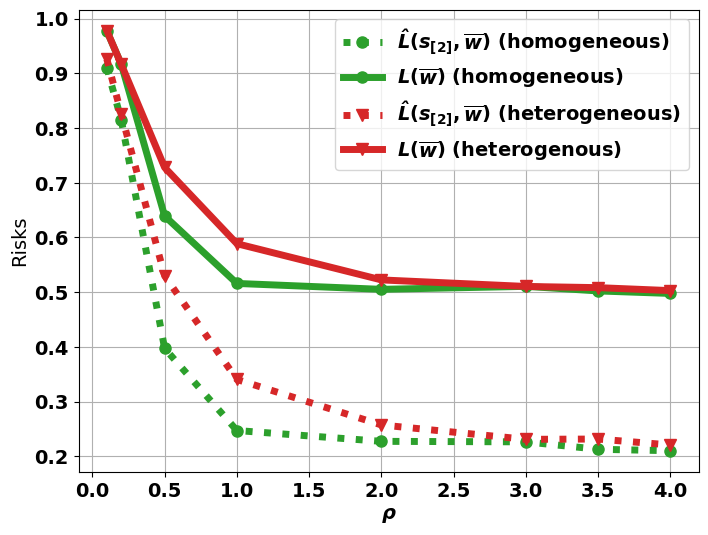}
    \end{subfigure}
    \caption{Measured gen. error (left), along with empirical and population risks (right) as function of radius $\rho$, for Experiment~1.}
    \label{fig:gen_and_risks_feature_balls}
\end{figure}

\textbf{Experiment 1 (Synthetic data with feature heterogeneity across clients)}: In this experiment, we consider binary classification using DSVM with synthetic data in dimension $d = 100$, generated as described in Section~\ref{sec:het_svm}. Fig.~\ref{fig:gen_and_risks_feature_balls} shows the evolution of the generalization error for the homogeneous and heterogeneous setups of Section~\ref{sec:het_svm} as a function of the balls' radius $\rho$. 
As can be seen, for all values of $\rho$, the across-clients heterogeneous training data procedure yields a better (i.e., smaller) generalization error than the associated across-clients homogeneous training data procedure. \\

\textbf{Experiment 2 (Parity prediction on MNIST with digit structure-induced feature heterogeneity)}:

In this experiment, we consider binary classification of a subset of the MNIST dataset having digits $\{0,1,2,3,4,6,7\}$ based on the parity of the depicted digit. 
\begin{figure}[htpb!]
    \centering
    \includegraphics[width=0.5\textwidth]{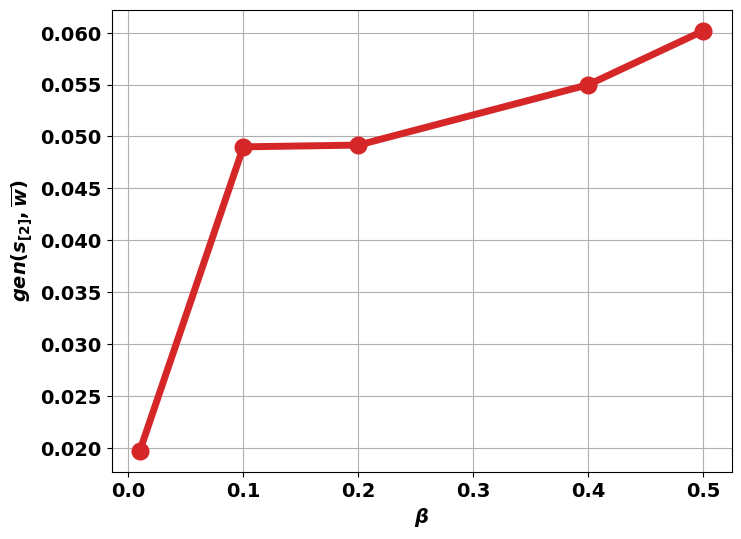}
    \caption{Measured gen. error for Experiment 2 vs. degree $(\beta)$ of feature heterogeneity across clients -- smaller $\beta$ means more heterogeneity.}
    \label{fig:gen_parity_mnist}
\end{figure}
There are two clients. For some $\beta \in [0,0.5]$, Client 1 is given $100 \beta \%$ of those training samples whose indices are in the set $\{0,1,2,3\}$ and $(100 - 100 \beta)\%$ of those training samples whose indices are in the set $\{4,5,6,7\}$. Similarly, Client 2 is given the complementary training set, i.e., $(100 - 100\beta) \%$ of the training samples with indices in $\{0,1,2,3\}$ and $100 \beta\%$ of the training samples with indices in $\{4,5,6,7\}$. The task is to predict the parity of the depicted digit.
Observe that with the considered training samples split between the two clients, both get half of the samples with label $0$ and half of the samples with label $1$, \ie there is \textit{no} label heterogeneity across clients in the training data. However, the clients’ training data exhibits feature heterogeneity that is induced \textit{naturally} by the structure of the digits themselves being different for the two clients. 
For instance, if $\beta = 0$ the training data of Client 1 is composed of digits $\{4,5,6,7\}$ whereas that of Client 2 is composed of digits $\{0,1,2,3\}$, a case which corresponds to \textit{extreme} feature heterogeneity across clients. If $\beta = 0.5$, for every one of the two clients, the training data is composed of $50\%$ of the digits $\{0,1,2,3\}$ and $50\%$ of the digits $\{4,5,6,7\}$, i.e., \textit{feature homogeneity across clients}. The two clients use the same CNN for local training; and we set the number of local epochs such that the training loss is sufficiently minimized. By varying $\beta \in [0,0.5]$, we simulate the effect of various degrees of feature heterogeneity on the generalization error. 

The results of this experiment, presented in Fig.~\ref{fig:gen_parity_mnist}, show that the measured generalization error of the aggregated model decreases as $\beta \rightarrow 0$ (extreme data heterogeneity); and it increases as $\beta \rightarrow 0.5$ (data homogeneity). \\

\textbf{Experiment 3 (FL with label heterogeneity across clients)}: In this experiment, we consider binary classification using the datasets MNIST and CIFAR10. For each dataset, the training samples are split equally among the two clients, but by inducing label heterogeneity among them. For instance, Client 1 is assigned a fixed proportion \(\alpha \in [0, 0.5]\) of the entire training samples from class 1 and a proportion \((1-\alpha)\) of the training samples from class 3. Client 2 has the remaining training samples, i.e., proportion \((1-\alpha)\) of the samples from class 1 and \(\alpha\) of the samples from class 3. Every client trains a CNN locally, and communicates with the server over $R$ rounds. It is important to note that the value of $R$ is set to be \textit{identical} for all degrees of heterogeneity $\alpha \in [0,1/2]$, i.e., $R(\alpha_1) = R(\alpha_2)$ for every $(\alpha_1,\alpha_2) \in [0,1/2]^2$ with $\alpha_1 \neq \alpha_2$. In our experiments, this value of $R$ is chosen large enough (300 for the MNIST dataset and 150 for the CIFAR dataset) such that the global training loss is kept below a targeted (small) threshold for every value of heterogeneity degree $\alpha \in [0,1/2]$. 
\begin{figure}[htpb!]
    \centering
    \includegraphics[width=0.5\textwidth]{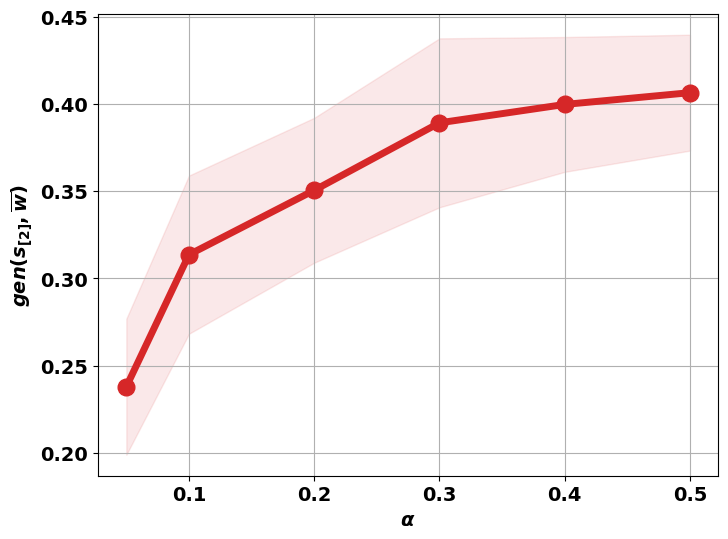}

    \caption{Measured gen. error for Experiment 3 vs. degree $(\alpha)$ of label heterogeneity -- smaller $\alpha$ means more heterogeneity.}
    \label{fig:gen_label_cifar}
\end{figure}
The results, 
are shown in Fig.~\ref{fig:gen_label_cifar} for the CIFAR10 dataset. Similar results using MNIST can be found in the appendices. 
As it is visible from Fig.~\ref{fig:gen_label_cifar}, bigger degrees of heterogeneity (\ie smaller \(\alpha \in [0, 1/2]\)) yield smaller generalization error. It is worth noting that this experiment, which somewhat stretches our problem setup by accommodating multiple rounds communication between the clients and the server, also indicates that the observations and insights of this paper (on the effect of data heterogeneity across clients on the generalization error) may hold more generally, beyond the setup of Section~\ref{sec:het_svm}. \\

\textbf{Experiment 4 (FL with label heterogeneity: the effect of variable-communication rounds)}: We re-run Experiment 4 with the difference that, this time, the number of communication rounds $R$ between the clients and the server depends on the degree of data heterogeneity $\alpha \in [0,1/2]$. Specifically, we fix a threshold ``TrainErr" on the training error of the aggregated model (e.g., TrainErr=0.05); and, for a given value of the degree of data heterogeneity $\alpha$, the clients communicate with the server over a number of rounds $R(\alpha)$ that is $\alpha$-dependent and stop doing so when the empirical risk of the aggregated model falls below the fixed threshold for the training error. In particular, for distinct values of heterogeneity $\alpha_1 < \alpha_2$ we have $R(\alpha_1) > R(\alpha_2)$.

\begin{figure}[htpb!]
    \begin{subfigure}{0.49\linewidth}
        \centering
        \includegraphics[width=\textwidth]{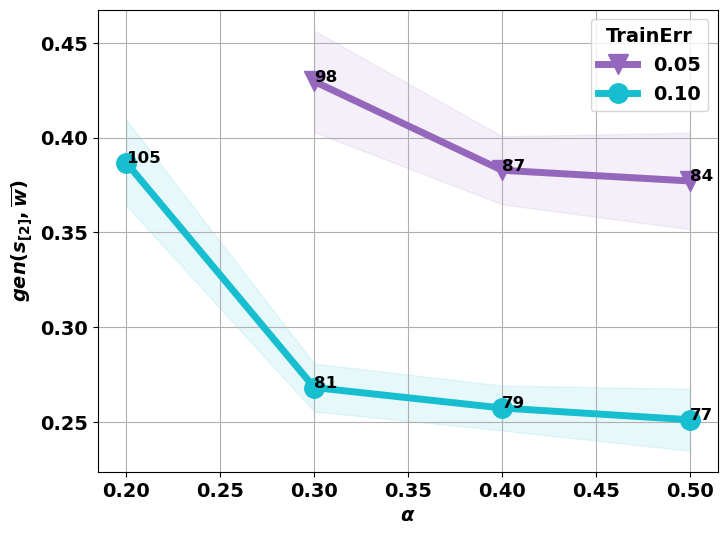}
    \end{subfigure}
    \begin{subfigure}{0.49\linewidth}
        \centering
        \includegraphics[width=\textwidth]{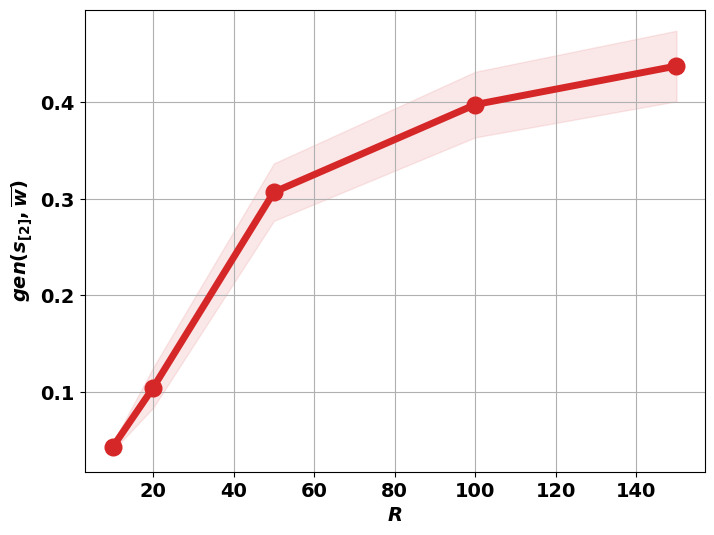}
    \end{subfigure}
    \caption{(Left) Measured gen. error for Experiment 4 vs. degree $(\alpha)$ of label heterogeneity -- smaller $\alpha$ means more heterogeneity. (Right) Gen. error for $\alpha=0.2$ vs. number of communication rounds $R$.}
    \label{fig:gen_comp_R}
\end{figure}

The left plot of Fig.~\ref{fig:gen_comp_R} shows the values of the measured generalization error for the MNIST dataset for two values of the threshold TraiErr on the empirical risk, TrainErr=0.05 and TrainErr=0.10. Observe that, now, like the experimental results reported in~\cite[Fig. 1(a)]{sun2024understanding} the generalization error decreases with the degree of data heterogeneity across clients $\alpha \in [0,1/2]$. That is, with a variable number of communication rounds a bigger degree of data heterogeneity across clients yields a larger generalization error of the aggregated model. \\


\textbf{Discussion:} As already mentioned, the setup of the experiment of~\cite[Fig. 1(a)]{sun2024understanding}, as well as that of our similar Experiment 4, set the interactions between the clients and the server to terminate when the training error reaches a fixed threshold TrainErr, at variable number of rounds for distinct levels of data heterogeneity across clients. It is important to note that, in this case, the comparison of the generalization errors at distinct levels of heterogeneity $\alpha_1 < \alpha_2$ actually amounts to merely comparing the associated test errors. Or, given that the empirical risk is known to typically increase with the degree of data heterogeneity across clients, it is \textit{somewhat natural} to expect that so does the test error in the variable number of communication rounds setting. 

Moreover, note that (i) prior work has established that for a given fixed degree of data heterogeneity the generalization error increases with the number of communication rounds between the clients and the server~\cite{sefidgaran2024lessons} as also visible from Fig.~\ref{fig:gen_comp_R} for Experiment 4 with $\alpha=0.2$ and (ii) in general the bigger the degree of heterogeneity (smaller $\alpha$) the more communication rounds needed to reach a prescribed small value TrainErr of the training error (i.e., if $\alpha_1 < \alpha_2$ then $R(\alpha_1) > R(\alpha_2)$), as per the annotations on each point of the curves in Fig.~\ref{fig:gen_comp_R} (left plot). Therefore, formally in the variable number of communication rounds setting, the increase of the generalization error with the degree of label heterogeneity across clients observed in~\cite[Fig. 1(a)]{sun2024understanding} and our Fig.~\ref{fig:gen_comp_R} can \textit{not} be attributed solely to the effect of the degree of label heterogeneity being larger at $\alpha_1 < \alpha_2$. 
In other words, these observations may be due to communicating more with the server for $\alpha_1$ than for $\alpha_2 > \alpha_1$. That is precisely the reason for which, in this paper, in order to isolate the effect of data heterogeneity across clients, we set the number of communication rounds to be identical for distinct heterogeneity levels.

\bibliographystyle{IEEEtran}
\bibliography{bib}


\onecolumn
\begin{center}
\Large \bf Appendices
\end{center}

The appendices are organized as follows:
\begin{itemize}
    \item Appendix~\ref{app:add_exp} contains additional experimental results.

    \item Appendix~\ref{app:exp_details} contains the details of the experiments presented in the paper.
        
    \item Appendix~\ref{sec:proofs} contains all the proofs of the results of the papers, in the order of their appearance, that is:
    \begin{itemize}
        \item Proof of Theorem~\ref{CMI:FL:TH} presented in Appendix~\ref{proof:inexp:cmi:general},

        \item Proof of Theorem~\ref{tailbound:cmi:general} presented in Appendix~\ref{Tail:proof:CMI},

        \item Proof of Theorem~\ref{Lossy:inexpectation:general} presented in Appendix~\ref{Lossy:general:CMI:1},

        \item Proof of Theorem~\ref{h_D:inexpectation:general} presented in Appendix~\ref{h_D:inexpectation:general:proof},
        
        \item Proof of Theorem~\ref{h_d:ingeneral} presented in Appendix~\ref{Lossy:h_d:inverse},

        \item Proof of Theorem~\ref{h_d:tail:general} presented in Appendix~\ref{Tail:h_d},

        \item Proof of Theorem~\ref{general:svm:cmi} presented in Appendix~\ref{Proof:SVM:General:K},

        \item Proof of Theorem~\ref{SVM:h_d} presented in Appendix~\ref{pr:svm_hd},

\item Proof of Theorem~\ref{Gaussian_generalization_theorem} presented in Appendix~\ref{Proof_Gaussian_generalization_theorem}

        \item Proof of Lemma~\ref{h_d:lemma} presented in Appendix~\ref{Proof:Lemma:auxiliary:h_d:property},
        
        \item  Proof of Lemma \ref{simp:expect:loss:1} presented in Appendix~\ref{Proof:auxiliary:lemma:h_d}.

    \end{itemize}
\end{itemize}


\appendices
\section{Additional experimental results} \label{app:add_exp}

\subsection{Additional experiment 1}

\begin{figure}[htpb!]
    \centering
    \includegraphics[width=0.5\linewidth]{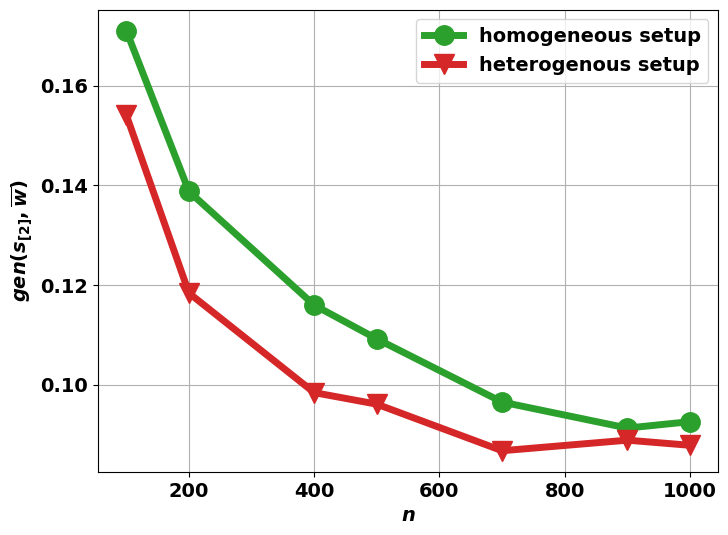}
   \caption{Measured generalization error vs. $n$ for Additional experiment 1.}
    \label{fig:gen_feature_mnist}
\end{figure}

In this experiment, we consider binary classification with two classes (here 1 and 6) of the MNIST dataset~\cite{lecun1998mnist}. In order to induce feature heterogeneity, we add Gaussian white noise with standard deviation $\sigma = 0.2$ to half of the training MNIST images. Then, two setups are compared. In the heterogeneous data setup, Client 1 possesses all the noisy data, while the second one has only the non-noisy original images. In the homogeneous setup, every client picks its data uniformly at random from noisy and non-noisy digits, thus resulting in half of its training samples being noisy and the other half non-noisy. Fig.~\ref{fig:gen_feature_mnist} shows the evolution of the generalization error for both homogeneous and heterogeneous setups. The reported values are averaged over 100 independent runs, each performed using 200 local SGD epochs before aggregation. Here too, as it is visible from the figure, feature-heterogeneity helps for a better, \ie smaller, generalization error. 

\subsection{Additional experiment 2}

\begin{figure}[htpb!]
    \centering
    \includegraphics[width=0.5\linewidth]{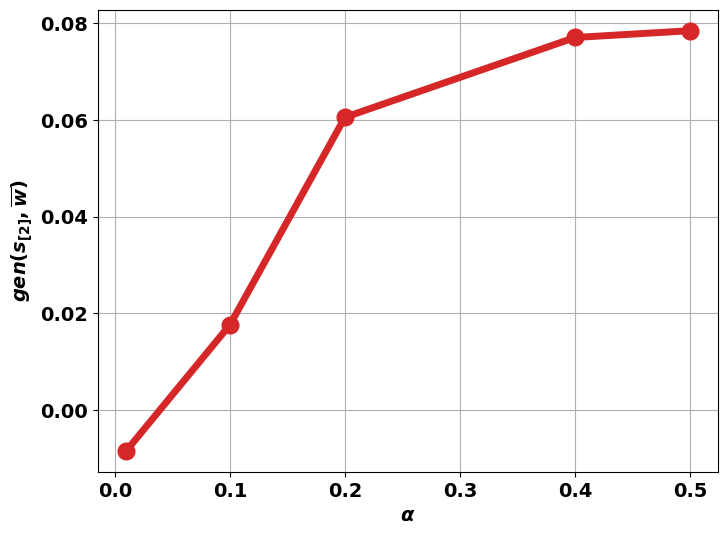}
    \caption{Measured gen. error for Add. experiment 2 vs. degree $(\alpha)$ of label heterogeneity -- smaller $\alpha$ means more heterogeneity.}
    \label{fig:gen_label_mnist}
\end{figure}

This experiment has the same setup as Experiment 3, but uses two classes (6 and 9) of the MNIST dataset. Fig.~\ref{fig:gen_label_mnist} shows the generalization error versus the degree of heterogeneity $\alpha$. One can see that, similarly to Experiement 3 that uses CIFAR-10, the generalization error is smaller when $\alpha$ decreases \ie when the data is more heterogeneous across clients.

\section{Details of experimental results}
\label{app:exp_details}

\subsection{Experiment 1}

\begin{figure}[htpb!]
    \centering
    \includegraphics[width=0.4\linewidth]{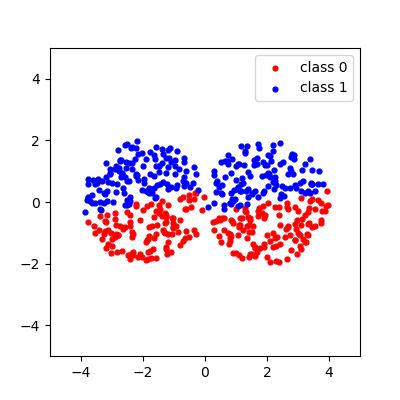}
    \caption{Synthetic data for Experiment 1, $d = 2$}
    \label{fig:balls_d=2}
\end{figure}

For the first experiments, we use synthetic data, generated as explained in Section~\ref{sec:experiments} of the paper. The data dimension is $d = 100$. The two balls have the following characteristics.
\begin{itemize}
    \item Ball 1: 
    \begin{itemize}
        \item Center: $a_1 = (-2, 0,  \ldots, 0)^\top$
        \item Radius: $\rho = 2.0$
        \item Labels: $y = \mathbbm 1_{w^T x + a_1 / 5 > 0}$, where $w = (-0.2, 1, \ldots, 1)$
    \end{itemize}
    \item Ball 2:
    \begin{itemize}
        \item Center: $a_2 =  (2, 0, \ldots, 0)^\top$ 
        \item Radius: $\rho = 2.0$
        \item Labels: $y = \mathbbm 1_{w^T x + a_2 / 5 > 0}$, where $w = (-0.2, 1, \ldots, 1)$
    \end{itemize}
\end{itemize}
See Fig.~\ref{fig:balls_d=2} for an illustration of the synthetic data for dimension $d = 2$.

To illustrate our theoretical results, in particular the generalization bounds of Theorems 4 and 5, the two clients train an SVM model. They each perform 300 epochs using SGD with a learning rate of 0.005. Moreover, the whole setup has been run 300 times to account for the overall randomness and estimate the expectation within the bounds of Theorems 4 and 5.

\subsection{Experiment 2}

This experiment uses digits 0 to 7 of MNIST. The feature heterogeneity generation is explained in the main paper.
Each client trains a CNN with a single convolutional layer and a fully-connected layer. The optimizer is mini-batch SGD with batch size 64 and learning rate 0.05, and the loss is the binary cross-entropy. Each client trains for 200 local epochs. The setup was run independently 150 times.

\subsection{Experiment 3}

In this experiment, beyond the setup considered for the theoretical results of our paper, we use the MNIST and CIFAR10 datasets. We extract two classes out of it (resp. 6+9 for MNIST and 1+3 for CIFAR10) to perform binary classification. 
The only preprocessing that has been performed is normalization of the images.

Each client here trains a CNN with 
\begin{itemize}
    \item two convolutional layers, a dropout layer, and two fully-connected layers for the MNIST experiment,
    \item two convolutional layers, a dropout layer, and three fully-connected layers for the CIFAR10 experiment.
\end{itemize}

We minimize the binary cross-entropy loss, using mini-batch SGD with batch size 64 and learning rate 0.01 (resp. 0.1 for CIFAR10). 300 (resp. 150) communication rounds were run, and simulations were performed 100 times, independently.

\subsection{Experiment 4}

The setup here is the same as Experiment 3. We considered only CIFAR10. The only difference is now that the number of communication rounds is different for each value of alpha, and the models are trained until they reach the training loss threshold TrainErr, or a maximum number of rounds of 300.

\subsection{Additional experiment 1} 

The data used for the first additional experiment consists in two classes extracted from the MNIST dataset (1 and 6). The images were normalized and projected into a space of dimension $d = 2000$ using a Gaussian kernel with scale parameter $\gamma = 0.01$. Then, AWGN with standard deviation $\sigma = 0.2$ was added to the images. We still consider a two-client distributed setup, where each client trains an SVM model using SGD with a learning rate of 0.01. 200 local epochs were run and the simulations were performed and simulations were performed 100 times.

\subsection{Additional experiment 2}

See details of Experiment 3.

\subsection{Implemental and hardware details}

All experiments were done using Python 3.12.7 on a machine with the following specifications:
\begin{itemize}
    \item CPU: AMD Ryzen 7 5800X (8 cores)
    \item GPU: Nvidia Geforce RTX 3070 
    \item RAM: 32 GB
\end{itemize}

SVM models were implemented using the Scikit-learn library. In particular, we used ``RBFSampler'' for kernel projection. CNN models were implemented using the Pytorch library.


\section{Proofs} \label{sec:proofs}

\subsection{Proof of Theorem \ref{CMI:FL:TH}}\label{proof:inexp:cmi:general} 
Recall Definition~\ref{def:symmetric}. Also, recall the definition of the membership vectors \(\mathbf{J}_k\) and  \(\mathbf{J}^c_k\) as given in the beginning of Section~\ref{sec-CMI-type-gneralization-bopunds}. Let, for \(k \in [K]\), \(\mathcal{Q}_k\) be the set of type-I symmetric priors on \(W_k\) conditionally given \((S_k,S'_k)\). The proof consists of two steps: In the first step, we prove that
\begin{align}
\inf_{{Q}_k \in \mathcal{Q}_k} \mathbb{E}_{S_k, S^{\prime}_k} \left[ D_{KL}\left(P_{W_k|S_k,S^{\prime}_k} \parallel Q_k \right) \right]=I\left( W_k ; \mathbf{J}_k|\mathfrak{Z}_k^{2n} \right). \label{eq:th1_step1}
\end{align}
In the second step, we show that for every $(Q_1,\hdots,Q_K) \in \mathcal{Q}_1\times \hdots \times \mathcal{Q}_K$ it holds that
\begin{align}
    \mathbb{E}_{S_{[K]}, \overline{W}} \left[ \gen(S_{[K]}, \overline{W}) \right] \leq \sqrt{\frac{2E}{n}},
    \end{align}
    where
    \begin{align}
        E &= \frac{1}{K} \sum_{k=1}^K \mathbb{E}_{S_k, S_k'} \left[ D_{KL}\left( P_{W_k \mid S_k, S_k'} \parallel {Q}_{k}\right) \right].\label{variational:form:mutualinf:3} 
    \end{align}

To show the first step, consider the set $\mathcal{Q}_k^{\prime}$ of all conditional priors ${Q}_k^{\prime}$ that can be expressed as
\begin{align}
Q_k^{\prime}\left(W_k|S_k,S_k^{\prime}\right) = \mathbb{E}_{\mathbf{J}_k}\left[Q_{k,1}^{\prime}\left(W_k| \mathfrak{Z}^{2n}_{\mathbf{J}_k}, \mathfrak{Z}^{2n}_{\mathbf{J}_k^c}\right) \right]
\label{type-I-symmetric-conditional-prior}
\end{align}
for some arbitrary conditional distribution $Q_{k,1}^{\prime}$. It is easy to verify that $\mathcal{Q}_k = \mathcal{Q}_k^{\prime}$. Therefore, we have
\begin{align}
\inf_{{Q}_k \in \mathcal{Q}_k} \mathbb{E}_{S_k, S^{\prime}_k} \left[ D_{KL} \left( P_{W_k \mid S_k, S^{\prime}_k} \parallel Q_k \right) \right]
= \inf_{{Q}_k^{\prime} \in \mathcal{Q}_k^{\prime}} \mathbb{E}_{S_k, S^{\prime}_k} \left[ D_{KL} \left( P_{W_k \mid S_k, S^{\prime}_k} \parallel Q_k^{\prime} \right) \right].
\end{align}
Recall that the vector $\mathfrak{Z}^{2n}_k$ is a re-arrangement of the elements of $(S_k, S_k^{\prime})$, indexed by the vector $\mathbf{J}_k$. Using this, we get \
\begin{align}
    \inf_{{Q}_k\in \mathcal{Q}_k}\mathbb{E}_{S_k,S^{\prime}_k}\left[D_{KL}\left(P_{W_k\lvert S_k,S^{\prime}_k}\parallel{Q}_k\right)\right]
    &=\inf_{{{Q}}^{\prime}_k\in \mathcal{Q}^{\prime}_k}\mathbb{E}_{S_k,S^{\prime}_k}\left[D_{KL}\left(P_{W_k\lvert S_k,S^{\prime}_k}\parallel{Q}^{\prime}_k\right)\right]\nonumber\\
    &=\inf_{{{Q}}^{\prime}_k\in \mathcal{Q}^{\prime}_k}\mathbb{E}_{\mathfrak{Z}_k^{2n}}\mathbb{E}_{\mathbf{J}_k}\left[D_{KL}\left(P_{W_k\lvert {\mathfrak{Z}^{2n}_{{\mathbf{J}}_k},{\mathfrak{Z}^{2n}_{\mathbf{J}_k^c}}}}\parallel{Q}^{\prime}_k\right)\right]\nonumber\\
    &=\inf_{{Q}^{\prime}_{k,1}}\mathbb{E}_{\mathfrak{Z}_k^{ 2n}}\mathbb{E}_{\mathbf{J}_k}\left[D_{KL}\left(P_{W_k\lvert {\mathfrak{Z}^{2n}_{\mathbf{J}_k},{\mathfrak{Z}^{ 2n}_{\mathbf{J}_k^c}}}}\parallel\mathbb{E}_{\mathbf{J}_k}\left[{Q}^{\prime}_{k,1}\left(W_k\lvert \mathfrak{Z}^{2n}_{{\mathbf{J}}_k},\mathfrak{Z}^{2n}_{{\mathbf{J}}^c_k}\right)\right]\right)\right]\nonumber\\
    &=I(W_k;\mathbf{J}_k\lvert \mathfrak{Z}_k^{2n});
    \label{eq:th1_step1_1}
\end{align}
where the third equality follows using~\eqref{type-I-symmetric-conditional-prior}; and this completes the proof of the first step.

We now turn to the proof of the second step. By~\eqref{definition-generalization-error}, for arbitrary \(\lambda\) we have  
    \begin{align}
\lambda\mathbb{E}_{S_{[K]},\overline{W}}\left[\gen\left(S_{[K]},\overline{W}\right)\right]&=\frac{\lambda}{K}\sum_{k\in[K]}\mathbb{E}_{S_k,\overline{W}}\left[\gen\left(S_k,\overline{W}\right)\right]\nonumber\\
        &=\frac{\lambda}{K}\sum_{k\in[K]}\mathbb{E}\left[\frac{1}{n}\sum_{i\in [n]}\bigg(\ell(z^{\prime}_{k,i},\overline{W})-\ell(z_{k,i},\overline{W})\bigg)\right]\label{CMI:generalizationbound:KL:3}\\
        &\leq\sum_{k\in[K]}\frac{1}{K}\Bigg(D_{KL}\left({\mu}^{\otimes 2n}_k\otimes P_{\overline{W},W_k\lvert S_k,S^{\prime}_k}\parallel{\mu}^{\otimes 2n}_k\otimes{Q}_k\otimes\overline{P}_k\right)\nonumber\\
&\hspace{2.5 cm}+\log\mathbb{E}\left[e^{\frac{\lambda}{n}\sum_{i\in [n]}\left(\ell(z^{\prime}_{k,i},\overline{W})-\ell(z_{k,i},\overline{W})\right)}\right]\Bigg),\label{CMI:generalizationbound:KL:2}
    \end{align}
    where:
    \begin{itemize}
    \item ${Q}_k\left(W_k|S_k, S^{\prime}_k\right)$ and $P_{\overline{W}|W_k, S^{\prime}_k, S_k}$ are abbreviated as ${Q}_k$ and $\overline{P}_k$, respectively,
    \item the expectation in \eqref{CMI:generalizationbound:KL:3} and \eqref{CMI:generalizationbound:KL:2} is taken with respect to \((S_k, S^{\prime}_k, \overline{W}, W_k)\), with the joint distribution being \(P_{S^{\prime}_k} \otimes P_{S_k, \overline{W}, W_k}\) for \eqref{CMI:generalizationbound:KL:3} and \(\mu^{\otimes 2n}_k \otimes Q_k \otimes \overline{P}_k\) for \eqref{CMI:generalizationbound:KL:2},
 \item and \eqref{CMI:generalizationbound:KL:2} follows by application of Donsker-Varadhan's variational representation, using that the loss is bounded and so sub-Gaussian.
     \end{itemize}

Now, we proceed to upper bound the second term of the RHS of~\eqref{CMI:generalizationbound:KL:2}. Recall that for a membership vector 
$\mathbf{J}_k=\{J_{k,1},\cdots,J_{k,n}\}$ the vector \(\mathfrak{Z}^{2n}_{\mathbf{J}_k} \in \mathcal{Z}^{2n}\) stands for the size-\(n\) sub-vector of vector \(\mathfrak{Z}^{2n}_{k}\) whose elements are indexed by $\mathbf{J}_k$.  Thus, we have
 \begin{align}
\log\mathbb{E}\Big[e^{\frac{\lambda}{n}\sum_{i\in [n]}\left(\ell(z^{\prime}_{k,i},\overline{W})-\ell(z_{k,i},\overline{W})\right)}\Big]
&=\log\mathbb{E}\left[e^{\frac{\lambda}{n}\sum_{i\in[n]}\ell(\mathfrak{Z}_{J^c_{k,i}},\overline{W})-\ell(\mathfrak{Z}_{J_{k,i}},\overline{W})}\right]\label{CMI:generalizationbound:KL:bounding:exponent:2}\\
&=\log\mathbb{E}\left[\mathbb{E}_{\mathbf{J}_k\sim\text{Bern}\left(\frac{1}{2}\right)^{\otimes n}}\left[e^{\frac{\lambda}{n}\sum_{i\in[n]}\ell(\mathfrak{Z}_{J^c_{k,i}},\overline{W})-\ell(\mathfrak{Z}_{J_{k,i}},\overline{W})}\right]\right]\label{CMI:generalizationbound:KL:bounding:exponent:3}\\
&\leq\log\left(\frac{e^{\frac{\lambda}{n}}+e^{-\frac{\lambda}{n}}}{2}\right)^{n}\label{CMI:generalizationbound:KL:bounding:exponent:4}\\
&\leq\frac{\lambda^2}{2n},
\label{CMI:generalizationbound:KL:bounding:exponent:5}
 \end{align}

 where:
 \begin{itemize}
\item the expectation in the LHS of \eqref{CMI:generalizationbound:KL:bounding:exponent:2} is taken over the random variables $(S_k, S^{\prime}_k, W_k, \overline{W})$, distributed according to ${\mu}^{\otimes 2n}_k \otimes {Q}_k\left(W_k\lvert S_k,S^{\prime}_k\right)\otimes P_{\overline{W}\lvert W_k,S^{\prime}_k,S_k}$.
    \item the expectation in the RHS of \eqref{CMI:generalizationbound:KL:bounding:exponent:2}  is taken over the random variables $({\mathfrak{Z}}_k^{2n}, W_k, \overline{W}, \mathbf{J}_k)$, with the joint distribution given by $\mu^{\otimes 2n}_k \otimes {Q}_k(W_k\lvert{\mathfrak{Z}_{\mathbf{J}_k}^{2n}},\mathfrak{Z}_{\mathbf{J}^c_k}^{2n})\otimes P_{\overline{W}\lvert W_k,\mathfrak{Z}_{\mathbf{J}_k}^{2n}}\otimes \text{Bern}\left(\frac{1}{2}\right)^{\otimes n}$.
    \item the expectation in \eqref{CMI:generalizationbound:KL:bounding:exponent:3} is taken over the random variables $({\mathfrak{Z}}_k^{2n}, W_k, \overline{W})$, with the joint distribution described by $\mu^{\otimes 2n}_k \otimes {Q}(W_k|\mathfrak{Z}_k^{2n}) \otimes P(\overline{W}|W_k, \mathfrak{Z}_k^{2n})$.
    \item the conditionals ${Q}_k(W_k | S_k, S^{\prime}_k)$ and $P_{\overline{W} \mid W_k, S^{\prime}_k, S_k}$ are both symmetric with respect to $ S^{\prime}_k, S_k$ -- the symmetry of ${Q}_k(W_k | S_k, S^{\prime}_k)$ holds by assumption and that of $P_{\overline{W} \mid W_k, S^{\prime}_k, S_k}$ follows by use of Markov's chain $\overline{W} - W_k - (S_k, S^{\prime}_k)$. This implies the symmetry over joint distribution of ${Q}_k(W_k|S_k, S^{\prime}_k) \otimes P_{\overline{W} \mid W_k, S^{\prime}_k, S_k}$ with respect to $(S_k,S'_k)$; and, so, the RHS of~\eqref{CMI:generalizationbound:KL:bounding:exponent:2} and that of~\eqref{CMI:generalizationbound:KL:bounding:exponent:3} are identical.

     \item \eqref{CMI:generalizationbound:KL:bounding:exponent:4} follows by using the inequality 
     \begin{align}
         \frac{e^{x}+e^{-x}}{2}\leq e^{\frac{x^2}{2}},
     \end{align}
     and the fact that $\ell(z,w)\in[0,1]$ for all realization of $(z,w)\in(\mathcal{Z},\mathcal{W})$.
     \end{itemize}
     Continuing from~\eqref{CMI:generalizationbound:KL:2} and substuting using~\eqref{CMI:generalizationbound:KL:bounding:exponent:5} we get
     \begin{align}
\mathbb{E}_{S_{[K]},\overline{W}}\left[\gen(S_{[K]},\overline{W})\right]\leq\frac{1}{K\lambda}\sum_{k\in[K]}D_{KL}\left({\mu}^{\otimes 2n}_k\otimes P_{\overline{W},W_k\lvert S_k,S^{\prime}_k}\parallel{\mu}^{\otimes 2n}_k\otimes{Q}_k\otimes\overline{P}_k\right)+\frac{\lambda}{2n}.\label{simplification:Hoefding}
     \end{align}
This inequality can be further simplified as:
    \begin{align}
\mathbb{E}_{S_{[K]},\overline{W}}\left[\gen(S_{[K]},\overline{W})\right]&\leq\frac{1}{K\lambda}\sum_{k\in[K]}D_{KL}({\mu}^{\otimes 2n}_k\otimes P_{\overline{W},W_k\lvert S_k,S^{\prime}_k}\parallel{\mu}^{\otimes 2n}_k\otimes{Q}_k\otimes\overline{P}_k)+\frac{\lambda}{2n}\\
&=\frac{1}{K\lambda}\sum_{k\in[K]}D_{KL}({\mu}^{\otimes 2n}_k\otimes P_{W_k\lvert S_k,S^{\prime}_k}\otimes \overline{P}_k\parallel{\mu}^{\otimes 2n}_k\otimes{Q}_k\otimes\overline{P}_{k})+\frac{\lambda}{2n}\\
&=\frac{1}{K\lambda}\sum_{k\in[K]}D_{KL}\left({\mu}^{\otimes 2n}_k\otimes P_{W_k\lvert S_k,S^{\prime}_k}\parallel{\mu}^{\otimes 2n}_k\otimes{Q}_k\right)+\frac{\lambda}{2n}.
\label{CMI:generalizationbound:KL:bounding:exponent:6}
    \end{align}
Finally, letting
\begin{align}
    \lambda &= \sqrt{\frac{2n}{K} \sum_{k \in [K]} D_{KL} \left( {\mu}^{\otimes 2n}_k \otimes P_{W_k|S_k, S^{\prime}_k} \parallel {\mu}^{\otimes 2n}_k \otimes {Q}_k \right)}\\
    &=\sqrt{\frac{2n}{K} \sum_{k \in [K]} \mathbb{E}_{S_k,S'_k}\left[D_{KL} \left( P_{W_k|S_k, S^{\prime}_k} \parallel  {Q}_k \right)\right]},
\end{align}
and substuting in~\eqref{CMI:generalizationbound:KL:bounding:exponent:6} completes the proof of the second step; and so that of the theorem.




   \subsection{Proof of Theorem \ref{tailbound:cmi:general}}\label{Tail:proof:CMI}
   Let us consider the random variable \(\Delta(S_{[K]}, {Q}_{[K]})\) as
        \begin{align}
\Delta(S_{[K]},{Q}_{[K]})=\sqrt{\frac{\sum_{k\in[K]}\mathbb{E}_{S^{\prime}_{[K]}}\left[D_{KL}\left(P_{{W_k}\lvert S_k,S^{\prime}_k}\parallel{Q}_k\right)\right]+K\log\left(\sqrt{2n}\right)+\log(1/\delta)}{K\lambda^\ast}},\label{Tail:Eq:simplification:2}
        \end{align}
        and $$\lambda^{\ast}=\frac{(2n-1)}{4}.$$ Then, we can write

\begin{align}
            \mathbb{P}\Big(\mathbb{E}_{P_{\overline{W},W_k\lvert S_{[K]}}}&\left[\gen(S_{[K]},\overline{W})\right]>\Delta(S_{[K]},{Q}_{[K]})
            \Big)\label{Tail:Eq:simplification:3}\\
            &\qquad=\mathbb{P}\Big(\frac{1}{K}\sum\limits_{k\in[K]}\mathbb{E}_{S^{\prime}_k}\mathbb{E}_{P_{\overline{W},W_k\lvert S_k,S^{\prime}_k}}\left[\hat{\mathcal{L}}(S^{\prime}_k,\overline{W})-\hat{\mathcal{L}}(S_k,\overline{W})\right]>\Delta(S_{[K]},{Q}_{[K]})\Big)\label{Tail:Eq:simplification:4}\\
&\qquad\leq\mathbb{P}\Bigg(\bigg(\frac{1}{K}\sum\limits_{k\in[K]}\mathbb{E}_{S^{\prime}_k}\left[\mathbb{E}_{P_{\overline{W},W_k\lvert S^{\prime}_k,S_k}}\left[\hat{\mathcal{L}}(S^{\prime}_k,\overline{W})-\hat{\mathcal{L}}(S_k,\overline{W})\right]\right]\bigg)^2>{\Delta^2(S_{[K]},{Q}_{[K]})}\Bigg)\label{Tail:Eq:simplification:6}\\
&\qquad\leq\mathbb{P}\Bigg(\frac{1}{K}\sum_{k\in[K]}\left(\mathbb{E}_{S^{\prime}_k}\mathbb{E}_{P_{\overline{W},W_k\lvert S^{\prime}_k,S_k}}\left[\hat{\mathcal{L}}(S^{\prime}_k,\overline{W})-\hat{\mathcal{L}}(S_k,\overline{W})\right]\right)^{2}>\Delta^2(S_{[K]},{Q}_{[K]})\Bigg)\label{Tail:Eq:simplification:7}\\
&\qquad=\mathbb{P}\Bigg(\lambda^\ast\sum_{k\in[K]}\Bigg(\mathbb{E}_{S^{\prime}_k}\mathbb{E}_{P_{\overline{W},W_k\lvert S^{\prime}_k,S_k}}\left[\hat{\mathcal{L}}(S^{\prime}_k,\overline{W})-\hat{\mathcal{L}}(S_k,\overline{W})\right]\Bigg)^2>{\lambda^{\ast}K{\Delta^2(S_{[K]},{Q}_{[K]})}}\Bigg)\label{Tail:Eq:simplification:9}\\
&\qquad\leq \mathbb{P}\Bigg(\lambda^\ast\sum_{k\in[K]}\mathbb{E}_{S^{\prime}_k}\mathbb{E}_{P_{\overline{W},W_k\lvert S^{\prime}_k,S_k}}\left[\left(\hat{\mathcal{L}}(S^{\prime}_k,\overline{W})-\hat{\mathcal{L}}(S_k,\overline{W})\right)^2\right]>{\lambda^{\ast}K{\Delta^2(S_{[K]},{Q}_{[K]})}}\Bigg)\label{Tail:Eq:simplification:10}\\
&\qquad\leq\mathbb{P}\bigg(\sum\limits_{k\in[K]}\mathbb{E}_{S^{\prime}_k}\left(D_{KL}\left(P_{\overline{W},W_k\lvert S_k,S^{\prime}_k}\parallel\overline{P}_k\otimes{Q}_k \right)\right)\label{Tail:Eq:simplification:12}\\
&\qquad\quad\quad+\sum\limits_{k\in[K]}\mathbb{E}_{S^{\prime}_k}\bigg(\log\mathbb{E}_{\overline{P}_k\otimes{Q}_k}\left[e^{
\lambda^\ast\left(\hat{\mathcal{L}}(S^{\prime}_k,\overline{W})-\hat{\mathcal{L}}(S_k,\overline{W})\right)^2}\right]\bigg)\geq{\lambda^{\ast}K{\Delta^2(S_{[K]},{Q}_{[K]})}}\bigg)\label{Tail:Eq:simplification:13}\\
&\qquad\leq\mathbb{P}\bigg(\sum\limits_{k\in[K]}\mathbb{E}_{{\mu_k^{\otimes n}}}\left(\log\mathbb{E}_{\overline{P}_k\otimes{Q}_k}\left[e^{
\lambda^\ast\left(\hat{\mathcal{L}}(S^{\prime}_k,\overline{W})-\hat{\mathcal{L}}(S_k,\overline{W})\right)^2}\right]\right)\geq\label{Tail:Eq:simplification:14}\\
&\quad\qquad\qquad\sum_{k\in[K]}\log\mathbb{E}_{\mu^{\otimes 2n}_k\otimes\overline{P}_k\otimes{Q}_k}\Big[e^{
\lambda^\ast\left(\hat{\mathcal{L}}(S^{\prime}_k,\overline{W})-\hat{\mathcal{L}}(S_k,\overline{W})\right)^2}\Big]+\log(1/\delta)\bigg)\\
&\qquad\leq\delta\label{Tail:Eq:simplification:15},
        \end{align}
        where
        \begin{itemize}
        \item  $S_{[K]}$ is distributed as $S_{[K]}\sim\prod_{k=1}^{K}\mu^{\otimes n}_k$,
        \item the probability distributions \( P_{\overline{W}| W_k,S_k,S^{\prime}_k}\) and \({Q}_k(\overline{W}|S_k, S^{\prime}_k)\) are denoted by \(\overline{P}_k\) and \({Q}_k\), respectively, as before, and \eqref{Tail:Eq:simplification:10} are due to Jensen's inequality for the convex function $f(x)=x^2$,
\item equations (\ref{Tail:Eq:simplification:12}-\ref{Tail:Eq:simplification:13}) are concluded using Donsker-Varadhan's variational representation lemma,
\item and Markov inequality yields the final inequality in \eqref{Tail:Eq:simplification:15}.
        \end{itemize}
        It remains to show that 
        \begin{align}
\sum_{k\in[K]}\log\mathbb{E}_{\mu^{\otimes 2n}_k\otimes\overline{P}_k\otimes{Q}_k}\Big[e^{
\lambda^\ast\left(\hat{\mathcal{L}}(S^{\prime}_k,\overline{W})-\hat{\mathcal{L}}(S_k,\overline{W})\right)^2}\Big]
&\leq K\log(\sqrt{2n}), \label{Tail:Eq:simplification:17}
        \end{align}
        where expectation is with respect to the probability distribution $\mu^{2n}_k\otimes\overline{P}_k\otimes{Q}_k$. 
        
To show this, the left-hand side of \eqref{Tail:Eq:simplification:17} can be re-written as
        \begin{align}
\sum_{k\in[K]}\log\mathbb{E}_{\mu^{\otimes 2n}_k\otimes\overline{P}_k\otimes{Q}_k}\Big[e^{
\lambda^\ast\left(\hat{\mathcal{L}}(S^{\prime}_k,\overline{W})-\hat{\mathcal{L}}(S_k,\overline{W})\right)^2}\Big]
&=\sum\limits_{k\in[K]}\log\mathbb{E}\Big[e^{
\lambda^\ast\left(\frac{1}{n}\sum_{i\in[n]}\left[\ell (Z^{\prime}_{k,i},\overline{W})-\ell (Z_{k,i},\overline{W})\right]\right)^2}\Big] \label{Tail:Eq:simplification:18}\\
&=\sum\limits_{k\in[K]}\log\mathbb{E}\left[e^{
\lambda^\ast\left(\frac{1}{n}\sum_{i\in[n]}\left[\ell(\mathfrak{Z}_{J_{k,i}},\overline{W})-\ell(\mathfrak{Z}_{J^c_{k,i}},\overline{W})\right]\right)^2}\right]\label{Tail:Eq:simplification:19} \\
&=\sum\limits_{k\in[K]}\log\mathbb{E}\Bigg[\mathbb{E}_{\mathbf{J}_k}\left[e^{
\lambda^\ast\left(\frac{1}{n}\sum_{i\in[n]}\left(\ell\left[\mathfrak{Z}_{J_{k,i}},\overline{W})-\ell(\mathfrak{Z}_{J^c_{k,i}},\overline{W}\right]\right)\right)^2}\right]\Bigg]\label{Tail:Eq:simplification:20}\\
&\leq K\log(\sqrt{2n})\label{Tail:Eq:simplification:21},
        \end{align}
where
         \begin{itemize}
         \item the expectation in the right-hand side of \eqref{Tail:Eq:simplification:18} is with respect to the probability distribution ${\mu^{2n}_k\otimes\overline{P}_k\otimes{Q}_k}$,
         \item the expectation in \eqref{Tail:Eq:simplification:19} is with respect to ${\mu^{\otimes 2n}_k\otimes P_{\overline{W}\lvert W_k,\mathfrak{Z}_k^{2n}}\otimes{Q}_k({W_k}\lvert\mathfrak{Z}_k^{2n})\otimes\text{Bern}\left(\frac{1}{2}\right)^{\otimes n}}$, 
         \item equation \eqref{Tail:Eq:simplification:20} uses ${\mu^{\otimes 2n}_k\otimes{Q}_k({W_k}\lvert\mathfrak{Z}_k^{2n})\otimes P_{\overline{W}\lvert W_k,\mathfrak{Z}_k^{2n}}}$ as join distribution for computing the expectation,
             \item the equation \eqref{Tail:Eq:simplification:19} follows from the symmetry of \(P_{\overline{W} \mid W_k, S_k, S^{\prime}_k} \otimes Q_k(\overline{W}| S_k, S^{\prime}_k)\) with respect to \((S_k, S^{\prime}_k)\). The symmetry in \(P_{\overline{W}| W_k, S_k, S^{\prime}_k}\) arises from the Markov chain \(\overline{W} - W_k - (S_k, S^{\prime}_k)\) in \(P_{\overline{W}|W_k, S_k, S^{\prime}_k}\), and the symmetry in \(Q_k(\overline{W}|S_k, S^{\prime}_k)\) follows from the assumptions. These two separate symmetric properties together imply the symmetry of \(P_{\overline{W}|W_k, S_k, S^{\prime}_k} \otimes Q_k(\overline{W} \mid S_k, S^{\prime}_k)\).
\item the expecationf in equation \eqref{Tail:Eq:simplification:20} is computed with respect to random variable $\mathbf{J}_k\sim\text{Bern}\left(\frac{1}{2}\right)^{\otimes n}$ 

\item the equation \eqref{Tail:Eq:simplification:21} is concluded since 
\begin{align}
    \frac{1}{n}\sum_{i\in[n]}\left[\ell\left(\mathfrak{Z}_{J_{k,i}},\overline{W})-\ell(\mathfrak{Z}_{J^c_{k,i}},\overline{W}\right)\right],\nonumber
\end{align}
is $1/\sqrt{n}$-subgaussian for any $k\in[K]$ and hence 
\begin{align}
    \mathbb{E}_{\mathbf{J}_k\sim\text{Bern}\left(\frac{1}{2}\right)^{\otimes n}}\left[e^\frac{
\left[\frac{1}{n}\sum_{i\in[n]}\left(\ell(\mathfrak{Z}_{J_{k,i}},\overline{W})-\ell(\mathfrak{Z}_{J^c_{k,i}},\overline{W})\right)\right]^2}{(4/2n-1)}\right]\leq\sqrt{2n},\nonumber
\end{align}
         \end{itemize}
where concluded from \cite[Theorem 2.6.VI]{wainwright2019high} and this completes the proof.



\subsection{Proof of Theorem \ref{Lossy:inexpectation:general}}\label{Lossy:general:CMI:1}

We have
\begin{align}
    \mathbb{E}\left[\gen(S_{[K]}, \overline{W})\right] &= \frac{1}{K} \sum_{k \in [K]} \mathbb{E}\left[\gen(S_k, \overline{W})\right] \nonumber \\
    &\leq \frac{1}{K} \sum_{k \in [K]} \Big(\mathbb{E} \left[\gen(S_k, \hat{\overline{W}}_k)\right] + \epsilon\Big) \nonumber \\
    &\leq \sqrt{\frac{2 \sum_{k \in [K]} I(\hat{\overline{W}}_k; \mathbf{J}_k \mid \mathfrak{Z}^{2n}_k, W_{[K] \setminus k})}{nK}} + \epsilon, \label{Lossy:in:exp:proof}
\end{align}
where:
\begin{itemize}
\item the first equality follows by~\eqref{definition-generalization-error},
    \item the first inequality follows by the (distortion) constraint \( \mathbb{E}\left[\gen(S_k, \overline{W}) - \gen(S_k, \hat{\overline{W}}_k)\right] \leq \epsilon \),
    \item the second inequality
   holds by application of Theorem \ref{CMI:FL:TH},
    \item the mutual information is calculated according to the joint distribution \( P_{\mathfrak{Z}_k^{2n}, W_{[K] \setminus k}, \mathbf{J}_k} \times P_{\hat{\overline{W}}_k \mid \mathfrak{Z}_k^{2n}, W_{[K] \setminus k}, \mathbf{J}_k} \), so by taking the infimum over all conditional distributions \( P_{\hat{\overline{W}}_k \mid \mathfrak{Z}_k^{2n}, W_{[K] \setminus k}, \mathbf{J}_k} \), the proof will then be completed.
\end{itemize}















\subsection{Proof of Theorem \ref{h_D:inexpectation:general}}\label{h_D:inexpectation:general:proof}

The proof consists of two steps. In the first step, we use the equivalence between the two terms in \eqref{general:h_d:1} and \eqref{general:h_d:2}, which has already been proved by Theorem \ref{CMI:FL:TH} and is therefore omitted. In the second step, we establish the main part of the Theorem. We have  
    \begin{align}
      nh_{D}&\Big(\mathbb{E}_{\overline{W}}\big[\mathcal{L}(\overline{W})\big],\mathbb{E}_{S_{[K]},\overline{W}}\big[{\hat{\mathcal{L}}(S_{[K]},\overline{W})}\big]\Big)\\
      &\leq\frac{n}{K}\sum_{k\in[K]}\mathbb{E}_{S^{\prime}_k,S_k,\overline{W}}\Big[h_{D}\Big(\hat{\mathcal{L}}\big(S^{\prime}_k,\overline{W}\big),{\hat{\mathcal{L}}\big(S_k,\overline{W}\big)}\Big)\Big]\\
      &=\frac{n}{K}\sum_{k\in[K]}\mathbb{E}_{S^{\prime}_k,S_k,W_k,\overline{W}}\left[h_{D}\left(\hat{\mathcal{L}}\big(S^{\prime}_k,\overline{W}\big),{\hat{\mathcal{L}}\big(S_k,\overline{W}\big)}\right)\right]\\
      &=\frac{n}{K}\sum_{k\in[K]}\mathbb{E}_{S^{\prime}_k,S_k,W_k,\overline{W}}\Big[h_D\Big(\frac{1}{n}\sum\nolimits_{i\in[n]}\ell(Z^{\prime}_{i,k},\overline{W}),\frac{1}{n}\sum\nolimits_{i\in[n]}\ell(Z_{i,k},\overline{W})\Big)\Big]\\
      &\leq\frac{1}{K}\sum_{k\in[K]}D_{KL}\left(\mu^{\otimes 2n}_kP_{\overline{W},W_k\lvert S_k,S^{\prime}_k}\parallel \mu^{\otimes 2n}_k\otimes Q_k\otimes\overline{P}_k\right)\label{CMI:hd:bound:expectation:2}\\
      &\qquad\quad+\sum_{k\in[K]}\frac{1}{K}\log\mathbb{E}_{S_k,S^{\prime}_k,\overline{W}\sim\mu^{\otimes 2n}_k\otimes Q_k\otimes\overline{P}_k}\left[e^{nh_D\left(\frac{1}{n}\sum_{i\in[n]}\ell(Z^{\prime}_{i,k},\overline{W}),\frac{1}{n}\sum_{i\in[n]}\ell(Z_{i,k},\overline{W})\right)}\right],\label{CMI:hd:bound:expectation:1}
     \end{align}
     where briefly ${Q}\left({W_k}\lvert S^{\prime}_k,S_k\right)\otimes P_{\overline{W}\lvert W_k,S^{\prime}_k,S_k}$ is denoted by $Q_k\otimes\overline{P}_k$.

    In the following, we compute the term in the equation \eqref{CMI:hd:bound:expectation:1}. We use $\unif(2n)$ as a distribution that picks uniformly $n$ indices among $2n$ indices with probability $\frac{1}{\binom{2n}{n}}$. This indices will be denote by $\mathbf{T}_k=(T_{k,1},\cdots,T_{k,n})$, and therefore for a vector $\mathfrak{Z}^{2n}$  with length $2n$, where rearranged by combining such that $\{\mathfrak{Z}^{2n}_k\}:=\{S_k\}\bigcup\{{S^{\prime}_k}\}$, therefore the elements corresponds to $n$ indices of $S_k$ will be in $\mathbf{T}_k$ and denote by $\mathfrak{Z}^{2n}_{\mathbf{T}_k}=(\mathfrak{Z}_{T_{k,1}},\cdots,\mathfrak{Z}_{T_{k,n}})$. The other $n$ indices are allocated to $S^{\prime}_k$,  they are not in  $\mathbf{T}_k$, and they denote by ${\mathbf{T}_k^{c}}=({T^c_{k,1}},\cdots,{T^c_{k,n}})$ and similarly its corresponds vector in $\mathfrak{Z}_k^{2n}$ will be denote by $\mathfrak{Z}^{2n}_{\mathbf{T}_k^c}=(\mathfrak{Z}_{T^c_{k,1}},\cdots,\mathfrak{Z}_{T^c_{k,n}})$. Therefore using Markov chain $\overline{W}-W_k-(S_k,S^{\prime}_k)$ we have symmetry on $P_{\overline{W}\lvert W_k,S_K,S^{\prime}_k}$ respect to $(S_k,S^{\prime}_k)$, So using Lemma \ref{simp:expect:loss:1}, it can be concluded that for $n\geq 10$ and all $k\in[K]$,
    \begin{align}
\mathbb{E}_{\mu^{\otimes2n}_k\otimes Q_k\otimes\overline{P}_k}&\left[e^{nh_D\left(\frac{1}{n}\sum_{i\in[n]}\ell(Z^{\prime}_{i,k},\overline{W}),\frac{1}{n}\sum_{i\in[n]}\ell(Z_{i,k},\overline{W})\right)}\right]\label{simplification:expect:2}\\
&=\mathbb{E}_{\mu^{\otimes 2n}_k\otimes{Q}_{T_k}\otimes\overline{P}_{T_k}\otimes\unif(2n)}\left[e^{nh_D\left(\frac{1}{n}\sum_{i\in[n]}\ell(\mathfrak{Z}_{T^c_{k,i}},\overline{W}),\frac{1}{n}\sum_{i\in[n]}\ell(\mathfrak{Z}_{T_{k,i}},\overline{W})\right)}\right]\label{simplification:h_d:expect:3}\\
&=\mathbb{E}_{\unif(2n)}\left[\mathbb{E}_{\mu^{\otimes 2n}_k\otimes{Q}_{T_k}\otimes\overline{P}_{T_k}}\left(e^{nh_D\left(\frac{1}{n}\sum_{i\in[n]}\ell(\mathfrak{Z}_{T^c_{k,i}},\overline{W}),\frac{1}{n}\sum_{i\in[n]}\ell(\mathfrak{Z}_{T_{k,i}},\overline{W})\right)}\right)\right]\label{simplification:h_d:expect:4}\\
&=\mathbb{E}_{\mu^{\otimes 2n}_k\otimes Q_k(W_k\lvert\mathfrak{Z}^{2n}_k)\otimes P_{\overline{W}\lvert W_k,\mathfrak{Z}^{2n}}}\left[\mathbb{E}_{\mathbf{T}_k\sim\unif(2n)}\left(e^{nh_D\left(\frac{1}{n}\sum_{i\in[n]}\ell(\mathfrak{Z}_{T^c_{k,i}},\overline{W}),\frac{1}{n}\sum_{i\in[n]}\ell(\mathfrak{Z}_{T_{k,i}},\overline{W})\right)}\right)\right]\label{simplification:h_d:expect:5}\\
&\leq n,\label{simplification:h_d:expect:6}
    \end{align}
    \begin{itemize}
    \item $Q_k(W_k\lvert S_k,S^{\prime}_k)\otimes P_{\overline{W}\lvert W_k,S_k,S^{\prime}_k}$ and ${Q}(W_k\lvert{\mathfrak{Z}}_{\mathbf{T}_k}^{\otimes 2n},\mathfrak{Z}^{\otimes 2n}_{{\mathbf{T}^c_k}})\otimes P_{\overline{W}\lvert W_k,{Z}_{\mathbf{T}_k}^{\otimes2n},{Z}_{\mathbf{T}^c_k}^{\otimes2n}}$
    briefly denoted by $Q_k\otimes\overline{P}_k$ and ${Q}_{T_k}\otimes\overline{P}_{T_k}$ respectively.
        \item the equivalency between (\ref{simplification:h_d:expect:3}-\ref{simplification:h_d:expect:5}) comes from symmetry of $Q_k$ respect to $S_k,S^{\prime}_k$ combining with Markov chain $\overline{W}-W_k-(S_k,S^{\prime}_k)$.
        \item and the equation \eqref{simplification:h_d:expect:6} concluded from Lemma \ref{simp:expect:loss:1}.
    \end{itemize}




\subsection{Proof of Theorem \ref{h_d:ingeneral}}\label{Lossy:h_d:inverse}

To prove this result, in the first step, we establish the following bound:
\begin{align*}
    nh_D\left(\frac{1}{K}\sum_{k\in[K]}\mathbb{E}_{{\hat{\overline{W}}_k}}\left[\mathcal{L}(\hat{\overline{W}}_k)\right],\frac{1}{K}\sum_{k\in[K]}\mathbb{E}_{S_k,\hat{\overline{W}}_k}\left[\hat{\mathcal{L}}(S_k,\hat{\overline{W}}_k)\right]\right) 
    \leq \frac{1}{K} \sum_{k\in[K]} I(\mathbf{T}_k, \hat{\overline{W}}_k \mid \mathfrak{Z}^{2n}_k, W_{[K] \setminus \{k\}}) + \log n.
\end{align*}
We then utilize the distortion criterion and the definition of the inverse function \( h^{-1}_D \) to complete the proof. 

To show the first step, we have  
\begin{align}
&nh_D\left(\frac{1}{K}\sum_{k\in[K]}\mathbb{E}_{{\hat{\overline{W}}_k}}\left[\mathcal{L}(\hat{\overline{W}}_k)\right],\frac{1}{K}\sum_{k\in[K]}\mathbb{E}_{S_k,\hat{\overline{W}}_k}\left[\hat{\mathcal{L}}(S_k,\hat{\overline{W}}_k)\right]\right)\nonumber\\
    &=nh_D\left(\frac{1}{K}\sum_{k\in[K]}\mathbb{E}_{{S_k,W_{[K]\setminus{k}}}}\left[\mathbb{E}_{{\hat{\overline{W}}_k|S_k,W_{[K]\setminus{k}}}}\left[{\mathcal{L}}(\hat{\overline{W}}_k)\right]\right],\frac{1}{K}\sum_{k\in[K]}\mathbb{E}_{{S_k,W_{[K]\setminus{k}}}}\left[\mathbb{E}_{{\hat{\overline{W}}_k|S_k,W_{[K]\setminus{k}}}}\left[\hat{\mathcal{L}}(S_k,\hat{\overline{W}}_k)\right]\right]\right)\nonumber\\
    &\leq\frac{n}{K}\sum_{k\in[K]}\mathbb{E}_{{S_k,W_{[K]\setminus{k}}}}\left[h_D\left(\mathbb{E}_{{\hat{\overline{W}}_k|S_k,W_{[K]\setminus{k}}}}\left[{\mathcal{L}}(\hat{\overline{W}}_k)\right],\mathbb{E}_{{\hat{\overline{W}}_k|S_k,W_{[K]\setminus{k}}}}\left[\hat{\mathcal{L}}(S_k,\hat{\overline{W}}_k)\right]\right)\right]\label{Proof:lossy:h_d:3}\\
    &\leq\frac{1}{K}\sum_{k\in[K]}I(\mathbf{T}_k,\hat{\overline{W}}_k|\mathfrak{Z}^{2n}_k,W_{[K]\setminus{k}})+\log n\label{Proof:lossy:h_d:4},
\end{align}
where 
\begin{itemize}
    \item equation \eqref{Proof:lossy:h_d:3} is due to the convexity of \(h_D\) with respect to both inputs \cite[Lemma~1]{sefidgaran2024minimum},
    \item and \eqref{Proof:lossy:h_d:4} holds due to Theorem \ref{h_D:inexpectation:general},
\end{itemize}
Next, recall that by the assumption of the theorem, we have  $P_{\hat{\overline{W}}_k|S_k,W_{[K]\setminus{k}}}$ satisfies the distortion criterion 
\begin{align}
    \frac{1}{K}\sum_{k\in[K]}\mathbb{E}\left[\gen(S_{k},\overline{W})-\gen(S_{k},\hat{\overline{{W}}}_k)\right]\leq\epsilon.
\end{align}
Hence, using this criterion, the expected generalization error can be upper-bounded as
\begin{align}
  \mathbb{E}\left[\gen(S_{[K]},\overline{W})\right]&\leq\frac{1}{K}\sum_{k\in[K]}\mathbb{E}\left[\gen(S_{k},\hat{\overline{{W}}}_k)\right]+\epsilon\label{Proof:lossy:h_d:5}\\
  &=\frac{1}{K}\sum_{k\in[K]}\left(\mathbb{E}_{{\hat{\overline{W}}_k}}\left[\mathcal{L}(\hat{\overline{W}}_k)\right]-\mathbb{E}_{S_k,\hat{\overline{W}}_k}\left[\hat{\mathcal{L}}(S_k,\hat{\overline{W}}_k)\right]\right)+\epsilon\label{Proof:lossy:h_d:6}\\
  &\leq h^{-1}_D\left(\frac{1}{nK}\sum_{k\in[K]}I(\mathbf{T}_k,\hat{\overline{W}}_k|\mathfrak{Z}^{2n}_k,W_{[K]\setminus{k}})+\log n\Bigg|\frac{1}{K}\sum_{k\in[K]}\mathbb{E}_{S_k,\hat{\overline{W}}_k}\left[\hat{\mathcal{L}}(S_k,\hat{\overline{W}}_k)\right]\right)\label{Proof:lossy:h_d:7}\\
  &\quad-\frac{1}{K}\sum_{k\in[K]}\mathbb{E}_{S_k,\hat{\overline{W}}_k}\left[\hat{\mathcal{L}}(S_k,\hat{\overline{W}}_k)\right]+\epsilon\label{Proof:lossy:h_d:8},
\end{align}
where equation \eqref{Proof:lossy:h_d:7} derived from 
\begin{align}
    \frac{1}{K}\sum_{k\in[K]}\mathbb{E}_{{\hat{\overline{W}}_k}}\left[\mathcal{L}(\hat{\overline{W}}_k)\right]\leq h^{-1}_D\left(\frac{1}{nK}\sum_{k\in[K]}I(\mathbf{T}_k,\hat{\overline{W}}_k|\mathfrak{Z}^{2n}_k,W_{[K]\setminus{k}})+\log n\Bigg|\frac{1}{K}\sum_{k\in[K]}\mathbb{E}_{S_k,\hat{\overline{W}}_k}\left[\hat{\mathcal{L}}(S_k,\hat{\overline{W}}_k)\right]\right),
\end{align}
using \eqref{Proof:lossy:h_d:4} and the definition of the inverse function $h^{-1}_D(\cdot|\cdot)$. This completes the proof.

\subsection{Proof of Theorem \ref{h_d:tail:general}}\label{Tail:h_d}
    Let define 
    \begin{align}
        \Delta(S_{[K]},{Q}_{[K]})=\frac{1}{K}\sum_{k\in[K]}D_{KL}\left(P_{W_k\lvert S_k,S^{\prime}_k}\parallel{Q}_{k}\right)+\log( n/\delta).
    \end{align}

    We have
        \begin{align}
&\mathbb{P}\left(nh_{D}\left({\hat{\mathcal{L}}(S^{\prime}_{[K]},\overline{W})},\hat{\mathcal{L}}(S_{[K]},\overline{W})\right)>\Delta(S_{[K]},{Q}_{[K]})\right)\nonumber\\
&\leq\mathbb{P}\Bigg(n\sum_{k\in[K]}\frac{1}{K}h_{D}\Big(\mathbb{E}_{P_{\overline{W}\lvert S_k}}\Big[n^{-1}\sum\nolimits_{i\in[n]}\ell\left(Z^{\prime}_{i,k},\overline{W}\right)\Big],\mathbb{E}_{P_{\overline{W}\lvert S_k}}\left[n^{-1}\sum\nolimits_{i\in[n]}\ell\left(Z_{i,k},\overline{W}\right)\right]\Big)>\Delta(S_{[K]},{Q}_{[K]})\Bigg)\label{hd:expectation:8}\\
&=\mathbb{P}\Bigg(\sum_{k\in[K]}\frac{n}{K}h_{D}\left(\mathbb{E}_{P_{\overline{W},W_k\lvert S_k}}\left[n^{-1}\sum\nolimits_{i\in[n]}\ell\left(Z^{\prime}_{i,k},\overline{W}\right)\right],\mathbb{E}_{P_{\overline{W},W_k\lvert S_k}}\left[n^{-1}\sum\nolimits_{i\in[n]}\ell\left(Z_{i,k},\overline{W}\right)\right]\right)>\Delta(S_{[K]},{Q}_{[K]})\Bigg)\nonumber\\
&=\mathbb{P}\Bigg(\sum_{k\in[K]}\frac{n}{K}h_{D}\left(\mathbb{E}_{P_{\overline{W},W_k\lvert S_k,S^{\prime}_k}}\left[\frac{\sum\nolimits_{i\in[n]}\ell\big(Z^{\prime}_{i,k},\overline{W}\big)}{n}\right],\mathbb{E}_{P_{\overline{W},W_k\lvert S_k,S^{\prime}_k}}\left[\frac{\sum\nolimits_{i\in[n]}\ell\left(Z_{i,k},\overline{W}\right)}{n}\right]\right)>\Delta(S_{[K]},Q_{[K]})\Bigg)\nonumber\\
&\leq\mathbb{P}\Bigg(\sum\nolimits_{k\in[K]}\frac{n}{K}\mathbb{E}_{P_{\overline{W},W_k\lvert S_k,S^{\prime}_k}}\left[h_{D}\left(\frac{\sum\nolimits_{i\in [n]}\ell(Z^{\prime}_{i,k},\overline{W})}{n},\frac{\sum\nolimits_{i\in [n]}\ell(Z_{i,k},\overline{W})}{n}\right)\right]>\Delta(S_{[K]},{Q}_{[K]})\Bigg)\label{hd:expectation:5}\\
&\leq\mathbb{P}\Bigg(\sum\nolimits_{k\in[K]}\frac{1}{K}D_{KL}\left(P_{\overline{W},W_k\lvert S_k,S^{\prime}_k}\parallel \overline{P}_k\otimes Q_k\right)\label{hd:expectation:4}\\
&\qquad+\sum\nolimits_{k\in[K]}\frac{1}{K}\log\mathbb{E}_{\overline{P}_k\otimes Q_k}\left[e^{\left(nh_{D}\left(\frac{1}{n}\sum\nolimits_{i\in [n]}\ell(Z^{\prime}_{i,k},\overline{W}),\frac{1}{n}\sum\nolimits_{i\in [n]}\ell(Z_{i,k},\overline{W})\right)\right)}\right]>\Delta(S_{[K]},{Q}_{[K]})\Bigg)\nonumber\\
&\leq\mathbb{P}\Bigg(\sum\nolimits_{k\in[K]}\frac{1}{K}\log\mathbb{E}_{\mu^{\otimes 2n}_k\otimes\overline{P}_k\otimes Q_k}\left[e^{\left[nh_{D}\left(\frac{1}{n}\sum\nolimits_{i\in [n]}\ell(Z^{\prime}_{i,k},\overline{W}),\frac{1}{n}\sum\nolimits_{i\in [n]}\ell(Z_{i,k},\overline{W})\right)\right]}\right]>\nonumber\\
&\qquad\quad\sum\nolimits_{k\in[K]}\frac{1}{K}\log\mathbb{E}_{\mu^{\otimes 2n}_k\otimes\overline{P}_k\otimes Q_k}\left[e^{\left[nh_{D}\left(\frac{1}{n}\sum_{i\in [n]}\ell(Z^{\prime}_{i,k},\overline{W}),\frac{1}{n}\sum_{i\in [n]}\ell(Z_{i,k},\overline{W})\right)\right]}\right]+\log(1/\delta)
\Bigg)\nonumber\\
&\leq\delta\label{hd:expectation:1}
\end{align}
        \begin{itemize}
        \item $\mathbb{P}(\cdot)$ is calculated respect to variables
    ${(S_{[K]},S^{\prime}_{[K]})\sim\prod_{k=1}^{K}\mu^{\otimes 2n}_k}$ in all of the above steps.
        \item $P_{\overline{W}\lvert W_k,S_k,S^{\prime}_k}\otimes Q_k$ is denoted by $\overline{P}_k\otimes Q_k$ for simplicity.
            \item equations \eqref{hd:expectation:8} and \eqref{hd:expectation:5} concluded from convexity of $h_D(x_1,x_2)$ in both of $x_1$ and $x_2$.
            \item Donsker-Varadhan variational representation implies the equation \eqref{hd:expectation:4}.
            \item and the \eqref{hd:expectation:1} is due to Markov's inequality. 
        \end{itemize}
       Now it is just enough to compute the
       \begin{align}
\mathbb{E}_{\mu^{\otimes 2n}_k\otimes\overline{P}_k\otimes Q_k }\left[e^{\left(nh_{D}\left(\frac{1}{n}\sum_{i\in [n]}\ell(Z^{\prime}_{i,k},\overline{W}),\frac{1}{n}\sum_{i\in[n]}\ell(Z_{i,k},\overline{W})\right)\right)}\right]\leq n,
       \end{align}
       for any $k\in[K]$, where already computed in \eqref{simplification:h_d:expect:6}
       And this completes the proof.


\subsection{Proof of Theorem~\ref{general:svm:cmi}}\label{Proof:SVM:General:K}
Fix some $r\in[M]$. In the rest of the proof, for better readability, we drop the dependence of the parameters on $r$, for example, we use the notations
\begin{align}
    \mu_k&\coloneqq  \mu_k^{(r)}\label{parametes_proof:general:svm:limited:rad:1}  \\
    b_k &\coloneqq   b^{(r)}_k{=}{\sum\nolimits_{m=c_k^{(r)}}^{c_k^{(r)}+r-1}\alpha^{(r)}_{k,m}a_m},\label{parametes_proof:general:svm:limited:rad:2}\\
    \rho_k &\coloneqq  \rho_k^{(r)} = \rho+{D_{k,r}}\label{parametes_proof:general:svm:limited:rad:3}  \\
    c_k &\coloneqq  c_k^{(r)}\label{parametes_proof:general:svm:4}\\
    \alpha^{(r)}_{k,m}&\coloneqq\alpha_{k,m}\label{parametes_proof:general:svm:5}
\end{align}
We prove this result using Theorem~\ref{Lossy:inexpectation:general}. To use Theorem~\ref{Lossy:inexpectation:general}, first, we need to define the space of ``auxiliary'' or ``lossy'' hypotheses. Let $\hat{\mathcal{W}}= \mathbb{R}^d \times \mathbb{R}$ be the space of auxiliary hypotheses. Every hypothesis $\hat{w}\in \hat{\mathcal{W}}$ is composed of two parts $\hat{w}=(\hat{w}_1,\hat{w}_2)$, where $\hat{w}_1 \in \mathbb{R}^d$ and $\hat{w}_2 \in \mathbb{R}$. 

Next, for every $k\in[K]$, define the auxiliary loss function $\tilde{\ell}_{\theta,k} \colon \mathcal{Z} \times \hat{\mathcal{W}} \to \{0,1\}$ as
\begin{align}\label{auxiliary:loss:general}
    \tilde{\ell}_{\theta,k}(z_k,\hat{w}) = \mathbbm{1}_{\left\{y_k\left(\langle x_k-b_k,\hat{w}_1\rangle+\hat{w}_2\right)<\frac{\theta}{2}\right\}},\quad \hat{w}\in \hat{\mathcal{W}}, z_k\in \mathcal{Z}.
\end{align}
For a given $\hat{w}\in \hat{\mathcal{W}}$ and $s_k=\{z_{k,1},\ldots,z_{k,n}\}$, define the generlization error $\gen(s_{k},\hat{w})$ with respect to this auxiliary loss function, \ie
\begin{align}
    \gen(s_{k},\hat{w}) = \mathbb{E}_{Z_k\sim \mu_k}\left[ \tilde{\ell}_{\theta,k}(Z_k,\hat{w})\right] - \frac{1}{n} \sum_{i\in[n]} \tilde{\ell}_{\theta,k}(z_{i,k},\hat{w}). \label{eq:svm_pr_gen}
\end{align}

Now, we are ready to present the outline of the proof using Theorem~\ref{Lossy:inexpectation:general}. First, for each client $k\in[K]$, we  define the auxiliary learning algorithm $P_{\hat{\overline{W}}_k | S_k, W_{[K]\setminus{k}}}$ such that 
\begin{align}
\mathbb{E}\left[\gen_{\theta}(S_{k},\overline{W})-\gen(S_{k},\hat{\overline{{W}}}_k)\right]\leq\mathcal{O}\left(\frac{1}{nK\sqrt{K}}\right),
\label{eq:svm_pr_dist}
\end{align}
where $\gen(S_{k},\hat{\overline{{W}}}_k)$ is defined as in \eqref{eq:svm_pr_gen}.

Next, we show that for these auxiliary learning algorithms,
\begin{align}
    R_{\mathcal{D}_k}(\epsilon) \leq \mathcal{O}\left(\Big(\frac
    {\rho_{k}}{K\theta}\Big)^2\log(nK)\log\left(\left[3,\frac{K\theta}{\rho_k}\right]^+\right){+}{\log}\bigg(\bigg[1,\frac{4 n\| b_k\|}{K\theta}\bigg]^+\right). \label{eq:svm_pr_rate}
\end{align} 
Combining \eqref{eq:svm_pr_dist} and \eqref{eq:svm_pr_rate} with Theorem~\ref{Lossy:inexpectation:general} yield
    \begin{equation*}
        \mathbb{E}[\text{gen}_{\theta}(S_{[K]},\overline{W})] = \mathcal{O}\left(\sqrt{\frac{\sum\limits_{k\in[K]}\Big(\frac
    {\rho_{k}}{K\theta}\Big)^2\log(nK)\log\left(\left[3,\frac{K\theta}{\rho_k}\right]^+\right){+}{\log}\bigg(\bigg[1,\frac{4 n\| b_k\|}{K\theta}\bigg]^+\bigg)}{nK}}\right),
        \end{equation*}
which completes the proof.

Hence, we start by defining the auxiliary learning algorithms $P_{\hat{\overline{W}}_k | S_k, W_{[K]\setminus{k}}}$ and then we upper bound the distortion and ``rates'' as in \eqref{eq:svm_pr_dist} and \eqref{eq:svm_pr_rate}, respectively, to complete the proof. 

In the rest of the proof, we use the following constants:
 \begin{align}
           &m_k\coloneqq\left\lceil 112\left(\frac{\rho_{k}}{K\theta_n}\right)^2\log (nK\sqrt{K})\right\rceil,\label{def_constants_gaussian_1}\\ &\tau_{k,1}=\tau_{k,2}\coloneqq\sqrt{1+\frac{K{\theta_n}}{4\rho_{k}}},\\
           &\nu_k\coloneqq\frac{1}{2\tau_{k,1}},\\
           & \theta_n \coloneqq \theta\left(1-\frac{1}{n}\right), \label{def:constants}
\end{align}
where $\lceil\cdot \rceil$ denotes the ceiling function.

\paragraph{Definition of the auxiliary learning algorithm.} To define $P_{\hat{\overline{W}}_k | S_k, W_{[K]\setminus{k}}}$, first we define $P_{\hat{\overline{W}}_k |  W_{[K]}}$. Then, we let
\begin{align}
P_{\hat{\overline{W}}_k | S_k, W_{[K]}} = P_{\hat{\overline{W}}_k |  W_{[K]}}, \label{eq:svm_pr_w_hat_def1}
\end{align}
\ie we define $P_{\hat{\overline{W}}_k | S_k, W_{[K]}}$ by imposing the Markov chain $\hat{\overline{W}}_k - W_{[K]} - S_k$. Having defined $P_{\hat{\overline{W}}_k | S_k, W_{[K]}}$ and since $P_{W_k | S_k, W_{[K]\setminus k}}$ is already defined, the joint conditional distribution $P_{\hat{\overline{W}}_k, W_k | S_k, W_{[K]\setminus{k}}}$ is well defined, and hence so does the conditional distribution $P_{\hat{\overline{W}}_k | S_k, W_{[K]\setminus{k}}}$.

Hence, we need to define $P_{\hat{\overline{W}}_k | W_{[K]}}$. Given $W_{[K]}=w_{[K]}$, let 
\begin{align}
    \hat{\overline{W}}_k = \left(\hat{\overline{W}}_{k,1},\hat{\overline{W}}_{k,2}\right), \label{eq:svm_pr_w_hat_def2}
\end{align}
where $\hat{\overline{W}}_{k,1}$ and $\hat{\overline{W}}_{k,2}$ are defined as in the following.

\textbf{Definition of $\hat{\overline{W}}_{k,1}$:} For a fixed matrix $A_k\in\mathbb{R}^{m_k\times d}$, that will be determined later, let
\begin{align}
    \hat{\overline{W}}_{k,1} = \frac{1}{K} \sum_{k'\neq k} w_{k'} + \frac{1}{K} A_k^{\top} W'_{k,1}, \label{def:svm_pr_w_hat_1}
\end{align}
where $W'_{k,1} \in \mathbb{R}^{m_k}$ is a random variable distributed uniformly over the $m_k$-dimensonal ball $B_{m_k}(A_k w_k,\nu_k)$, if $\|A_k w_k\|\leq \tau_{k,2}$, and otherwise  distributed uniformly over the $m_k$-dimensonal ball $B_{m_k}(\vc{0},\nu_k)$. To summarize,
 \begin{align}
            W^\prime_{k,1}\sim\begin{cases}
                \unif(B_{m_k}(A_k w_k,\nu_k)), & \text{if } \| A_k w_k\|\leq \tau_{k,2},\\
                \unif(B_{m_k}(0,\nu_k)), &\quad \text{otherwise}.
            \end{cases} \label{def:svm_pr_w_p_1}
\end{align}

\textbf{Definition of $\hat{\overline{W}}_{k,2}$:} Let 
\begin{align}
    N_k \coloneqq \left\lceil \frac{4 n \|b_k\|}{K\theta} \right\rceil,
\end{align}
and
\begin{align}
     u_{k,t} = & -\frac{b_k}{K} + \frac{2t \|b_k\|}{K N_k}, \quad t \in [N_k].
\end{align}
Hence $u_{k,1}=-\frac{\|b_k\|}{K}+\frac{2 \|b_k\|}{N_k}$, $u_{k,N_k}=\frac{\|b_k\|}{K}$, and $u_{k,t}$ are chosen with distance at most $\frac{\theta}{2n}$.

Now, given $w_{[K]}$ and having defined $u_{k,t}$, $t\in[N_k]$, we choose $\hat{\overline{W}}_{k,2}$ as a deterministic discrete random variable taking the value
\begin{align}
   \hat{\overline{W}}_{k,2} =   \frac{1}{K} \sum_{k'\neq k} \left\langle b_k , w_{k'}\right\rangle+w^\prime_{k,2}, \label{def:svm_pr_w_hat_2}
\end{align}
where $w^\prime_{k,2}=u_{k,t^\ast}$ is a deterministic discrete random variable, where
\begin{align}
            t^\ast=\argmin_{t \in [N_k]}\left|\frac{1}{K} \langle b_k,w_k\rangle-u_{k,t}\right|.\label{def:svm_pr_w_p_2}
\end{align}
This completes the definition of $\hat{\overline{W}}_k$ for a given $w_{[K]}$. Hence, as explained above, this well defines the auxiliary learning algorithms $P_{\hat{\overline{W}}_k | S_k, W_{[K]\setminus{k}}}$. It remains then to prove the upper bounds \eqref{eq:svm_pr_dist} and \eqref{eq:svm_pr_rate} on the distortion and rates, respectively. 

\paragraph{Upper bounding the distortion:} In this part, for the above-defined auxiliary learning algorithm $P_{\hat{\overline{W}}_k | S_k, W_{[K]\setminus{k}}}$, we upper bound the distortion term as
\begin{align}
\mathbb{E}\left[\gen_{\theta}(S_{k},\overline{W})-\gen(S_{k},\hat{\overline{{W}}}_k)\right]\leq\mathcal{O}\left(\frac{1}{nK\sqrt{K}}\right), \label{eq:svm_pr_dist_rep}
\end{align}
where $\gen(S_{k},\hat{\overline{{W}}}_k)$ is defined as in \eqref{eq:svm_pr_gen}.

Recall that for a given $z\in \mathcal{Z}$ and the above-defined $\hat{\overline{W}}_k$ (see \eqref{def:svm_pr_w_hat_1} and \eqref{def:svm_pr_w_hat_2}), we have
\begin{align}
    \tilde{\ell}_{\theta}(z_k,\hat{\overline{W}}_k) = &  \mathbbm{1}_{\left\{y_k\left(\langle x_k-b_k,\hat{\overline{W}}_{k,1}\rangle+\hat{\overline{W}}_{k,2}\right)<\frac{\theta}{2}\right\}}.
\end{align}

To show \eqref{eq:svm_pr_dist_rep}, we first show that
\begin{align}
\mathbb{E}_{A_k}\mathbb{E}_{S_{k},\overline{W},\hat{\overline{{W}}}_k}\left[\gen_{\theta}(S_{k},\overline{W})-\gen(S_{k},\hat{\overline{{W}}}_k)\right]\leq\mathcal{O}\left(\frac{1}{nK\sqrt{K}}\right), \label{eq:svm_k2_het_dist_2}
\end{align}
Here the outer expectation is concerning random matrices $A_k\in\mathbb{R}^{m_k\times d}$ whose elements are generated i.i.d.. according to the distribution $\mathcal{N}(0,\frac{1}{m_k})$. Once \eqref{eq:svm_k2_het_dist_2} is shown, since the expectation over $A_k$ is upper bounded as desired; then this implies that there exists at least one suitable choice of $A_k$, for which  \eqref{eq:svm_pr_dist_rep} holds. This completes the proof of the upper bound on the distortion for this suitable choice of $A_k$.

Now, a sufficient condition to show \eqref{eq:svm_k2_het_dist_2} is to prove that for any fixed $(s_{[K]},w_{[K]})$ and for $\overline{w}=\frac{1}{K}\sum\limits_{k\in[K]} w_{k}$, it holds that\footnote{Recall that by definition $P_{{\hat{\overline{W}}_k}|s_k,w_{[K]}}=P_{{{\hat{\overline{W}}_k}|w_{[K]}}}$.}
\begin{align}
\mathbb{E}_{A_k}\mathbb{E}_{\hat{\overline{{W}}}_k \sim P_{{\hat{\overline{W}}}_k|w_{[K]},A_k}}\left[\gen_{\theta}(s_{k},\overline{w})-\gen(s_{k},\hat{\overline{{W}}}_k)\right]\leq\mathcal{O}\left(\frac{1}{nK\sqrt{K}}\right). \label{eq:svm_k2_het_dist_3}
\end{align}
Hence, we continue to prove \eqref{eq:svm_k2_het_dist_3}. We have
\begin{align}
\mathbb{E}_{A_k}\mathbb{E}_{\hat{\overline{{W}}}_k}&\left[\gen_{\theta}(s_{k},\overline{w})-\gen(s_{k},\hat{\overline{{W}}}_k)\right]\nonumber\\
        =&\mathbb{E}_{Z_k\sim \mu_k}\mathbb{E}_{A_k,\hat{\overline{{W}}}_k}\left[\ell_0(Z_k,\overline{w})-\tilde{\ell}_{\theta,k}(Z_k,\hat{\overline{W}}_k)\right] +\frac{1}{n}\sum_{i\in[n]} \mathbb{E}_{A_k,\hat{\overline{{W}}}_k}\left[\tilde{\ell}_{\theta,k}(z_{k,i},\hat{\overline{W}}_k)- \ell_{\theta}(z_{k,i},\overline{w})\right] \label{eq:svm_k2_het_dist_4}\\ 
        =&\mathbb{E}_{Z_k\sim \mu_k}\mathbb{E}_{A_k,\hat{\overline{{W}}}_k}\left[\mathbbm{1}_{\left\{Y_k\left(\left\langle X_k,\overline{w}\right\rangle\right)<0\right\}} -\mathbbm{1}_{\left\{Y_k\left(\langle X_k-b_k,\hat{\overline{W}}_{k,1}\rangle+\hat{\overline{W}}_{k,2}\right)<\frac{\theta}{2}\right\}}\right] \nonumber \\
        & +\frac{1}{n}\sum_{i\in[n]} \mathbb{E}_{A_k,\hat{\overline{{W}}}_k}\left[\mathbbm{1}_{\left\{y_{k,i}\left(\langle x_{k,i}-b_k,\hat{\overline{W}}_{k,1}\rangle+\hat{\overline{W}}_{k,2}\right)<\frac{\theta}{2}\right\}} - \mathbbm{1}_{\left\{y_{k,i}\left(\left\langle x_{k,i},\overline{w}\right\rangle\right)<\theta\right\}}\right]\nonumber \\
        \leq&\mathbb{E}_{Z_k\sim \mu_k}\mathbb{E}_{A_k,\hat{\overline{{W}}}_k}\left[\mathbbm{1}_{\left\{\left|\left\langle X_k,\overline{w}\right\rangle-\langle X_k-b_k,\hat{\overline{W}}_{k,1}\rangle-\hat{\overline{W}}_{k,2}\right|>\frac{\theta}{2}\right\}}\right]  +\frac{1}{n}\sum_{i\in[n]} \mathbb{E}_{A_k,\hat{\overline{{W}}}_k}\left[\mathbbm{1}_{\left\{\left|\left\langle x_{k,i},\overline{w}\right\rangle-\langle x_{k,i}-b_k,\hat{\overline{W}}_{k,1}\rangle-\hat{\overline{W}}_{k,2}\right|>\frac{\theta}{2}\right\}}\right]. \label{eq:svm_k2_het_dist_5} \end{align}
 where 
 \begin{itemize}
     \item $\hat{\overline{W}}_k\sim P_{{\hat{\overline{W}}_k}|w_{[K]},A_k}$,
     \item \eqref{eq:svm_k2_het_dist_4} is derived using the definitions of $\gen_{\theta}(s_{k},\overline{w})$ and $\gen\big(s_{k},\hat{\overline{{W}}}_k\big)$ (see \eqref{eq:svm_pr_gen}) and using the linearity of the expectation,
     \item and \eqref{eq:svm_k2_het_dist_5} is derived since for any $a,b\in \mathbb{R}$, we have 
     \begin{align}
         \mathbbm{1}_{\{a<0\}}-\mathbbm{1}_{\left\{b<\frac{\theta}{2}\right\}} \leq & \mathbbm{1}_{\left\{|a-b|>\frac{\theta}{2}\right\}}, \nonumber \\
         \mathbbm{1}_{\left\{a<\frac{\theta}{2}\right\}}-\mathbbm{1}_{\left\{b<\theta\right\}} \leq & \mathbbm{1}_{\left\{|a-b|>\frac{\theta}{2}\right\}},
     \end{align}
     and since $|Y|=|y_{k,i}|=1$, 
 \end{itemize}
To further upper bound \eqref{eq:svm_k2_het_dist_5}, we first show that far any $x_k \in \supp (\mu_{k,x})$, we have
 \begin{align}
     \mathbb{E}_{A_k,\hat{\overline{{W}}}_k}\left[\mathbbm{1}_{\left\{\left|\left\langle x_k,\overline{w}\right\rangle-\langle x_k-b_k,\hat{\overline{W}}_{k,1}\rangle-\hat{\overline{W}}_{k,2}\right|>\frac{\theta}{2}\right\}}\right] \leq\mathcal{O}\left(\frac{1}{nK\sqrt{K}}\right),\label{eq:svm_k2_het_dist_6}
 \end{align}
 where $\hat{\overline{W}}_k \sim P_{{\hat{\overline{W}}_k}|w_{[K]},A_k}$. Combining \eqref{eq:svm_k2_het_dist_6} with \eqref{eq:svm_k2_het_dist_5} proves \eqref{eq:svm_k2_het_dist_3}, and hence, as explained above, this completes the proof of the upper bound \eqref{eq:svm_pr_dist_rep} on the distortion. Hence, to complete the proof of \eqref{eq:svm_pr_dist_rep}, it remains to show \eqref{eq:svm_k2_het_dist_6}.

By using \eqref{def:svm_pr_w_hat_1} and \eqref{def:svm_pr_w_hat_2}, we 
have that 
\begin{align}
    \langle x_k-b_k,\hat{\overline{W}}_{k,1}\rangle+\hat{\overline{W}}_{k,2} = \frac{1}{K} \sum_{k'\neq k} \left\langle x_k , w_{k'}\right\rangle + \frac{1}{K} \langle x_k-b_k, A_k^{\top} W'_{k,1}\rangle+W^\prime_{k,2}.
\end{align}
Furthermore since $\overline{w}=\frac{1}{K}\sum_{k} w_{k}$, we have that
\begin{align}
    \left|\left\langle x_k,\overline{w}\right\rangle-\langle x_k-b_k,\hat{\overline{W}}_{k,1}\rangle-\hat{\overline{W}}_{k,2} \right| = &   \left|\frac{1}{K}  \langle x_k-b_k, w_k-A_k^{\top} W'_{k,1}\rangle +  \frac{1}{K}\left\langle b_k,w_k\right\rangle-W^\prime_{k,2}\right| \nonumber\\
    \leq & \left|\frac{1}{K}  \langle x_k-b_k, w_k-A_k^{\top} W'_{k,1}\rangle\right| + \left|  \frac{1}{K}\left\langle b_k,w_k\right\rangle-W^\prime_{k,2}\right| \label{eq:svm_pr_triangle} \\
    \leq & \left|\frac{1}{K}  \langle x_k-b_k, w_k-A_k^{\top} W'_{k,1}\rangle\right| + \frac{\theta}{2n}, \label{eq:svm_pr_w_2_quant}
\end{align}
where \eqref{eq:svm_pr_triangle} is derived using the triangle inequality and \eqref{eq:svm_pr_w_2_quant} follows by the definition of $W^\prime_{k,2}$ (see \eqref{def:svm_pr_w_p_2}).

Using \eqref{eq:svm_pr_w_2_quant}, the left-hand side of \eqref{eq:svm_k2_het_dist_6} can be upper bounded as 
\begin{align}
     \mathbb{E}_{A_k,\hat{\overline{{W}}}_k}\left[\mathbbm{1}_{\left\{\left|\left\langle x_k,\overline{w}\right\rangle-\langle x_k-b_k,\hat{\overline{W}}_{k,1}\rangle-\hat{\overline{W}}_{k,2}\right|>\frac{\theta}{2}\right\}}\right] \leq & \mathbb{E}_{A_k,\hat{\overline{{W}}}_k}\left[\mathbbm{1}_{\left\{\left|  \langle  \bar{x}_k , w_k-A_k^{\top} W'_{k,1}\rangle\right|>\frac{K \theta_n}{2}\right\}}\right], \label{eq:svm_k2_het_dist_7}
 \end{align}
 where  $\theta_n = \theta(1-1/n)$ and
 \begin{align}
     \bar{x}_k \coloneqq x_k-b_k.
 \end{align}
 Note that $\|\bar{x}_k\| \leq \rho_k$.

Recall that $W'_{k,1} \in \mathbb{R}^{m_k}$ is a random variable distributed uniformly over the $m_k$-dimensonal ball $B_{m_k}(A_k w_k,\nu_k)$, if $\| A_k w_k\|\leq \tau_{k,2}$, and otherwise  distributed uniformly over the $m_k$-dimensonal ball $B_{m_k}(0,\nu_k)$. Using simple algebras, we further upper bound the right-hand side of \eqref{eq:svm_k2_het_dist_7} as
\begin{align}
     \mathbb{E}_{A_k,\hat{\overline{{W}}}_k}\bigg[&\mathbbm{1}_{\left\{\left|\left\langle x_k,\overline{w}\right\rangle-\langle x_k-b_k,\hat{\overline{W}}_{k,1}\rangle-\hat{\overline{W}}_{k,2}\right|>\frac{\theta}{2}\right\}}\bigg]\nonumber \\
     \leq & \mathbb{E}_{A_k,\hat{\overline{{W}}}_k}\left[\mathbbm{1}_{\left\{\left|  \langle   \bar{x}_k , w_k- A_k^\top W'_{k,1}\rangle\right|>\frac{K \theta_n}{2},\,\| A_k \bar{x}_k\| \leq \rho_k \tau_{k,1},\,  \| A_k w_k\| \leq \tau_{k,2} \right\}} \right]\nonumber\\
     &+ \mathbb{E}_{A_k,\hat{\overline{{W}}}_k}\left[\mathbbm{1}_{\left\{\| A_k \bar{x}_k\| > \rho_k \tau_{k,1}\right\}} \right] + \mathbb{E}_{A_k,\hat{\overline{{W}}}_k}\left[\mathbbm{1}_{\left\{\| A_k w_k\| > \tau_{k,2}\right\}} \right]  \nonumber \\
     = & \mathbb{E}_{A_k}\mathbb{E}_{W'_{k,1}\sim \unif\left(B_{m_k}(A_k w_k,\nu_k)\right)}\left[\mathbbm{1}_{\left\{\left|  \langle   \bar{x}_k , w_k- A_k^\top W'_{k,1}\rangle\right|>\frac{K \theta_n}{2},\,\| A_k \bar{x}_k\| \leq \rho_k \tau_{k,1},\,  \| A_k w_k\| \leq \tau_{k,2} \right\}}  \right] \nonumber \\
     &+ \mathbb{E}_{A_k}\left[\mathbbm{1}_{\left\{\| A_k \bar{x}_k\| > \rho_k \tau_{k,1}\right\}} \right] + \mathbb{E}_{A_k}\left[\mathbbm{1}_{\left\{\| A_k w_k\| > \tau_{k,2}\right\}} \right]  \label{eq:svm_pr_steps_1} \\
     = & \mathbb{E}_{A_k}\mathbb{E}_{W'\sim \unif\left(B_{m_k}(\vc{0},\nu_k)\right)}\left[\mathbbm{1}_{\left\{\left| \langle   \bar{x}_k , w_k-A_k^\top W'-A_k^\top A_k w_k\rangle\right|>\frac{K \theta_n}{2},\,\| A_k \bar{x}_k\| \leq \rho_k \tau_{k,1},\,  \| A_k w_k\| \leq \tau_{k,2} \right\}}  \right] \nonumber \\
     &+ \mathbb{E}_{A_k}\left[\mathbbm{1}_{\left\{\| A_k \bar{x}_k\| > \rho_k \tau_{k,1}\right\}} \right] + \mathbb{E}_{A_k}\left[\mathbbm{1}_{\left\{\| A_k w_k\| > \tau_{k,2}\right\}} \right]  \nonumber \\
     \leq & \mathbb{E}_{A_k}\left[\mathbbm{1}_{\left\{\left|  \langle   \bar{x}_k , w_k-A_k^\top A_k w_k\rangle\right|>\frac{K \theta_n}{4},\,\| A_k \bar{x}_k\| \leq \rho_k \tau_{k,1}\,  ,\| A_k w_k\| \leq \tau_{k,2} \right\}}  \right] \nonumber \\
     &+ \mathbb{E}_{A_k}\mathbb{E}_{W'\sim \unif\left(B_{m_k}(\vc{0},\nu_k)\right)}\left[\mathbbm{1}_{\left\{\left| \langle   \bar{x}_k , A_k^\top W'\rangle\right|>\frac{K \theta_n}{4},\,\| A_k \bar{x}_k\| \leq \rho_k \tau_{k,1},\,  \| A_k w_k\| \leq \tau_{k,2} \right\}}  \right] \nonumber \\
     &+ \mathbb{E}_{A_k}\left[\mathbbm{1}_{\left\{\| A_k \bar{x}_k\| > \rho_k \tau_{k,1}\right\}} \right] + \mathbb{E}_{A_k}\left[\mathbbm{1}_{\left\{\| A_k w_k\| > \tau_{k,2}\right\}} \right]  \label{eq:svm_pr_steps_2} \\
     = & \mathbb{E}_{A_k}\left[\mathbbm{1}_{\left\{\left| \langle   \bar{x}_k , w_k\rangle -\langle A_k \bar{x}_k, A_k w_k\rangle\right|>\frac{K \theta_n}{4},\,\| A_k \bar{x}_k\| \leq \rho_k \tau_{k,1},\,  \| A_k w_k\| \leq \tau_{k,2} \right\}}  \right] \label{eq:svm_pr_steps_3}  \\
     &+ \mathbb{E}_{A_k}\mathbb{E}_{W'\sim \unif\left(B_{m_k}(\vc{0},\nu_k)\right)}\left[\mathbbm{1}_{\left\{\left| \langle   \bar{x}_k , A_k^\top W'\rangle\right|>\frac{K \theta_n}{4},\,\| A_k \bar{x}_k\| \leq \rho_k \tau_{k,1},\,  \| A_k w_k\| \leq \tau_{k,2}\right\}}  \right]\label{eq:svm_pr_steps_4}  \\
     &+ \mathbb{E}_{A_k}\left[\mathbbm{1}_{\left\{\| A_k \bar{x}_k\| > \rho_k \tau_{k,1}\right\}} \right] + \mathbb{E}_{A_k}\left[\mathbbm{1}_{\left\{\| A_k w_k\| > \tau_{k,2}\right\}} \right]  \label{eq:svm_pr_steps_5} 
 \end{align}
where
\begin{itemize}
    \item \eqref{eq:svm_pr_steps_1} is derived since whenever $\| A_k w_k\|\leq \tau_{k,2}$, $W'_{k,1} \sim \unif\left(B_{m_k}(A_k w_k,\nu_k)\right)$, 
    \item \eqref{eq:svm_pr_steps_2} is achieved using the triangle inequality and since $\langle   \bar{x}_k , w_k-A_k^\top A_k w_k\rangle $ does not depend on $W'$,
    \item and the last equality is deduced using the fact that
    \begin{align}
        \langle   \bar{x}_k , w_k-A_k^\top A_k w_k\rangle =  \langle   \bar{x}_k , w_k\rangle -\langle A_k \bar{x}_k, A_k w_k\rangle.
    \end{align}    
\end{itemize}
Finally, we bound each of the terms \eqref{eq:svm_pr_steps_3}, \eqref{eq:svm_pr_steps_4}, and \eqref{eq:svm_pr_steps_5}:
\begin{itemize}
    \item Using \cite[Lemma 8, part 2.]{gronlund2020}, \eqref{eq:svm_pr_steps_3} is upper bounded by
    \begin{align}
       \mathbb{E}_{A_k}\left[\mathbbm{1}_{\left\{\left| \langle   \bar{x}_k , w_k\rangle -\langle A_k \bar{x}_k, A_k w_k\rangle\right|>\frac{K \theta_n}{4},\,\| A_k \bar{x}_k\| \leq \rho_k \tau_{k,1}\, \| A_k w_k\| \leq \tau_{k,2} \right\}}  \right]  \leq &  4 e^{-\frac{m_k}{7}\left(\frac{K\theta_n}{4\rho_k}\right)^2} \nonumber \\
       =& \mathcal{O}\left(\frac{1}{nK\sqrt{K}}\right),
    \end{align}
    \item Using \cite[Lemma 3]{sefidgaran2022rate}, \eqref{eq:svm_pr_steps_4} is upper bounded by
    \begin{align}
      \mathbb{E}_{A_k}\mathbb{E}_{W'\sim \unif\left(B_{m_k}(\vc{0},\nu_k)\right)}\left[\mathbbm{1}_{\left\{\left| \langle   \bar{x}_k , A_k^\top W'\rangle\right|>\frac{K \theta_n}{4},\,\| A_k \bar{x}_k\| \leq \rho_k \tau_{k,1}\, \| A_k w_k\| \leq \tau_{k,2} \right\}}  \right]  \leq&  \frac{m_k{\nu_k}^{m_k}}{\sqrt{\pi}}e^{-\frac{(m_k+1)}{2}\left(\frac{K \theta_n}{4\tau_{k,1}\nu_k\rho_{k}}\right)^2} \nonumber \\
       =& \mathcal{O}\left(\frac{1}{nK\sqrt{K}}\right),
    \end{align}
    \item Using \cite[Lemma 8, part 1.]{gronlund2020}, \eqref{eq:svm_pr_steps_5} is upper bounded by
    \begin{align}
       \mathbb{E}_{A_k}\left[\mathbbm{1}_{\left\{\| A_k \bar{x}_k\| > \rho_k \tau_{k,1}\right\}} \right] + \mathbb{E}_{A_k}\left[\mathbbm{1}_{\left\{\| A_k w_k\| > \tau_{k,2}\right\}} \right]  \leq &  2e^{-0.21m_k(\tau_{k,1}^2-1)^2}+2e^{-0.21m_k(\tau_{k,2}^2-1)^2} \nonumber \\
       =& \mathcal{O}\left(\frac{1}{nK\sqrt{K}}\right).
    \end{align}
\end{itemize}
Combining the above bounds with \eqref{eq:svm_pr_steps_3}, \eqref{eq:svm_pr_steps_4}, and \eqref{eq:svm_pr_steps_5} proves \eqref{eq:svm_k2_het_dist_6} and hence, as explained above, this completes the proof of the upper bound \eqref{eq:svm_pr_dist} on the distortion.

\paragraph{Upper bounding the rate:}
Thus, it remains to upper bound the rate as in \eqref{eq:svm_pr_rate}. Fix $A_k$ as a matrix that satisfies the distortion constraint \eqref{eq:svm_pr_dist}. We have
   \begin{align}
       R_{\mathcal{D}_k}(\epsilon)\leq & I\Big(\hat{\overline{W}}_k;\mathbf{J}_k\lvert \mathfrak{Z}_{k}^{2n},W_{[K]\setminus{k}}\Big)  \label{eq:svm_pr_rate_def} \\
       = & I\Big(\hat{\overline{W}}_{k,1},\hat{\overline{W}}_{k,2};\mathbf{J}_k\lvert \mathfrak{Z}_{k}^{2n},W_{[K]\setminus{k}}\Big) \nonumber  \\
       = & I\Big(A_k^\top W'_{k,1},W'_{k,2};\mathbf{J}_k\lvert \mathfrak{Z}_{k}^{2n},W_{[K]\setminus{k}}\Big)  \label{eq:svm_pr_rate_def_2}    \\
       \leq & I\Big( W'_{k,1},W'_{k,2};\mathbf{J}_k\lvert \mathfrak{Z}_{k}^{2n},W_{[K]\setminus{k}}\Big)  \label{eq:svm_pr_rate_def_3}     \\
       = & h\Big( W'_{k,1},W'_{k,2}\lvert \mathfrak{Z}_{k}^{2n},W_{[K]\setminus{k}}\Big) - h\Big( W'_{k,1},W'_{k,2}\lvert \mathfrak{Z}_{k}^{2n},W_{[K]\setminus{k}},\mathbf{J}_k\Big) \nonumber\\
       = & h\Big( W'_{k,1},W'_{k,2}\lvert \mathfrak{Z}_{k}^{2n},W_{[K]\setminus{k}}\Big) - h\Big( W'_{k,1},W'_{k,2}\lvert S_k,W_{[K]\setminus{k}}\Big) \label{eq:svm_pr_rate_def_4}     \\
       \leq & h\Big( W'_{k,1},W'_{k,2}\lvert \mathfrak{Z}_{k}^{2n},W_{[K]\setminus{k}}\Big) - h\Big( W'_{k,1},W'_{k,2}\lvert S_k,W_{[K]}\Big) \label{eq:svm_pr_rate_def_5}   \\
       = & h\Big( W'_{k,1},W'_{k,2}\lvert \mathfrak{Z}_{k}^{2n},W_{[K]\setminus{k}}\Big) - h\Big( W'_{k,1}\lvert W_{k}\Big) \label{eq:svm_pr_rate_def_6}  \\
       \leq & h\big( W'_{k,1}\big) + H\big( W'_{k,2}\big) - h\Big( W'_{k,1}\lvert W_{k}\Big) \label{eq:svm_pr_rate_def_7}  \\
       \leq & \log\left(\text{Volume}\big(B_{m_k}(\vc{0},\tau_{k,2}+\nu_k)\big)\right) + \log(N_k)- \log\left(\text{Volume}\big(B_{m_k}(\vc{0},\nu_k)\big)\right) \label{eq:svm_pr_rate_def_8}   
        \\
       = &  m_k \log\left(\frac{\tau_{k,2}+\nu_k}{\nu_k}\right)+\log\left(\left[1,\frac{4 n \|b_k\|}{K\theta} \right]^+\right) \label{eq:svm_pr_rate_def_9}\\
       =&\mathcal{O}\left(\Big(\frac
    {\rho_{k}}{K\theta}\Big)^2\log(nK)\log\left(\left[3,\frac{K\theta}{\rho_k}\right]^+\right){+}{\log}\bigg(\bigg[1,\frac{4n\| b_k\|}{K\theta}\bigg]^+\right)\label{eq:svm_pr_rate_def_10},
    \end{align}
where
\begin{itemize}
    \item \eqref{eq:svm_pr_rate_def} follows by the definition of the rate-distortion function in  \eqref{rate-distortion-term} and since $P_{\hat{\overline{W}}_k | S_k, W_{[K]\setminus{k}}}$ satisfies the distortion criterion \eqref{eq:svm_pr_dist},
    \item \eqref{eq:svm_pr_rate_def_2} is derived using the definitions of $\hat{\overline{W}}_{k,1}$ and $\hat{\overline{W}}_{k,2}$ in \eqref{def:svm_pr_w_hat_1} and \eqref{def:svm_pr_w_hat_2}, respectively,
    \item \eqref{eq:svm_pr_rate_def_3} follows by data processing inequality,
    \item \eqref{eq:svm_pr_rate_def_4} holds since by the assumption of Theoerm~\ref{Lossy:inexpectation:general}, we have      \(P_{\hat{\overline{W}}_k\lvert \mathfrak{Z}_{k}^{2n},W_{[K]\setminus{k}},\mathbf{J}_k} = P_{\hat{\overline{W}}_k\lvert \mathfrak{Z}_{\mathbf{J}_k}^{2n},W_{[K]\setminus{k}}} \),
     \item \eqref{eq:svm_pr_rate_def_5} is deduced since conditioning reduces the entropy,
     \item \eqref{eq:svm_pr_rate_def_6} is derived since due the definitions of $(W'_{k,1},W'_{k,2})$, the Markov chain $(W'_{k,1},W'_{k,2})-W_k-(\mathfrak{Z}_{k}^{2n},W_{[K]\setminus{k}})$  holds, and since $W'_{k,2}$ is a deterministic function of $W_k$,
     \item \eqref{eq:svm_pr_rate_def_7} is deduced since conditioning reduces the entropy (note that $W'_{k,2}$ is a discrete random variable),
     \item and \eqref{eq:svm_pr_rate_def_9} is derived due to the following facts:
     \begin{itemize}
         \item[i)] $W'_{k,1}$ by definition \eqref{def:svm_pr_w_hat_1} is bounded in the $m_k$ dimensional ball with radius $(\tau_{k,2}+\nu_k)$, 
         \item[ii)] the differential entropy of a bounded variable is maximized under the uniform distribution, 
         \item[iii)] given $W_k$, $W'_{k,1}$ is distributed uniformly over either $B_{m_k}(A_k w_k,\nu_k)$ or $B_{m_k}(\vc{0},\nu_k)$, depending on the value of $\|A_k w_k\|$; which conclude that $ h\Big( W'_{k,1}\lvert W_{k}\Big)=\log\left(\text{Volume}\big(B_{m_k}(\vc{0},\nu_k)\big)\right)$ (note that the entropy is invariant under the translation, 
         \item[iv)] $W'_{k,2}$, by definition \eqref{def:svm_pr_w_hat_2}, takes at most $N_k=\left\lceil \frac{4 n \|b_k\|}{K\theta} \right\rceil$ different values and hence its entropy is bounded by $\log(N_k)$.
     \end{itemize}
\end{itemize}

This completes the proof of the upper bound \eqref{eq:svm_pr_rate}; and hence completes the proof of Theorem~\ref{general:svm:cmi}.


\subsection{Proof of Theorem \ref{SVM:h_d}}\label{pr:svm_hd}
We prove this result using Theorem \ref{h_d:ingeneral}, similar to how Theorem~\ref{general:svm:cmi} is proved using Theorem~\ref{Lossy:inexpectation:general} in Appendix~\ref{Proof:SVM:General:K}. More specifically:

 \begin{itemize}
     \item We consider the same auxiliary learning algorithm $P_{\hat{\overline{W}}_k|S_k,W_{[K]\setminus{k}}}$ as the one defined in the proof of Theorem \ref{general:svm:cmi} (see \eqref{eq:svm_pr_w_hat_def1} and \eqref{eq:svm_pr_w_hat_def2}). Hence, using \eqref{eq:svm_pr_dist}, we have  
\begin{align}
\mathbb{E}\left[\gen_{\theta}(S_{k},\overline{W})-\gen(S_{k},\hat{\overline{{W}}}_k)\right]\leq\mathcal{O}\left(\frac{1}{nK\sqrt{K}}\right). \label{eq:svm_hd_pr_dist}
\end{align}

\item Next, similar to how the ``rate'' was upper-bounded in the proof of Theorem~\ref{general:svm:cmi} (see \eqref{eq:svm_pr_rate_def_10}), it is straightforward to establish the below upper bound

\begin{align}
    I\Big(\hat{\overline{W}}_k;\mathbf{T}_k\lvert \mathfrak{Z}_{k}^{2n},W_{[K]\setminus{k}}\Big)
    &\leq\mathcal{O}\left(\Big(\frac
    {\rho_{k}}{K\theta}\Big)^2\log(nK)\log\left(\left[3,\frac{K\theta}{\rho_k}\right]^+\right){+}{\log}\bigg(\bigg[1,\frac{4n\| b_k\|}{K\theta}\bigg]^+\right).
\end{align}
 \end{itemize}

Now, applying the above bounds in Theorem~\ref{h_d:ingeneral} and using Lemma~\ref{h_d:lemma}, yield
\begin{equation}
  \mathbb{E}[\gen_{\theta}(S_{[K]},\overline{W})] \leq \mathcal{O}\left( h^{-1}_D\bigg( \hat{E}+\log(n)\Big| \hat{\mathcal{L}}_{\theta,k}\bigg)-\hat{\mathcal{L}}_{\theta,k}+\frac{1}{nK\sqrt{K}}\right),\label{eq:svm_hd_bound_init}
    \end{equation}
where
\begin{align}
\hat{\mathcal{L}}_{\theta,k} =& \mathbb{E}_{S_k,\hat{\overline{{W}}}_k}\left[\hat{\mathcal{L}}_{\theta,k}(S_k,\hat{\overline{{W}}}_k)\right], \nonumber\\
    \hat{\mathcal{L}}_{\theta,k}(S_k,\hat{\overline{{W}}}_k) = & \frac{1}{n}\sum_{i\in[n]} \tilde{\ell}_{\theta,k}(z_{k,i},\hat{\overline{W}}_k),
\end{align}
and $\tilde{\ell}_{\theta,k}$ is defined in \eqref{auxiliary:loss:general}.

Next, we establish the below upper bound on the difference of the empirical risks between the original and auxiliary learning algorithms:
\begin{align}   \mathbb{E}_{S_k,\overline{{W}},\hat{\overline{{W}}}_k}\left[{\hat{\mathcal{L}}_{\theta,k}(S_k,\hat{\overline{W}}_k)}-\hat{\mathcal{L}}_{\theta}(S_k,\overline{W})\right]& = \frac{1}{n}\sum_{i\in[n]} \mathbb{E}_{S_k,\overline{{W}},\hat{\overline{{W}}}_k}\left[\tilde{\ell}_{\theta,k}(z_{k,i},\hat{\overline{W}}_k)- \ell_{\theta}(z_{k,i},\overline{w})\right]  \leq \frac{9}{nK\sqrt{K}}, \label{eq:svm_hd_emp_dist}
\end{align}
This claim is, in fact, already shown (implicitly) in the proof of Theorem \ref{general:svm:cmi}, as it is equal to the expectation over $S_k$ of the second term in \eqref{eq:svm_k2_het_dist_4}, which is bounded as desired therein.

Finally, using item VII in Lemma~\ref{h_d:lemma} and \eqref{eq:svm_hd_emp_dist}, \eqref{eq:svm_hd_bound_init} can be upper bounded as:
\begin{equation}
  \mathbb{E}[\gen_{\theta}(S_{[K]},\overline{W})] \leq \mathcal{O}\left( h^{-1}_D\bigg( \hat{E}+\log(n)\Big| \mathbb{E}\left[\hat{\mathcal{L}}_{\theta}(S_{[K]},\overline{W})\right]-\frac{9}{nK\sqrt{K}}\bigg)-\mathbb{E}\left[\hat{\mathcal{L}}_{\theta}(S_{[K]},\overline{W})\right]+\frac{1}{nK\sqrt{K}}\right).\label{eq:svm_hd_bound_final}
    \end{equation}


\subsection{Proof of Theorem \ref{Gaussian_generalization_theorem}}\label{Proof_Gaussian_generalization_theorem}
The proof is similar to that of Theorem \ref{general:svm:cmi}. For every $r\in[M]$, consider the substitutions
\begin{align}
    \mu_k&\coloneqq\mu^{(r)}_k,\nonumber\\
    b_k&\coloneqq b_k^{(r)}=\sum\nolimits_{m=c^{(r)}_k}^{c^{(r)}_k+r-1}\alpha^{(r)}_{k,m}{a_{m}},\nonumber\\
    \rho_k&\coloneqq\rho^{(r)}_{k}={D_{k,r}}+\sigma\sqrt{\log(nK)},\nonumber\\
    c_k&\coloneqq c^{(r)}_k,\nonumber\\
    \alpha_{k,m}&\coloneqq\alpha^{(r)}_{k,m}.\nonumber
\end{align}

Also, for every $k\in[K]$ consider the loss function $\tilde{\ell}_{\theta,k} \colon \mathcal{Z} \times \hat{\mathcal{W}} \to \{0,1\}$ defined as
\begin{align}\label{auxiliary:gaussian:loss:general}
    \tilde{\ell}_{\theta,k}(z_k,\hat{w}) = \mathbbm{1}_{\left\{y_k\left(\langle x_k-b_k,\hat{w}_1\rangle+\hat{w}_2\right)<\frac{\theta}{2}\right\}},\quad \hat{w}\in \hat{\mathcal{W}},\: z_k\in \mathcal{Z},
\end{align}
where $\hat{w}=(\hat{w}_1,\hat{w}_2)$ with $\hat{w}_1\in\mathbb{R}^{d}$ and $\hat{w}_2\in\mathbb{R}$.
Recall that
\begin{align}
    \gen(s_{k},\hat{w}) = \mathbb{E}_{Z_k\sim \mu_k}\left[ \tilde{\ell}_{\theta,k}(Z_k,\hat{w})\right] - \frac{1}{n} \sum_{i\in[n]} \tilde{\ell}_{\theta,k}(z_{i,k},\hat{w}).
    \label{eq:gen-error-quantized-model-SVM}
\end{align}
Also, consider 
auxiliary algorithm $P_{\hat{\overline{W}}_k|S_k,W_{[K]\setminus{k}}}$ such that
\begin{align}
\mathbb{E}\left[\gen_{\theta}(S_{k},\overline{W})-\gen(S_{k},\hat{\overline{{W}}}_k)\right]\leq\mathcal{O}\left(\frac{1}{nK\sqrt{K}}\right).
\end{align}
The rest of the proof follows essentially by showing (see below) that $R_{\mathcal{D}_k}(\epsilon)$ can be upper bounded as 
\begin{equation}
R_{\mathcal{D}_k}(\epsilon) \leq \mathcal{O}\left(\left(\frac{\rho_k}{K\theta}\right)^2\log\left(\bar{E}^{(r)}_k\right)\log (nK) +\log\left(\tilde{E}_k^{(r)}\right)\right),
\label{eq:upper-bound-rate-distortion-term-theorem10}
    \end{equation}
    where $\bar{E}^{(r)}_k=\left[ 3,\frac{K\theta}{\sigma}\right]^{+}$ and $\tilde{E}_k^{(r)}=\left[1,\frac{4n\parallel b_{k}\parallel}{K\theta}\right]^{+}$; and then combining with Theorem~\ref{Lossy:inexpectation:general} to get the desired result.

We now show~\eqref{eq:upper-bound-rate-distortion-term-theorem10}. Let 
\begin{align}
    &m_k\coloneqq\left\lceil 112\left(\frac{\rho_k}{K\theta}\right)^2\log(nK\sqrt{K})\right\rceil\label{constants_gaussian_1}\\
    &\tau_{k,1}=\tau_{k,2}=\sqrt{1+\frac{K\theta_n}{4\sigma}}\label{constants_gaussian_2}\\
    &\nu_k\coloneqq\frac{1}{\tau_{k,1}}\label{constants_gaussian_3}\\
    &\theta_n\coloneqq\theta\left(1-\frac{1}{n}\right)\label{constants_gaussian_4}
\end{align}
Consider $\hat{\overline{W}}_k=(\hat{\overline{W}}_{k,1},\hat{\overline{W}}_{k,2})$, where $\hat{\overline{W}}_{k,1}$ and $\hat{\overline{W}}_{k,2}$ are defined as in~\eqref{def:svm_pr_w_p_1} and~\eqref{def:svm_pr_w_hat_2}, respectively. We start by showing that
\begin{align}
\mathbb{E}\left[\gen_{\theta}(S_{k},\overline{W})-\gen(S_{k},\hat{\overline{{W}}}_k)\right]\leq\mathcal{O}\left(\frac{1}{nK\sqrt{K}}\right), \label{eq:svm_pr_dist_rep:gaussian}
\end{align}
where $\gen(S_{k},\hat{\overline{{W}}}_k)$ is defined as in\eqref{eq:gen-error-quantized-model-SVM}. To this end, we show that 
\begin{align}
\mathbb{E}_{A_k}\mathbb{E}_{S_{k},\overline{W},\hat{\overline{{W}}}_k}\left[\gen_{\theta}(S_{k},\overline{W})-\gen(S_{k},\hat{\overline{{W}}}_k)\right]\leq\mathcal{O}\left(\frac{1}{nK\sqrt{K}}\right), \label{eq:svm_k2_het_dist_2:gaussian}
\end{align}
where $A_k\in\mathbb{R}^{m_k\times d}$ is generated exactly in the same manner as in the proof of Theorem \ref{general:svm:cmi}. For that choice of $A_k$, we get 
\begin{align}
    \mathbb{E}_{S_k}\mathbb{E}_{A_k}&\mathbb{E}_{\hat{\overline{{W}}}_k}\left[\gen_{\theta}(S_{k},\overline{w})-\gen(S_{k},\hat{\overline{{W}}}_k)\right]\nonumber\\
    \leq&\mathbb{E}_{Z\sim \mu_k}\mathbb{E}_{A_k,\hat{\overline{{W}}}_k}\left[\mathbbm{1}_{\left\{\left|\left\langle X_k,\overline{w}\right\rangle-\langle X_k-b_k,\hat{\overline{W}}_{k,1}\rangle-\hat{\overline{W}}_{k,2}\right|>\frac{\theta}{2}\right\}}\right]\\  &+\frac{1}{n}\sum_{i\in[n]} \mathbb{E}_{S_k}\mathbb{E}_{A_k,\hat{\overline{{W}}}_k}\left[\mathbbm{1}_{\left\{\left|\left\langle X_{k,i},\overline{w}\right\rangle-\langle X_{k,i}-b_k,\hat{\overline{W}}_{k,1}\rangle-\hat{\overline{W}}_{k,2}\right|>\frac{\theta}{2}\right\}}\right].\label{difference_population_gneral_svm_1}
\end{align}
By considering $\bar{X}_k\coloneqq X_k-b_k$, we obtain
\begin{align}
  &\mathbb{E}_{Z\sim \mu_k}\mathbb{E}_{A_k,\hat{\overline{{W}}}_k}\left[\mathbbm{1}_{\left\{\left|\left\langle X_k,\overline{w}\right\rangle-\langle X_k-b_k,\hat{\overline{W}}_{k,1}\rangle-\hat{\overline{W}}_{k,2}\right|>\frac{\theta}{2}\right\}}\right]\nonumber\\
  &\qquad\qquad=\mathbb{E}_{A_k}\left[\mathbbm{1}_{\left\{\left| \langle   \bar{X}_k , w_k\rangle -\langle A_k \bar{X}_k, A_k w_k\rangle\right|>\frac{K \theta_n}{4},\,\| A_k \bar{X}_k\| \leq \tau_{k,1}\|\bar{X}_k\| ,\,  \| A_k w_k\| \leq \tau_{k,2} \right\}}  \right]\nonumber\\
     &\quad\qquad\qquad+ \mathbb{E}_{A_k}\mathbb{E}_{W'\sim \unif\left(B_{m_k}(\vc{0},\nu_k)\right)}\left[\mathbbm{1}_{\left\{\left| \langle   \bar{X}_k , A_k^\top W'\rangle\right|>\frac{K \theta_n}{4},\,\| A_k \bar{X}_k\| \leq \tau_{k,1}\|\bar{X}_k\| ,\,  \| A_k w_k\| \leq \tau_{k,2}\right\}}\right]\nonumber\\
     &\quad\qquad\qquad+ \mathbb{E}_{A_k}\left[\mathbbm{1}_{\left\{\| A_k \bar{X}_k\|> \tau_{k,1}\|\bar{X}_k\| \right\}} \right] + \mathbb{E}_{A_k}\left[\mathbbm{1}_{\left\{\| A_k w_k\| > \tau_{k,2}\right\}} \right]\label{splitting_gaussian_loss_4}
\end{align}
\begin{itemize}
    \item Using \cite[Lemma 8, part 2.]{gronlund2020}, the first term of the sum of the RHS of \eqref{splitting_gaussian_loss_4} satisfies
    \begin{align}
         \mathbb{E}_{\bar{X}_k}&\mathbb{E}_{A_k}\left[\mathbbm{1}_{\left\{\left| \langle   \bar{X}_k , w_k\rangle -\langle A_k \bar{X}_k, A_k w_k\rangle\right|>\frac{K \theta_n}{4},\,\| A_k \bar{X}_k\| \leq \tau_{k,1}\|\bar{X}_k\| \, ,\| A_k w_k\| \leq \tau_{k,2}\right\}}\right]\label{expectation:normal:1}\\
         &\leq\mathbb{E}_{\bar{X}_k}\left[4 e^{-\frac{m_k}{7}\left(\frac{K\theta_n}{4\|\bar{X}_k\|}\right)^2}\right]\nonumber\\
          &=\mathbb{E}_{\|\bar{X}_k\|}\left[4 e^{-\frac{m_k}{7}\left(\frac{K\theta_n}{4\|\bar{X}_k\|}\right)^2}\right]\label{expectation:normal:2}\\
         &=2\sum_{m\in[c_k,c_k+r-1]}\alpha_{k,m}\int_{0}^{+\infty}e^{-\frac{m_k}{7}\left(
         \frac{K\theta_n}{\sigma\left(u+\frac{\|a_m-b_k\|}{\sigma}\right)}\right)^2}e^{-\frac{u^2}{2}}du\label{expectation:normal:5}\\
         &\leq 2\sum_{m\in[c_k,c_k+r-1]}\alpha_{k,m}\int_{0}^{t}e^{-\frac{m_k}{7}\left(
         \frac{K\theta_n}{\sigma\left(t+\frac{\|a_m-b_k\|}{\sigma}\right)}\right)^2}e^{-\frac{u^2}{2}}du\label{expectation:normal:7}\\
         &\quad+2\sum_{m\in[c_k,c_k+r-1]}\alpha_{k,m}\int_{t}^{\infty}e^{-\frac{m_k}{7}\left(
         \frac{K\theta_n}{\sigma\left(u+\frac{\|a_m-b_k\|}{\sigma}\right)}\right)^2}e^{-\frac{u^2}{2}}du\label{expectation:normal:8}\\
         &\leq2\sum_{m\in[c_k,c_k+r-1]}\alpha_{k,m}\left[e^{-\frac{m_k}{7}\left(
         \frac{K\theta_n}{\sigma\left(t+\frac{\|a_m-b_k\|}{\sigma}\right)}\right)^2}+e^{-\frac{t^2}{2}}\right],\label{expectation:normal:9}\\
         &\leq\mathcal{O}\left(\frac{1}{nK\sqrt{K}}\right)\label{expectation:normal:10}
    \end{align}
    where, using the fact the random variable $\|X\|$ is Gaussian distributed with mean $\|a_m\|$ and variance $\sigma^2$, we have:
    \begin{itemize}
    \item \eqref{expectation:normal:5} follows by the definition of mixture distribution of $\|\bar{X}_k\|$ and combining with the inequality $\|X+a_m-b_k\|\leq \|X\|+\|a_m-b_k\|$.
    \item \eqref{expectation:normal:9} holds since the Gausian distribution is  clearly subgaussian and; so, the following inequality holds,
    \begin{align}
        \int_{t}^{\infty}e^{-\frac{u^2}{2}}du\leq e^{-\frac{t^2}{2}}.
    \end{align}
    \end{itemize}
        Then, using the that for every $k\in[K]$ and $m\in[{c_k},{c_k+r-1}]$ we have $\|a_m-b_k\|\leq{D_{k,r}}$, letting $t=\sqrt{\log(nK\sqrt{K})}$ and choosing $m_k$ as in~\eqref{constants_gaussian_1} we get that the first term of the sum of the RHS of~\eqref{splitting_gaussian_loss_4} is upper bounded by $\mathcal{O}\left(\frac{1}{{nK\sqrt{K}}}\right)$.

        \item  Using \cite[Lemma 3]{sefidgaran2022rate}, the second term of the RHS of~\eqref{splitting_gaussian_loss_4} is such that
    \begin{align}
      &\mathbb{E}_{\bar{X}_k}\left[\mathbb{E}_{A_k}\mathbb{E}_{W'\sim \unif\left(B_{m_k}(\vc{0},\nu_k)\right)}\left[\mathbbm{1}_{\left\{\left| \langle   \bar{x}_k , A_k^\top W'\rangle\right|>\frac{K \theta_n}{4},\,\| A_k \bar{x}_k\| \leq \tau_{k,1}\|\bar{X}_k\| \, ,\| A_k w_k\| \leq \tau_{k,2} \right\}}\right]\right]\label{expectation:normal:11}\\
      &\qquad\leq\mathbb{E}_{\bar{X}_k}\left[\frac{m_k{\nu_k}^{m_k}}{\sqrt{\pi}}e^{-\frac{(m_k+1)}{2}\left(\frac{K \theta_n}{4\tau_{k,1}\nu_k\|\bar{X}_k\|}\right)^2}\right]\nonumber\\
      &\qquad=\mathbb{E}_{\|\bar{X}_k\|}\left[\frac{m_k{\nu_k}^{m_k}}{\sqrt{\pi}}e^{-\frac{(m_k+1)}{2}\left(\frac{K \theta_n}{4\tau_{k,1}\nu_k\|\bar{X}_k\|}\right)^2}\right]\nonumber\\
      &\qquad\leq2\sum_{m\in[c_k,c_k+r-1]}\alpha_{k,m}\int_{0}^{+\infty}\frac{m_k{\nu_k}^{m_k}}{\sqrt{\pi}}e^{-\frac{(m_k+1)}{2}\left(\frac{K \theta_n}{4\sigma\tau_{k,1}\nu_k\left(u+\frac{\|a_m-b_k\|}{\sigma}\right)}\right)^2}e^{-\frac{u^2}{2}}d{u}\label{expectation:normal:12}\\
      &\qquad\leq 2\sum_{m\in[c_k,c_k+r-1]}\alpha_{k,m}\int_{0}^{t}\frac{m_k{\nu_k}^{m_k}}{\sqrt{\pi}}e^{-\frac{(m_k+1)}{2}\left(\frac{K \theta_n}{4\sigma\tau_{k,1}\nu_k\left(t+\frac{\|a_m-b_k\|}{\sigma}\right)}\right)^2}e^{-\frac{u^2}{2}}d{u}\label{expectation:normal:13}\\
      &\qquad\qquad +2\sum_{m\in[c_k,c_k+r-1]}\alpha_{k,m}\int_{t}^{\infty}\frac{m_k{\nu_k}^{m_k}}{\sqrt{\pi}}e^{-\frac{(m_k+1)}{2}\left(\frac{K \theta_n}{4\sigma\tau_{k,1}\nu_k\left(u+\frac{\|a_m-b_k\|}{\sigma}\right)}\right)^2}e^{-\frac{u^2}{2}}d{u}\label{expectation:normal:14}\\
      &\qquad\leq2\sum_{m\in[c_k,c_k+r-1]}\alpha_{k,m}\left[\frac{m_k{\nu_k}^{m_k}}{\sqrt{\pi}}e^{-\frac{(m_k+1)}{2}\left(\frac{K \theta_n}{4\sigma\tau_{k,1}\nu_k\left(t+\frac{\|a_m-b_k\|}{\sigma}\right)}\right)^2}+e^{-\frac{t^2}{2}}\right]\label{expectation:normal:15}\\
       &\qquad=\mathcal{O}\left(\frac{1}{nK\sqrt{K}}\right),\label{expectation:normal:16}
    \end{align}
    where~\eqref{expectation:normal:12} follows by noticing that $\|{X}\|$ is Gaussian distributed with mean $\|a_m\|$ and variance $\sigma^2$, and combining using $\|{X}+a_m-b_k\|\leq\|X\|+\|a_m-b_k\|$ with $t=\sqrt{\log(nK\sqrt{K})}$ and $m_k$ chosen as in~\eqref{constants_gaussian_1}.
  
\item  Using \cite[Lemma 8, part 1.]{gronlund2020}, the third term of the sum of the RHS of~\eqref{splitting_gaussian_loss_4} is upper bounded as
    \begin{align}
       \mathbb{E}_{A_k}\left[\mathbbm{1}_{\left\{\| A_k \bar{x}_k\| >  \tau_{k,1}\|X_k\|\right\}} \right] + \mathbb{E}_{A_k}\left[\mathbbm{1}_{\left\{\| A_k w_k\| > \tau_{k,2}\right\}} \right]  \leq &  2e^{-0.21m_k(\tau_{k,1}^2-1)^2}+2e^{-0.21m_k(\tau_{k,2}^2-1)^2} \nonumber \\
       =& \mathcal{O}\left(\frac{1}{nK\sqrt{K}}\right).
    \end{align}
\end{itemize}
    
Combining the above, we get \eqref{difference_population_gneral_svm_1}; and this establishes the distortion constraint \eqref{eq:svm_pr_dist_rep:gaussian}.

It remain to bound $R_{D_k}(\epsilon)$ as desired. This is done by fixing a matrix $A_k$ such that~\eqref{eq:svm_pr_dist_rep:gaussian} is satisfied and proceeding as in the steps~\eqref{eq:svm_pr_rate_def}-\eqref{eq:svm_pr_rate_def_9} while substituting using \eqref{constants_gaussian_1}-\eqref{constants_gaussian_4} to get 
    \begin{align}
    R_{D_k}(\epsilon)\leq\mathcal{O}\left(\left(\frac{\rho_k}{K\theta}\right)^2\log\left(\bar{E}^{(r)}_k\right)\log (nK) +\log\left(\tilde{E}_k^{(r)}\right)\right),
    \end{align}
    where $\bar{E}^{(r)}_k=\left[ 3,\frac{K\theta}{\sigma}\right]^{+}$ and $\tilde{E}_k^{(r)}=\left[1,\frac{4n\parallel b_{k}\parallel}{K\theta}\right]^{+}$. This completes the proof of Theorem~\ref{Gaussian_generalization_theorem}


\subsection{Proof of Lemma~\ref{h_d:lemma}}\label{Proof:Lemma:auxiliary:h_d:property}
\paragraph{Proof of $h^{-1}_D(y|0)\leq y$} For $y\in[0,2]$, define the set $A_y$  as
\begin{align}
A_y=\{x\in[0,1]:h_D(x,0)\leq y\}. \label{def:setA}
\end{align}
Using the definition of $h_D^{-1}(y|0)$, it is easy to see that $h_D^{-1}(y|0)=\sup {A_y}$. 
Now, By Lemma \ref{h_d:lemma} we know that $h_D(y,0) \geq y$. This combining with the monotonicity increasing of $h_D(x,0)$ in $x$ implies that $y\geq\sup {A_y}=h_D^{-1}(y|0)$, which completes the proof.

\paragraph{Proof of $h^{-1}_D(y|c)\leq c+\sqrt{y}$} Similar to the first part, for $c\in[0,1]$ and $y\in[0,2]$, define the set $B_y$  as 
\begin{align}
    B_y=\{x\in[0,1]:h_D(x,c)\leq y\}.
\end{align}
It is easy to see that $h^{-1}_D(y|c)=\sup B_y$ . Now using Lemma \ref{h_d:lemma} we know that $h_D(\sqrt{y}+c,c)\geq y$. This combining with the monotonicity increasing of $h_D(x,c)$ for $x\in[c,1]$ implies that $c+\sqrt{y}\geq\sup B_y=h^{-1}_D(y|c)$, and this completes the proof. 

\paragraph{Proof of item G}

For simplicity, let's denote $f_{a,b}(x) \coloneqq h_D(a+x,b+X)$ and without loss of generality assume that $a\geq b$.  It is sufficient to show that 
\begin{align}
f'_{a,b}(x) \coloneqq \frac{\partial f_{a,b}(x)}{\partial x} \leq 0, \quad  \text{for } x \in \left[0,\frac{1}{2}-a\right].      
\end{align}
Simple algebra yields
\begin{align}
f'_{a,b}(x) = -2 \log\left(\frac{a+b+2x}{2-(a+b+2x)}\right)+\log\left(\frac{a+x}{1-(a+x)}\right)+\log\left(\frac{b+x}{1-(b+x)}\right).      
\end{align}
To show that $f'_{a,b}(x)\leq 0$, we derive the $\max_{a\in[b,1/2]} f'_{a,b}(x)$. We have
\begin{align}
     \frac{\partial f'_{a,b}(x)}{\partial a} = \frac{1}{(a+x)(1-(a+x))}-\frac{1}{\left(\frac{a+x+b+x}{2}\right)\left(1-\frac{a+x+b+x}{2}\right)} \leq 0,
\end{align}
where the inequality is achieved since $0 \leq \frac{a+x+b+x}{2} \leq a+x \leq \frac{1}{2}$ and the function $y(1-y)$ is increasing in the range $y\in[0,\frac{1}{2}]$.

Hence $\max_{a\in[b,1/2]} f'_{a,b}(x)$ is achieved for $a=b$. Thus,
\begin{align}
f'_{a,b}(x) \leq f'_{b,b}(x) =0;      
\end{align}
And this completes the proof.




\subsection{Proof of Lemma \ref{simp:expect:loss:1}}\label{Proof:auxiliary:lemma:h_d}

Let us consider the set of independent binary random variables $\{V_1, V_2, \dots, V_{2n}\}$, where $V_i \in \{0, 1\}$ is independent of the others, and $V_i \sim \text{Bern}(\ell_i)$, for $i\in[2n]$. Then, we have

\begin{align}
&\mathbb{E}_{\mathbf{T}\sim\unif(2n)}\left[e^{nh_D\left(\frac{1}{n}\sum\nolimits_{i\in\mathbf{T}}\ell_i,\frac{1}{n}\sum\nolimits_{i'\in\mathbf{T}^c}\ell_i'\right)}\right]\nonumber\\
&\qquad\qquad=\mathbb{E}_{\mathbf{T}\sim\unif(2n)}\bigg[e^{nh_D\left(\mathbb{E}_{V_{T_1}},\cdots,\mathbb{E}_{V_{T_n}}\left[\frac{1}{n}\sum_{i\in\mathbf{T}}V_{i}\right],   \mathbb{E}_{V_{T^c_1}},\cdots,\mathbb{E}_{V_{T^c_n}}\left[\frac{1}{n}\sum_{i'\in{\mathbf{T}^c}}V_{i'}\right]\right)}\bigg]\label{simplification:expect:5}\\
&\qquad\qquad\leq\mathbb{E}_{\mathbf{T}\sim\unif(2n)}\left[\mathbb{E}_{V_{T_1}},\mathbb{E}_{V_{T^c_1}},\cdots,\mathbb{E}_{V_{T_n}},\mathbb{E}_{V_{T^c_n}}\left[e^{nh_D\left(\frac{1}{n}\sum_{i\in\mathbf{T}}V_{i},\frac{1}{n}\sum\nolimits_{i'\in\mathbf{T}^c}V_{i'}\right)}\right]\right]\label{simplification:expect:6}\\
&\qquad\qquad=\mathbb{E}_{\mathbf{T}\sim\unif(2n)}\left[\mathbb{E}_{V_1},\mathbb{E}_{V_{2}},\cdots,\mathbb{E}_{V_{2n-1}},\mathbb{E}_{V_{2n}}\left[e^{nh_D\left(\frac{1}{n}\sum_{i\in\mathbf{T}}V_{i},\frac{1}{n}\sum\nolimits_{i'\in\mathbf{T}^c}V_{i'}\right)}\right]\right]\nonumber\\
&\qquad\qquad=\mathbb{E}_{V_1},\mathbb{E}_{V_{2}},\cdots,\mathbb{E}_{V_{2n-1}},\mathbb{E}_{V_{2n}}\bigg[\mathbb{E}_{\mathbf{T}\sim\unif(2n)}\left[e^{nh_D\left(\frac{1}{n}\sum_{i\in\mathbf{T}}V_{i},\frac{1}{n}\sum_{i'\in\mathbf{T}^c}V_{i'}\right)}\right]\bigg]\nonumber\\
&\qquad\qquad\leq n,\label{simplification:expect:8}
    \end{align}
    where 
  
    \begin{itemize}
        \item\eqref{simplification:expect:5} is derived using the fact that  $\forall i\in[2n]$, we have $\mathbb{E}\left[V_{i}\right]=\ell_{i}$.
        \item the convexity of $f(x) = \exp(x)$ in $x$ and $g(x_1, x_2) = h_D(x_1, x_2)$ in both $x$ and $x^\prime$ \cite{sefidgaran2024minimum} implies the inequality in \eqref{simplification:expect:6}.
        \item and equation \eqref{simplification:expect:8} is concluded from equation (32) in \cite{sefidgaran2024minimum}, where, based on that, 
        \begin{align}
\mathbb{E}_{\mathbf{T}\sim\unif(2n)}\left[e^{nh_D\left(\frac{1}{n}\sum_{i\in\mathbf{T}}V_{i},\frac{1}{n}\sum_{i'\in\mathbf{T}^c}V_{i'}\right)}\right]\leq n.
        \end{align}
    \end{itemize}








\end{document}

%% file: defs_IT.tex
\usepackage{bm,amsfonts,bbm,float,euscript,setspace,tikz,steinmetz,enumitem,wrapfig}

\usepackage[normalem]{ulem}
\usepackage[mode=buildnew]{standalone}
\usepackage{mathabx} 
\newtheorem{theorem}{Theorem}
\newtheorem{lemma}{Lemma}
\newtheorem{definition}{Definition}

\usepackage{mleftright}\mleftright

\definecolor{darkgreen}{rgb}{0, 0.5, 0}
\definecolor{darkred}{RGB}{128, 0, 0}

\newcommand{\stkout}[1]{\ifmmode\text{\sout{\ensuremath{#1}}}\else\sout{#1}\fi}

\newcommand{\vc}[1]{\mathbf{#1}} 

\newcommand{\ie}{\emph{i.e., }}

\DeclareMathOperator{\gen}{gen}

\DeclareMathOperator{\unif}{Unif}

\DeclareMathOperator{\supp}{supp}

\DeclareMathOperator*{\argmin}{arg\,min}

\makeatletter
\DeclareFontFamily{OMX}{MnSymbolE}{}
\DeclareSymbolFont{MnLargeSymbols}{OMX}{MnSymbolE}{m}{n}
\SetSymbolFont{MnLargeSymbols}{bold}{OMX}{MnSymbolE}{b}{n}
\DeclareFontShape{OMX}{MnSymbolE}{m}{n}{
	<-6>  MnSymbolE5
	<6-7>  MnSymbolE6
	<7-8>  MnSymbolE7
	<8-9>  MnSymbolE8
	<9-10> MnSymbolE9
	<10-12> MnSymbolE10
	<12->   MnSymbolE12
}{}
\DeclareFontShape{OMX}{MnSymbolE}{b}{n}{
	<-6>  MnSymbolE-Bold5
	<6-7>  MnSymbolE-Bold6
	<7-8>  MnSymbolE-Bold7
	<8-9>  MnSymbolE-Bold8
	<9-10> MnSymbolE-Bold9
	<10-12> MnSymbolE-Bold10
	<12->   MnSymbolE-Bold12
}{}

\let\llangle\@undefined
\let\rrangle\@undefined
\DeclareMathDelimiter{\llangle}{\mathopen}%
{MnLargeSymbols}{'164}{MnLargeSymbols}{'164}
\DeclareMathDelimiter{\rrangle}{\mathclose}%
{MnLargeSymbols}{'171}{MnLargeSymbols}{'171}
\makeatother


%% file: main_TIT2025.bbl
\begin{thebibliography}{10}
\providecommand{\url}[1]{#1}
\csname url@samestyle\endcsname
\providecommand{\newblock}{\relax}
\providecommand{\bibinfo}[2]{#2}
\providecommand{\BIBentrySTDinterwordspacing}{\spaceskip=0pt\relax}
\providecommand{\BIBentryALTinterwordstretchfactor}{4}
\providecommand{\BIBentryALTinterwordspacing}{\spaceskip=\fontdimen2\font plus
\BIBentryALTinterwordstretchfactor\fontdimen3\font minus \fontdimen4\font\relax}
\providecommand{\BIBforeignlanguage}[2]{{%
\expandafter\ifx\csname l@#1\endcsname\relax
\typeout{** WARNING: IEEEtran.bst: No hyphenation pattern has been}%
\typeout{** loaded for the language `#1'. Using the pattern for}%
\typeout{** the default language instead.}%
\else
\language=\csname l@#1\endcsname
\fi
#2}}
\providecommand{\BIBdecl}{\relax}
\BIBdecl

\bibitem{mcmahan2017communication}
B.~McMahan, E.~Moore, D.~Ramage, S.~Hampson, and B.~A. y~Arcas, ``Communication-efficient learning of deep networks from decentralized data,'' in \emph{Artificial intelligence and statistics}.\hskip 1em plus 0.5em minus 0.4em\relax PMLR, 2017, pp. 1273--1282.

\bibitem{konevcny2016federated}
J.~Kone{\v{c}}n{\`y}, H.~B. McMahan, F.~X. Yu, P.~Richt{\'a}rik, A.~T. Suresh, and D.~Bacon, ``Federated learning: Strategies for improving communication efficiency,'' \emph{arXiv preprint arXiv:1610.05492}, 2016.

\bibitem{yuan2021we}
H.~Yuan, W.~Morningstar, L.~Ning, and K.~Singhal, ``What do we mean by generalization in federated learning?'' \emph{arXiv preprint arXiv:2110.14216}, 2021.

\bibitem{gupta2018distributed}
O.~Gupta and R.~Raskar, ``Distributed learning of deep neural network over multiple agents,'' \emph{Journal of Network and Computer Applications}, vol. 116, pp. 1--8, 2018.

\bibitem{aguerri2019distributed}
I.~E. Aguerri and A.~Zaidi, ``Distributed variational representation learning,'' \emph{IEEE transactions on pattern analysis and machine intelligence}, vol.~43, no.~1, pp. 120--138, 2019.

\bibitem{moldoveanu2023network}
M.~Moldoveanu and A.~Zaidi, ``In-network learning: Distributed training and inference in networks,'' \emph{Entropy}, vol.~25, no.~6, p. 920, 2023.

\bibitem{li2020federatedsurvey}
T.~Li, A.~K. Sahu, A.~Talwalkar, and V.~Smith, ``Federated learning: Challenges, methods, and future directions,'' \emph{IEEE Signal Processing Magazine}, vol.~37, no.~3, pp. 50--60, 2020.

\bibitem{kairouz2021advances}
P.~Kairouz, H.~B. McMahan, B.~Avent, A.~Bellet, M.~Bennis, A.~N. Bhagoji, K.~Bonawitz, Z.~Charles, G.~Cormode, R.~Cummings \emph{et~al.}, ``Advances and open problems in federated learning,'' \emph{Foundations and trends{\textregistered} in machine learning}, vol.~14, no. 1--2, pp. 1--210, 2021.

\bibitem{dwork2014algorithmic}
C.~Dwork and A.~Roth, \emph{The Algorithmic Foundations of Differential Privacy}.\hskip 1em plus 0.5em minus 0.4em\relax Foundations and Trends in Theoretical Computer Science, 2014.

\bibitem{geyer2017differentially}
R.~C. Geyer, T.~Klein, and M.~Nabi, ``Differentially private federated learning: A client level perspective,'' in \emph{NIPS Workshop on Machine Learning on the Phone and other Consumer Devices}, 2017.

\bibitem{abadi2016deep}
M.~Abadi, A.~Chu, and I.~e.~a. Goodfellow, ``Deep learning with differential privacy,'' in \emph{Proceedings of the ACM SIGSAC Conference on Computer and Communications Security}, 2016.

\bibitem{mohri2019agnostic}
M.~Mohri, G.~Sivek, and A.~T. Suresh, ``Agnostic federated learning,'' in \emph{International conference on machine learning}.\hskip 1em plus 0.5em minus 0.4em\relax PMLR, 2019, pp. 4615--4625.

\bibitem{zinkevich2010parallelized}
M.~Zinkevich, M.~Weimer, L.~Li, and A.~Smola, ``Parallelized stochastic gradient descent,'' \emph{Advances in neural information processing systems}, vol.~23, 2010.

\bibitem{yagli2020information}
S.~Yagli, A.~Dytso, and H.~V. Poor, ``Information-theoretic bounds on the generalization error and privacy leakage in federated learning,'' in \emph{2020 IEEE 21st International Workshop on Signal Processing Advances in Wireless Communications (SPAWC)}.\hskip 1em plus 0.5em minus 0.4em\relax IEEE, 2020, pp. 1--5.

\bibitem{barnes2022improved}
L.~P. Barnes, A.~Dytso, and H.~V. Poor, ``Improved information theoretic generalization bounds for distributed and federated learning,'' in \emph{2022 IEEE International Symposium on Information Theory (ISIT)}.\hskip 1em plus 0.5em minus 0.4em\relax IEEE, 2022, pp. 1465--1470.

\bibitem{zhang2024improving}
H.~Zhang, C.~Li, N.~Kan, Z.~Zheng, W.~Dai, J.~Zou, and H.~Xiong, ``Improving generalization in federated learning with model-data mutual information regularization: A posterior inference approach,'' \emph{Advances in Neural Information Processing Systems}, vol.~37, pp. 136\,646--136\,678, 2024.

\bibitem{Sefidgaran2022}
M.~Sefidgaran, A.~Gohari, G.~Richard, and U.~Simsekli, ``Rate-distortion theoretic generalization bounds for stochastic learning algorithms,'' in \emph{Conference on Learning Theory}.\hskip 1em plus 0.5em minus 0.4em\relax PMLR, 2022, pp. 4416--4463.

\bibitem{chor2023more}
R.~Chor, M.~Sefidgaran, and A.~Zaidi, ``More {{Communication Does Not Result}} in {{Smaller Generalization Error}} in {{Federated Learning}},'' in \emph{2023 {{IEEE International Symposium}} on {{Information Theory}} ({{ISIT}})}, 2023, pp. 48--53.

\bibitem{sefidgaran2024lessons}
M.~Sefidgaran, R.~Chor, A.~Zaidi, and Y.~Wan, ``Lessons from generalization error analysis of federated learning: You may communicate less often!'' in \emph{Forty-first International Conference on Machine Learning}, 2024.

\bibitem{hu2023generalization}
X.~Hu, S.~Li, and Y.~Liu, ``Generalization bounds for federated learning: Fast rates, unparticipating clients and unbounded losses,'' in \emph{The Eleventh International Conference on Learning Representations}, 2023.

\bibitem{aleks2020pac}
D.~Aleks and M.~Jaggi, ``Pac-bayesian analysis of decentralized learning algorithms,'' in \emph{Advances in Neural Information Processing Systems}, 2020.

\bibitem{neyshabur2017pac}
B.~Neyshabur, S.~Bhojanapalli, D.~McAllester, and N.~Srebro, ``A pac-bayesian approach to spectrally-normalized margin bounds for neural networks,'' in \emph{International Conference on Learning Representations}, 2018.

\bibitem{li2020federatedpac}
M.~Li, T.~Yang, S.~Song, and T.~Wang, ``Federated pac-bayes learning,'' in \emph{Proceedings of the AAAI Conference on Artificial Intelligence}, 2020.

\bibitem{smith2017federated}
V.~Smith, C.~K. Chiang, M.~Sanjabi, and A.~Talwalkar, ``Federated multi-task learning,'' in \emph{Advances in Neural Information Processing Systems}, 2017.

\bibitem{arivazhagan2019federated}
M.~G. Arivazhagan, V.~Aggarwal, A.~K. Singh, and S.~Choudhary, ``Federated learning with personalization layers,'' in \emph{NeurIPS Workshop on Federated Learning}, 2019.

\bibitem{sattler2020clustered}
F.~Sattler, S.~Wiedemann, K.~R. Müller, and W.~Samek, ``Clustered federated learning: Model-agnostic distributed multi-task optimization under privacy constraints,'' in \emph{IEEE Transactions on Neural Networks and Learning Systems}, 2020.

\bibitem{aji2017sparse}
A.~F. Aji and K.~Heafield, ``Sparse communication for distributed gradient descent,'' in \emph{Proceedings of the Empirical Methods in Natural Language Processing}, 2017.

\bibitem{reisizadeh2020fedpaq}
A.~Reisizadeh, A.~Mokhtari, H.~Hassani, A.~Jadbabaie, and R.~Pedarsani, ``Fedpaq: A communication-efficient federated learning method with periodic averaging of quantized gradients,'' in \emph{Proceedings of the AAAI Conference on Artificial Intelligence}, 2020.

\bibitem{lin2018deep}
Y.~Lin, S.~Han, H.~Mao, Y.~Wang, and W.~J. Dally, ``Deep gradient compression: Reducing the communication bandwidth for distributed training,'' in \emph{International Conference on Learning Representations}, 2018.

\bibitem{wang2022unreasonable}
J.~Wang, R.~Das, G.~Joshi, S.~Kale, Z.~Xu, and T.~Zhang, ``On the unreasonable effectiveness of federated averaging with heterogeneous data,'' \emph{arXiv preprint arXiv:2206.04723}, 2022.

\bibitem{zhang2021fedpd}
X.~Zhang, M.~Hong, S.~Dhople, W.~Yin, and Y.~Liu, ``Fedpd: A federated learning framework with adaptivity to non-iid data,'' \emph{IEEE Transactions on Signal Processing}, vol.~69, pp. 6055--6070, 2021.

\bibitem{mitra2021linear}
A.~Mitra, R.~Jaafar, G.~J. Pappas, and H.~Hassani, ``Linear convergence in federated learning: Tackling client heterogeneity and sparse gradients,'' \emph{Advances in Neural Information Processing Systems}, vol.~34, pp. 14\,606--14\,619, 2021.

\bibitem{steinke2020reasoning}
T.~Steinke and L.~Zakynthinou, ``Reasoning about generalization via conditional mutual information,'' in \emph{Conference on Learning Theory}.\hskip 1em plus 0.5em minus 0.4em\relax PMLR, 2020, pp. 3437--3452.

\bibitem{sefidgaran2024data}
M.~Sefidgaran and A.~Zaidi, ``Data-dependent generalization bounds via variable-size compressibility,'' \emph{IEEE Transactions on Information Theory}, 2024.

\bibitem{sefidgaran2024minimum}
M.~Sefidgaran, A.~Zaidi, and P.~Krasnowski, ``Minimum description length and generalization guarantees for representation learning,'' \emph{Advances in Neural Information Processing Systems}, vol.~36, 2024.

\bibitem{sefidgaran2022rate}
M.~Sefidgaran, R.~Chor, and A.~Zaidi, ``Rate-distortion theoretic bounds on generalization error for distributed learning,'' \emph{Advances in Neural Information Processing Systems}, vol.~35, pp. 19\,687--19\,702, 2022.

\bibitem{sun2024understanding}
Z.~Sun, X.~Niu, and E.~Wei, ``Understanding generalization of federated learning via stability: Heterogeneity matters,'' in \emph{International Conference on Artificial Intelligence and Statistics}.\hskip 1em plus 0.5em minus 0.4em\relax PMLR, 2024, pp. 676--684.

\bibitem{wang2025generalization}
Z.~Wang, C.~Long, and Y.~Mao, ``Generalization in federated learning: A conditional mutual information framework,'' \emph{arXiv preprint arXiv:2503.04091}, 2025.

\bibitem{liu2023exploiting}
R.~Liu, J.~Yang, and C.~Shen, ``Exploiting feature heterogeneity for improved generalization in federated multi-task learning,'' in \emph{2023 IEEE International Symposium on Information Theory (ISIT)}.\hskip 1em plus 0.5em minus 0.4em\relax IEEE, 2023, pp. 180--185.

\bibitem{li2020federated2}
T.~Li, A.~K. Sahu, M.~Zaheer, M.~Sanjabi, A.~Talwalkar, and V.~Smith, ``Federated optimization in heterogeneous networks,'' \emph{Proceedings of Machine learning and systems}, vol.~2, pp. 429--450, 2020.

\bibitem{karimireddy2020scaffold}
S.~P. Karimireddy, S.~Kale, M.~Mohri, S.~Reddi, S.~Stich, and A.~T. Suresh, ``Scaffold: Stochastic controlled averaging for federated learning,'' in \emph{International conference on machine learning}.\hskip 1em plus 0.5em minus 0.4em\relax PMLR, 2020, pp. 5132--5143.

\bibitem{wang2020tackling}
J.~Wang, Q.~Liu, H.~Liang, G.~Joshi, and H.~V. Poor, ``Tackling the objective inconsistency problem in heterogeneous federated optimization,'' \emph{Advances in neural information processing systems}, vol.~33, pp. 7611--7623, 2020.

\bibitem{JMLR:v24:21-0224}
\BIBentryALTinterwordspacing
S.~Chen, Q.~Zheng, Q.~Long, and W.~J. Su, ``Minimax estimation for personalized federated learning: An alternative between fedavg and local training?'' \emph{Journal of Machine Learning Research}, vol.~24, no. 262, pp. 1--59, 2023. [Online]. Available: \url{http://jmlr.org/papers/v24/21-0224.html}
\BIBentrySTDinterwordspacing

\bibitem{woodworth2020local}
B.~Woodworth, K.~K. Patel, S.~Stich, Z.~Dai, B.~Bullins, B.~Mcmahan, O.~Shamir, and N.~Srebro, ``Is local sgd better than minibatch sgd?'' in \emph{International Conference on Machine Learning}.\hskip 1em plus 0.5em minus 0.4em\relax PMLR, 2020, pp. 10\,334--10\,343.

\bibitem{woodworth2020minibatch}
B.~E. Woodworth, K.~K. Patel, and N.~Srebro, ``Minibatch vs local sgd for heterogeneous distributed learning,'' \emph{Advances in Neural Information Processing Systems}, vol.~33, pp. 6281--6292, 2020.

\bibitem{wang2021cooperative}
J.~Wang and G.~Joshi, ``Cooperative sgd: A unified framework for the design and analysis of local-update sgd algorithms,'' \emph{Journal of Machine Learning Research}, vol.~22, no. 213, pp. 1--50, 2021.

\bibitem{gronlund2020}
A.~Gr{\o}nlund, L.~Kamma, and K.~G. Larsen, ``Near-tight margin-based generalization bounds for support vector machines,'' in \emph{Proceedings of the 37th International Conference on Machine Learning}, ser. Proceedings of Machine Learning Research, H.~D. III and A.~Singh, Eds., vol. 119.\hskip 1em plus 0.5em minus 0.4em\relax PMLR, 13--18 Jul 2020, pp. 3779--3788.

\bibitem{lecun1998mnist}
Y.~LeCun, ``The mnist database of handwritten digits,'' \emph{http://yann. lecun. com/exdb/mnist/}, 1998.

\bibitem{wainwright2019high}
M.~J. Wainwright, \emph{High-dimensional statistics: A non-asymptotic viewpoint}.\hskip 1em plus 0.5em minus 0.4em\relax Cambridge university press, 2019, vol.~48.

\end{thebibliography}
